\documentclass[10pt,journal,compsoc]{IEEEtran}

\usepackage{cite}
\usepackage{amsmath,amssymb,amsfonts}
\usepackage{algorithmic}
\usepackage{times}
\usepackage[ruled,vlined,linesnumbered]{algorithm2e}
\usepackage{graphicx}
\usepackage{float}
\usepackage{rotating, caption}
\usepackage{subfigure}
\usepackage{textcomp}
\usepackage{xcolor}
\usepackage{multirow}
\usepackage{array}
\usepackage{mathrsfs}
\usepackage{booktabs}
\usepackage{varwidth}
\usepackage{threeparttable}
\usepackage{caption} 
\usepackage{bm}
\usepackage{pifont}
\usepackage{bibunits}
\usepackage[colorlinks]{hyperref}

  
\def\ie{\textit{i.e.,}\xspace}

\def\eg{\textit{e.g.,}\xspace}
\def\etal{\textit{et al.}\xspace}

\ifCLASSINFOpdf

\else

\fi
\begin{document}
\title{A Comprehensive Survey on\\ Graph Anomaly Detection with Deep Learning}

\author{Xiaoxiao Ma,
        Jia Wu, \IEEEmembership{Senior Member,~IEEE,}
        Shan Xue,
        Jian Yang,
        Chuan Zhou \\
        Quan Z. Sheng,
        and Hui Xiong,~\IEEEmembership{Fellow,~IEEE}
        and Leman Akoglu
\IEEEcompsocitemizethanks{\IEEEcompsocthanksitem X. Ma, J. Wu, S. Xue, J. Yang, Q. Z. Sheng are with Department
of Computing, Macquarie University, Sydney, NSW 2109, Australia. 
E-mail: \{xiaoxiao.ma2@students., jia.wu, emma.xue, jian.yang, michael.sheng\}mq.edu.au. 

\IEEEcompsocthanksitem Chuan Zhou is with the Academy of Mathematics and Systems Science,
Chinese Academy of Sciences, Beijing, China. E-mail: zhouchuan@amss.ac.cn.

\IEEEcompsocthanksitem H. Xiong is with Department of Management Science and Information Systems, Rutgers, the State University of New Jersey, USA. Email: hxiong@rutgers.edu.

\IEEEcompsocthanksitem Leman Akoglu is the Heinz College Dean's Associate Professor at Carnegie Mellon University's Heinz College of Information Systems and Public Policy, USA. Email: lakoglu@andrew.cmu.edu.\protect\\
Corresponding author: Jia Wu}

}

\markboth{Journal of \LaTeX\ Class Files,~Vol.~, No.~, August~2021}%
{Shell \MakeLowercase{\textit{et al.}}: Bare Demo of IEEEtran.cls for Computer Society Journals}

\IEEEtitleabstractindextext{%
\begin{abstract}
Anomalies are rare observations (\eg data records or events) that deviate significantly from the others in the sample. Over the past few decades, research on anomaly mining has received increasing interests due to the implications of these occurrences in a wide range of disciplines - for instance, security, finance, and medicine. For this reason, anomaly detection, which aims to identify these rare observations, has become one of the most vital tasks in the world and has shown its power in preventing detrimental events, such as financial fraud, network intrusions, and social spam. The detection task is typically solved by identifying outlying data points in the feature space, which, inherently, overlooks the relational information in real-world data. At the same time, graphs have been prevalently used to represent the structural/relational information, which raises the \textit{graph anomaly detection problem} - identifying anomalous graph objects (\ie nodes, edges and sub-graphs) in a single graph, or anomalous graphs in a set/database of graphs. Conventional anomaly detection techniques cannot tackle this problem well because of the complexity of graph data (\eg irregular structures, relational dependencies, node/edge types/attributes/directions/multiplicities/weights, large scale, etc.). However, thanks to the advent of deep learning in breaking these limitations, graph anomaly detection with deep learning has received a growing attention recently. 
In this survey, we aim to provide a systematic and comprehensive review of the contemporary deep learning techniques for graph anomaly detection. 
Specifically, we provide a taxonomy that follows a task-driven strategy and categorizes existing work according to the anomalous graph objects that they can detect. We especially focus on the challenges in this research area and discuss the key intuitions, technical details as well as relative strengths and weaknesses of various techniques in each category.
From the survey results, we highlight 12 future research directions spanning unsolved and emerging problems introduced by graph data, anomaly detection, deep learning and real-world applications. 
Additionally, to provide a wealth of useful resources for future studies, we have compiled a set of open-source implementations, public datasets, and commonly-used evaluation metrics. 
With this survey, our goal is to create a ``one-stop-shop'' that provides a unified understanding of the problem categories and existing approaches, publicly available hands-on resources, and high-impact open challenges for graph anomaly detection using deep learning.  
\end{abstract}

}

\maketitle
\IEEEdisplaynontitleabstractindextext
\IEEEpeerreviewmaketitle

\IEEEraisesectionheading{\section{Introduction}\label{sec:introduction}}
\IEEEPARstart{A}{nomalies} were first defined by Grubbs in 1969~\cite{grubbs1969procedures} as \textit{``one that appears to deviate markedly from other members of the sample in which it occurs''} and the studies on anomaly detection were initiated by the statistics community in the 19th century.
To us, anomalies might appear as social spammers or misinformation in social media; fraudsters, bot users or sexual predators in social networks; network intruders or malware in computer networks and broken devices or malfunctioning blocks in industry systems, and they often introduce huge damage to the real-world systems they appear in.
According to FBI's 2014 Internet Crime Report\footnote{https://www.fbi.gov/file-repository/2014\_ic3report.pdf/view}, the financial loss due to crime on social media reached more than \$60 million in the second half of the year alone and a more up-to-date report\footnote{https://www.zdnet.com/article/online-fake-news-costing-us-78-billion-globally-each-year/} indicates that the global economic cost of online fake news reached around \$78 billion a year in 2020.

In computer science, the research on anomaly detection dates back to the 1980s, and detecting anomalies on graph data has been an important data mining paradigm since the beginning.
However, the extensive presence of connections between real-world objects and advances in graph data mining in the last decade have revolutionized our understanding of the graph anomaly detection problems such that this research field has received a dramatic increase in interest over the past five years.
One of the most significant changes is that graph anomaly detection has evolved from relying heavily on human experts' domain knowledge into machine learning techniques that eliminate human intervention, and more recently, to various deep learning technologies.
These deep learning techniques are not only capable of identifying potential anomalies in graphs far more accurately than ever before, but they can also do so in real-time.
 
\begin{figure}[!t]
\setlength{\belowcaptionskip}{-0.5cm}
\centering
\subfigure[Conventional Anomaly Detection]{
\label{Fig1a}
\includegraphics[scale=0.45]{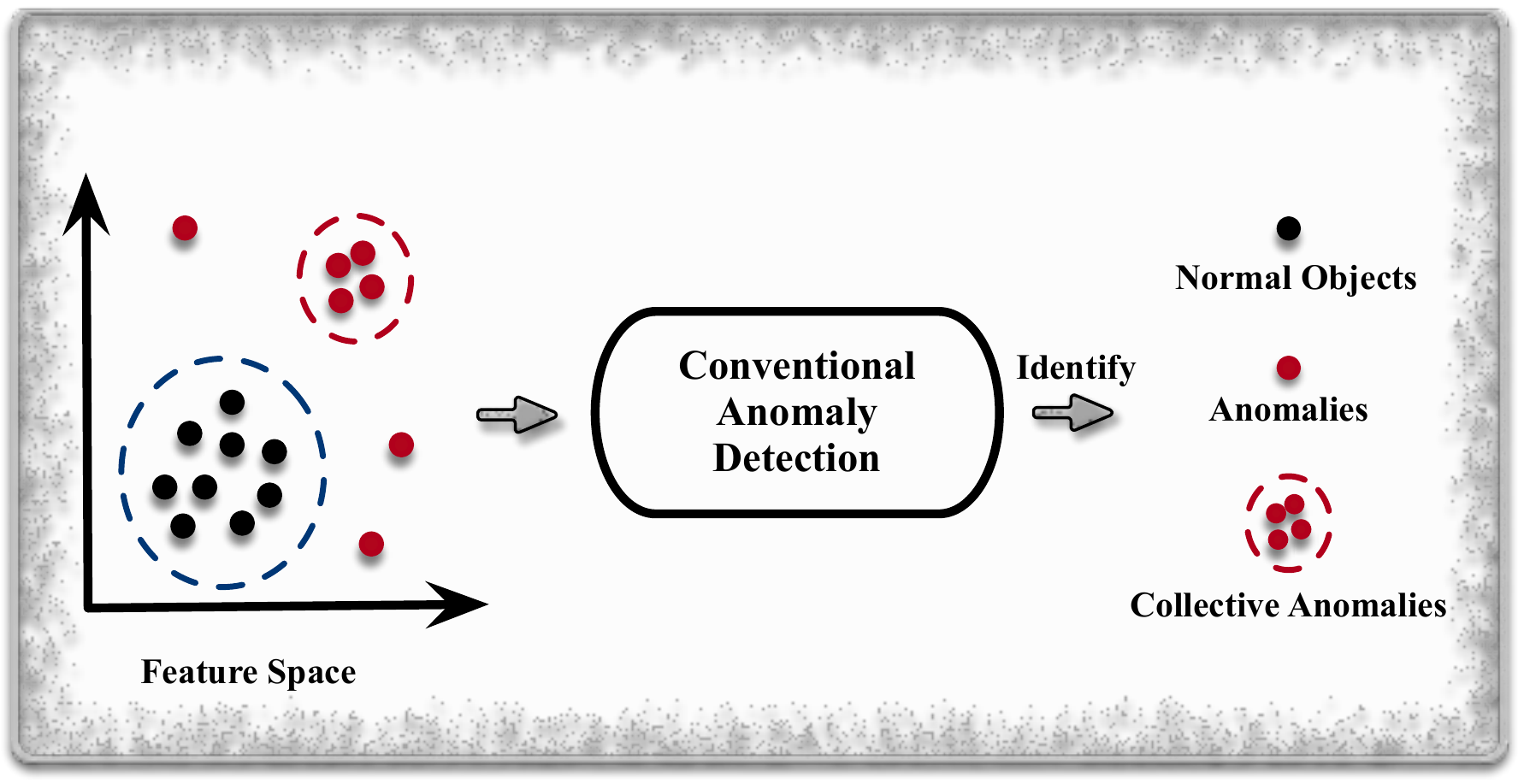}}
\subfigure[Graph Anomaly Detection]{
\label{Fig1b}
\includegraphics[scale=0.45]{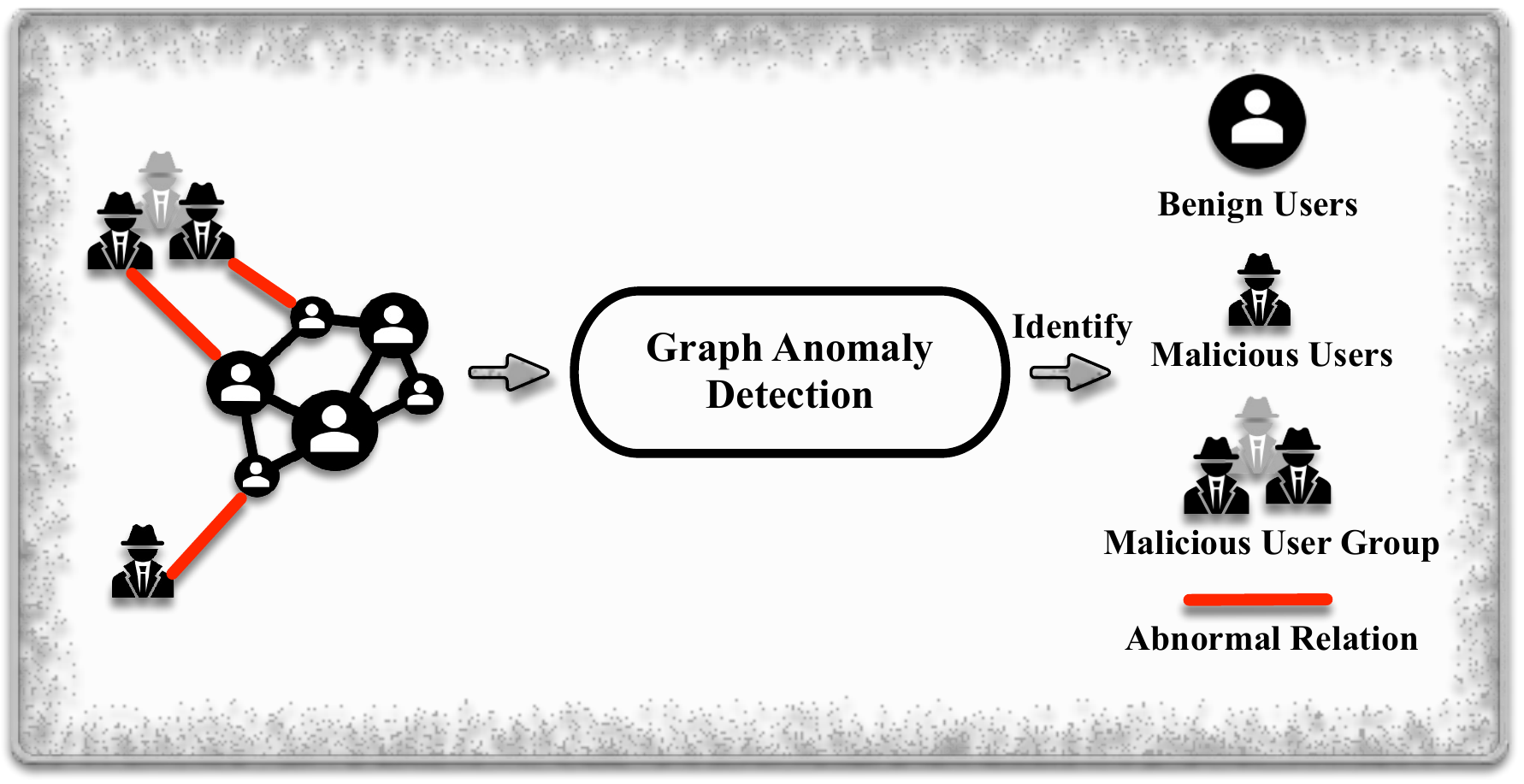}} 

\caption{Toy Examples of Conventional Anomaly Detection and Graph Anomaly Detection. Apart from anomalies shown in (b), graph anomaly detection also identifies graph-level anomalies, detailed in Sections~\ref{sec:anosgd:db} and~\ref{sec:anosgd:dynamic}.} 
\label{Toy}
\end{figure}

For our purposes today, anomalies, which are also known as outliers, exceptions, peculiarities, rarities, novelties, etc., in different application fields, refer to abnormal objects that are significantly different from the standard, normal, or expected. 
Although these objects rarely occur in real-world, they contain critical information to support downstream applications. For example, the behaviors of fraudsters provide evidences for anti-fraud detection and abnormal network traffics reveal signals for network intrusion protection. Anomalies, in many cases, may also have real and adverse impacts, for instance, fake news in social media can create panic and chaos with misleading beliefs~\cite{hooi2016birdnest,ahmed2019combining,nguyen2020fang,tam2019anomaly}, untrustworthy reviews in online review systems can affect customers' shopping choices~\cite{yu2016survey,benamira2019semi,kumar2018rev2}, network intrusions might leak private personal information to hackers~\cite{mongiovi2013netspot,miller2013efficient,DBLP:journals/compsec/MiaoSZ20,perozzi2016scalable}, and financial frauds can cause huge damage to economic systems~\cite{xuexiong2022,10.1145/3394486.3403361,10.1145/3394486.3403354,DBLP:conf/www/GuoLAHHZZ19}.

Anomaly detection is the data mining process that aims to identify the unusual patterns that deviate from the majorities in a dataset~\cite{iglewicz1993detect,chandola2009anomaly,Fraud2020}. In order to detect anomalies, conventional techniques typically represent real-world objects as feature vectors (\eg news in social media are represented as bag-of-words~\cite{sun2018detecting}, and images in web pages are represented as color histograms~\cite{DBLP:conf/icdm/WuHPZCZ14}), and then detect outlying data points in the vector space~\cite{DBLP:conf/ijcai/WangL20,DBLP:conf/kdd/PangCCL18,DBLP:journals/corr/abs-2007-02500}, as shown in Fig.~\ref{Fig1a}. 
Although these techniques have shown power in locating deviating data points under tabulated data format, they inherently discard the complex relationships between objects~\cite{akoglu2015graph}.

Yet, in reality, many objects have rich relationships with each other, which can provide valuable complementary information for anomaly detection.
Take online social networks as an example, fake users can be created using valid information from normal users or they can camouflage themselves by mimicking benign users' attributes~\cite{hooi2017graph,CARE-GNN}. 
In such situations, fake users and benign users would have near-identical features, and conventional anomaly detection techniques might not be able to identify them using feature information only. 
Meanwhile, fake users always build relationships with a large number of benign users to increase their reputation and influence so they can get unexpected benefits, whereas benign users rarely exhibit such activities~\cite{pandit2007netprobe,Densealert}.
Hence, these dense and unexpected connections formed by fake users denote their deviations to the benigns and more comprehensive detection techniques should take these structural information into account to pinpoint the deviating patterns of anomalies.

To represent the structural information, \textit{Graphs}, in which nodes/vertices denote real objects, and the edges denote their relationships, have been prevalently used in a range of application fields~\cite{liu2008spotting,wu2014multi,gao2017collaborative,aggarwal2011outlier,wu2017multiple,NIPS20201}, including social activities, e-commerce, biology, academia and communication. With the structural information contained in graphs, detecting anomalies in graphs raises a more complex anomaly detection problem in non-Euclidean space - graph anomaly detection (GAD) that aims to identify anomalous graph objects (\ie nodes, edges or sub-graphs) in a single graph as well as anomalous graphs among a set/database of graphs~\cite{akoglu2015graph,DBLP:journals/jiis/ChenHS12,GBGP}.
As a toy example shown in Fig.~\ref{Fig1b}, given an online social network, graph anomaly detection aims to identify anomalous nodes (\ie malicious users), anomalous edges (\ie abnormal relations) and anomalous sub-graphs (\ie malicious user groups).
But, because the copious types of graph anomalies cannot be directly represented in Euclidean feature space, it is not feasible to directly apply traditional anomaly detection techniques to graph anomaly detection, and researchers have intensified their efforts to GAD recently.

Amongst earlier works in this area, the detection methods relied heavily on handcrafted feature engineering or statistical models built by domain experts~\cite{akoglu2010oddball,eswaran2018spotlight,li2014probabilistic}.
This inherently limits these techniques' capability to detect unknown anomalies, and the exercise tended to be very labor-intensive. 
Many machine learning techniques, such as matrix factorization~\cite{li2017radar,DBLP:journals/pnas/MahoneyD09} and SVM~\cite{DBLP:journals/pr/ErfaniRKL16}, have also been applied to detect graph anomalies.
However, real-world networks often contain millions of nodes and edges that result in extremely high dimensional and large-scale data, and these techniques do not easily scale up to such data efficiently.
Practically, they exhibit high computational overhead in both the storage and execution time~\cite{DBLP:journals/jbd/ThudumuBJS20}.
These general challenges associated with graph data are significant for the detection techniques, and we categorize them as data-specific challenges (Data-CHs) in this survey.
A summary of them is provided in Appendix~\ref{appendix:data-challenges}.

\begin{table*}[!h]
\centering 
\footnotesize
\renewcommand\arraystretch{1}
\setlength{\tabcolsep}{2.8mm}{
\caption{A Comparison Between Existing Surveys on Anomaly Detection. We mark edge, sub-graph and graph detection as \includegraphics[scale=0.2]{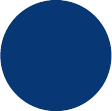} in our survey because we review more deep learning based works than any previous surveys.} 
\resizebox{0.96\textwidth}{!}{
\begin{tabular}{l|c|c|c|c|c|c|c|c|c|c}
\toprule[1 pt]
\multirow{2}{*}{\textbf{Surveys}} & \multirow{2}{*}{\textbf{AD}} & \multirow{2}{*}{\textbf{DAD}} & \multirow{2}{*}{\textbf{GAD}} 
& \multicolumn{4}{c|}{\textbf{GADL}} & \multirow{2}{*}{\textbf{Source Code}} & \multicolumn{2}{c}{\textbf{Dataset}}  \\ \cline{5-8}  \cline{10-11}
& & & &\textbf{Node}  &\textbf{Edge}  & \textbf{Sub-graph} & \textbf{Graph} & &\textbf{Real-world}  &\textbf{Synthetic} \\ 
\midrule[1 pt]

Our Survey & \includegraphics[scale=0.15]{pics/Introduction/4.pdf} & \includegraphics[scale=0.15]{pics/Introduction/4.pdf}& \includegraphics[scale=0.15]{pics/Introduction/4.pdf}& \includegraphics[scale=0.15]{pics/Introduction/4.pdf}& \includegraphics[scale=0.15]{pics/Introduction/4.pdf}& \includegraphics[scale=0.15]{pics/Introduction/4.pdf}& \includegraphics[scale=0.15]{pics/Introduction/4.pdf}& \includegraphics[scale=0.15]{pics/Introduction/4.pdf}& \includegraphics[scale=0.15]{pics/Introduction/4.pdf}& \includegraphics[scale=0.15]{pics/Introduction/4.pdf} \\

\midrule[1 pt]

Chandola \etal~\cite{chandola2009anomaly} & \includegraphics[scale=0.15]{pics/Introduction/4.pdf} & - & - & - & - & - & -  & - & - & - \\
Boukerche \etal~\cite{DBLP:journals/csur/BoukercheZA20} & \includegraphics[scale=0.15]{pics/Introduction/4.pdf} & \includegraphics[scale=0.15]{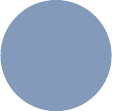} & - & - & - & - & -  & - & - & - \\
Bulusu \etal~\cite{DBLP:journals/corr/abs-2003-06979} & \includegraphics[scale=0.15]{pics/Introduction/4.pdf} & \includegraphics[scale=0.15]{pics/Introduction/4.pdf} & - & - & - & - & - & - & - & - \\
Thudumu \etal~\cite{DBLP:journals/jbd/ThudumuBJS20} & \includegraphics[scale=0.15]{pics/Introduction/4.pdf} & \includegraphics[scale=0.15]{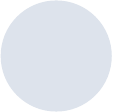} & \includegraphics[scale=0.15]{pics/Introduction/1.pdf} & - & - & - & -  & - & - & - \\

\midrule[1 pt]

Pang \etal~\cite{DBLP:journals/corr/abs-2007-02500} & \includegraphics[scale=0.15]{pics/Introduction/4.pdf} & \includegraphics[scale=0.15]{pics/Introduction/4.pdf} & \includegraphics[scale=0.15]{pics/Introduction/1.pdf} & \includegraphics[scale=0.15]{pics/Introduction/1.pdf} & \includegraphics[scale=0.15]{pics/Introduction/1.pdf} & - & - & \includegraphics[scale=0.15]{pics/Introduction/4.pdf} & \includegraphics[scale=0.15]{pics/Introduction/4.pdf} & - \\

Chalapathy and Chawla~\cite{chalapathy2019deep} & \includegraphics[scale=0.15]{pics/Introduction/4.pdf} & \includegraphics[scale=0.15]{pics/Introduction/4.pdf} & - & - & - & - & -  & \includegraphics[scale=0.15]{pics/Introduction/4.pdf} & \includegraphics[scale=0.15]{pics/Introduction/4.pdf} & - \\

\midrule[1 pt]

Akoglu \etal~\cite{akoglu2015graph} & \includegraphics[scale=0.15]{pics/Introduction/4.pdf} & - & \includegraphics[scale=0.15]{pics/Introduction/4.pdf} & - & - & - & - & - & - & - \\
Ranshous \etal~\cite{ranshous2015anomaly} & \includegraphics[scale=0.15]{pics/Introduction/4.pdf} & - & \includegraphics[scale=0.15]{pics/Introduction/4.pdf} & - & - & -& -  & \includegraphics[scale=0.15]{pics/Introduction/4.pdf} & - & - \\
Jennifer and Kumar~\cite{d2021anomaly} & \includegraphics[scale=0.15]{pics/Introduction/4.pdf} & - & \includegraphics[scale=0.15]{pics/Introduction/4.pdf} & - & - & - & - & - & - & - \\

\midrule[1 pt]

Eltanbouly \etal~\cite{DBLP:conf/iciot3/EltanboulyBACE20} & \includegraphics[scale=0.15]{pics/Introduction/4.pdf} & \includegraphics[scale=0.15]{pics/Introduction/2.pdf} & \includegraphics[scale=0.15]{pics/Introduction/1.pdf} & - & - & - & -  & - & - & - \\
Fernandes \etal~\cite{DBLP:journals/telsys/FernandesRCAP19} & \includegraphics[scale=0.15]{pics/Introduction/4.pdf} & \includegraphics[scale=0.15]{pics/Introduction/2.pdf} & \includegraphics[scale=0.15]{pics/Introduction/4.pdf} & - & - & - & -  & - & - & - \\
Kwon \etal~\cite{kwon2019survey} & \includegraphics[scale=0.15]{pics/Introduction/4.pdf} & \includegraphics[scale=0.15]{pics/Introduction/2.pdf} & - & - & - & -  & - & - & \includegraphics[scale=0.15]{pics/Introduction/1.pdf} & - \\
Gogoi \etal~\cite{DBLP:journals/cj/GogoiBBK11} & \includegraphics[scale=0.15]{pics/Introduction/4.pdf} & \includegraphics[scale=0.15]{pics/Introduction/1.pdf} & - & - & - & - & - & - & - & - \\

\midrule[1 pt]

Savage \etal~\cite{savage2014anomaly} & \includegraphics[scale=0.15]{pics/Introduction/4.pdf} & - & \includegraphics[scale=0.15]{pics/Introduction/4.pdf} & - & - & - & - & - & - & - \\
Yu \etal~\cite{yu2016survey} & \includegraphics[scale=0.15]{pics/Introduction/4.pdf} & - & \includegraphics[scale=0.15]{pics/Introduction/4.pdf} & - & - & - & - & - & - & - \\
Hunkelmann \etal~\cite{ahmed2019combining} & \includegraphics[scale=0.15]{pics/Introduction/4.pdf} & - & \includegraphics[scale=0.15]{pics/Introduction/1.pdf} & - & - & - & - & - & - & - \\
Pourhabibi \etal~\cite{Fraud2020} & \includegraphics[scale=0.15]{pics/Introduction/4.pdf} & \includegraphics[scale=0.15]{pics/Introduction/2.pdf} & \includegraphics[scale=0.15]{pics/Introduction/4.pdf} & \includegraphics[scale=0.15]{pics/Introduction/2.pdf} & - & - & -  & - & - & - \\

\bottomrule[1 pt]
\multicolumn{11}{l}{* AD: Anomaly Detection, DAD: Anomaly Detection with Deep Learning, GAD: Graph Anomaly Detection.} \\
\multicolumn{11}{l}{* GADL: Graph Anomaly Detection with Deep Learning.}\\
\multicolumn{11}{l}{* -: not included, \includegraphics[scale=0.2]{pics/Introduction/1.pdf} (1-2 references included), \includegraphics[scale=0.2]{pics/Introduction/2.pdf} (3-10 references included), \includegraphics[scale=0.2]{pics/Introduction/4.pdf} (10+ references included).}
\end{tabular}
\label{table:comparison}}}
\end{table*}
 
Non-deep learning based techniques also lack the capability to capture the non-linear properties of real objects~\cite{DBLP:journals/corr/abs-2007-02500}.
Hence, the representations of objects learned by them are not expressive enough to fully support graph anomaly detection.
To tackle these problems, more recent studies seek the potential of adopting deep learning techniques to identify anomalous graph objects. 
As a powerful tool for data mining, deep learning has achieved great success in data representation and pattern recognition~\cite{DBLP:conf/wsdm/WangNWYL20,zhang2019unsupervised, DBLP:conf/icdm/WangZ00ZX20}.
Its deep architecture with layers of parameters and transformations appear to suit the aforementioned problems well. The more recent studies, such as deep graph representation learning and graph neural networks (GNNs), further enrich the capability of deep learning for graph data mining~\cite{ijcai2020-693,wu2020comprehensive,cui2018survey,NIPS20202,su2021comprehensive}. 
By extracting expressive representations such that graph anomalies and normal objects can be easily separated, or the deviating patterns of anomalies can be learned directly through deep learning techniques, graph anomaly detection with deep learning (GADL) is starting to take the lead in the forefront of anomaly detection.
As a frontier technology, graph anomaly detection with deep learning, hence, is expected to generate more fruitful results on detecting anomalies and secure a more convenient life for the society.

\subsection{Challenges in GAD with Deep Learning} \label{sec:introdution:challenges}

Due to the complexity of anomaly detection and graph data mining~\cite{noble2003graph,DBLP:conf/ijcai/TengYEL18,shah2016edgecentric,DBLP:conf/icdm/WangGF17,GAL}, in addition to the prior mentioned data-specific challenges, adopting deep learning techniques for graph anomaly detection also faces a number of challenges from the technical side. 
These challenges associated with deep learning are categorized as technique-specific challenges (Tech-CHs), and they are summarized as follows.

\textbf{Tech-CH1. Anomaly-aware training objectives.} Deep learning models rely heavily on the training objectives to fine-tune all the trainable parameters. 
For graph anomaly detection, this necessitates appropriate training objectives or loss functions such that the GADL models can effectively capture the differences between benign and anomalous objects.
Designing anomaly-aware objectives is very challenging because there is no prior knowledge about the ground-truth anomalies as well as their deviating patterns versus the majority.
How to effectively separate anomalies from normal objects through training remains critical for deep learning-based models.

\textbf{Tech-CH2. Anomaly interpretability.} In real-world scenarios, the interpretability of detected anomalies is also vital because we need to provide convincing evidence to support the subsequent anomaly handling process. For example, the risk management department of a financial organization must provide lawful evidence before blocking the accounts of identified anomalous users. 
As deep learning has been limited for its interpretability~\cite{DBLP:journals/jzusc/ZhangZ18,DBLP:journals/corr/abs-2007-02500}, how to justify the detected graph anomalies remains a big challenge for deep learning techniques.

\textbf{Tech-CH3. High training cost.} Although D(G)NNs are capable of digesting rich information (\eg structural information and attributes) in graph data for anomaly detection, these GADL models are more complex than conventional deep neural networks or machine learning methods due to the anomaly-aware training objectives. Such complexity inherently leads to high training costs in both time and computing resources.

\textbf{Tech-CH4. Hyperparameter tuning.} D(G)NNs naturally exhibit a large set of hyperparameters, such as the number of neurons in each neural network layer, the learning rate, the weight decay and the number of training epochs. Their learning performance is significantly affected by the values of these hyperparameters. However, it remains a serious challenge to effectively select the optimal/sub-optimal settings for the detection models due to the lack of labeled data in real scenarios.

Because deep learning models are sensitive to their associated hyperparameters, setting well-performing values for the hyperparameters is vital to the success of a task.
Tuning hyperparameter is relatively trivial in supervised learning when labeled data are available. For instance, users can find an optimal/sub-optimal set of hyperparameters (\eg through random search, grid search) by comparing the model's outputs with the ground-truth. However, unsupervised anomaly detection has no accessible labeled data to judge the model's performance under different hyperparameter settings~\cite{akoglu2021anomaly,zhao2020automating}. Selecting the ideal hyperparameter values for unsupervised detection models persists as a critical obstacle to applying them in a wide range of real scenarios.

\subsection{Existing Anomaly Detection Surveys}

\begin{figure*}[!t]
\setlength{\belowcaptionskip}{-0.25cm}
    \centerline{\includegraphics[width=0.98\textwidth]{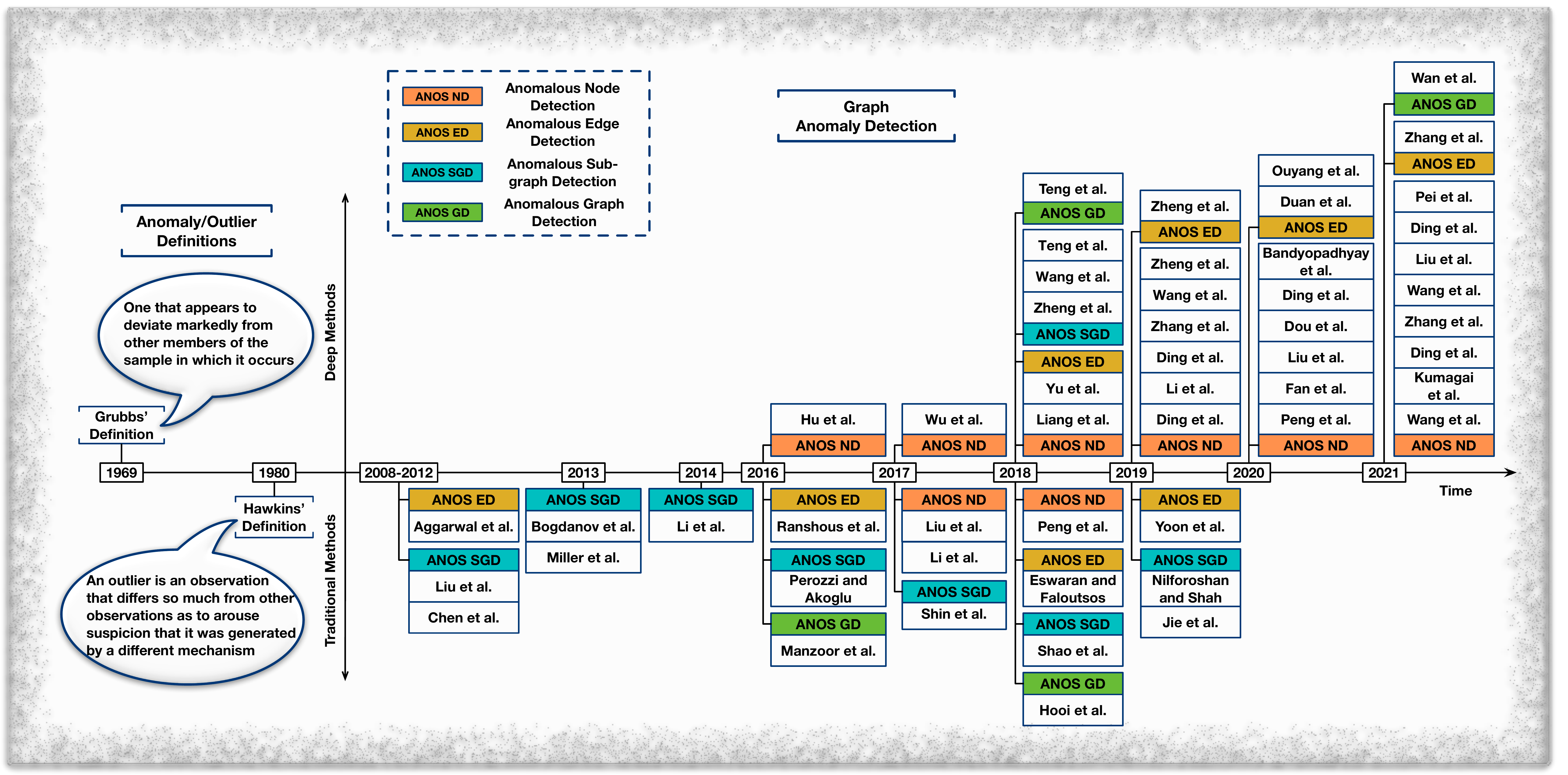}}
    \caption{A Timeline of Graph Anomaly Detection and Reviewed Techniques.}
    \label{pic:timeline}
\end{figure*} 

Recognizing the significance of anomaly detection, many review works have been conducted in the last ten years covering a range of anomaly detection topics: anomaly detection with deep learning, graph anomaly detection, graph anomaly detection with deep learning, and particular applications of graph anomaly detection such as social media, social networks, fraud detection and network security, etc.

There are some representative surveys on generalized anomaly detection techniques -~\cite{chandola2009anomaly},~\cite{DBLP:journals/csur/BoukercheZA20} and~\cite{DBLP:journals/jbd/ThudumuBJS20}.
But only the most up-to-date work in Thudumu \etal~\cite{DBLP:journals/jbd/ThudumuBJS20} covers the topic of graph anomaly detection.
Recognizing the power of deep learning, the three contemporary surveys, Ruff \etal~\cite{ruff2021unifying}, Pang \etal~\cite{DBLP:journals/corr/abs-2007-02500} and Chalapathy and Chawla~\cite{chalapathy2019deep} specifically review deep learning based anomaly detection techniques specifically.

As for graph anomaly detection, Akoglu \etal~\cite{akoglu2015graph}, Ranshous \etal~\cite{ranshous2015anomaly}, and Jennifer and Kumar~\cite{d2021anomaly} put their concentration on graph anomaly detection, reviewing many conventional approaches in this area, including statistical models and machine learning techniques.
Other surveys are dedicated to particular applications of graph anomaly detection, such as computer network intrusion detection and anomaly detection in online social networks, \eg~\cite{yu2016survey,ahmed2019combining,Fraud2020}, and~\cite{DBLP:conf/iciot3/EltanboulyBACE20,DBLP:journals/telsys/FernandesRCAP19,kwon2019survey,DBLP:journals/cj/GogoiBBK11,savage2014anomaly}.
These works provided solid reviews of the application of anomaly detection/graph anomaly detection techniques in these high demand and vital domains. 
However, none of the mentioned surveys are dedicated to techniques on graph anomaly detection with deep learning, as shown in Table~\ref{table:comparison}, and hence do not provide a systematic and comprehensive review of these techniques.

\subsection{Contributions}
Our contributions are summarized as follows:
\begin{itemize}
  \item \textbf{The first survey in graph anomaly detection with deep learning.} To the best of our knowledge, our survey is the first to review the state-of-the-art deep learning techniques for graph anomaly detection. Most of the relevant surveys focus either on conventional graph anomaly detection methods using non-deep learning techniques or on generalized anomaly detection techniques (for tabular/point data, time series, etc.). Until now, there has been no dedicated and comprehensive survey on graph anomaly detection with deep learning. Our work bridges this gap, and we expect that an organized and systematic survey will help push forward research in this area. 

  \item \textbf{A systematic and comprehensive review.} In this survey, we review the most up-to-date deep learning techniques for graph anomaly detection published in influential international conferences and journals in the area of deep learning, data mining, web services, and artificial intelligence, including: TKDE, TKDD, TPAMI, NeurIPS, SIGKDD, ICDM, WSDM, SDM, SIGMOD, IJCAI, AAAI, ICDE, CIKM, ICML, WWW, CVPR, and others. We first summarize seven data-specific and four technique-specific challenges in graph anomaly detection with deep learning. We then comprehensively review existing works from the perspectives of: 1) the motivations behind the deep methods; 2) the main ideas for identifying graph anomalies; 3) a brief introduction to conventional non-deep learning techniques; and 4) the technical details of deep learning algorithms. A brief timeline of graph anomaly detection and reviewed works is given in Fig.~\ref{pic:timeline}.
  
  \item \textbf{Future directions.} From the survey results, we highlight 12 future research directions covering emerging problems introduced by graph data, anomaly detection, deep learning models, and real-world applications. These future opportunities indicate challenges that have not been adequately tackled, and so more effort is needed in the future.

  \item \textbf{Affluent resources.} Our survey also provides an extensive collection of open-sourced anomaly detection algorithms, public datasets, synthetic dataset generating techniques, as well as commonly used evaluation metrics to push forward the state-of-the-art in graph anomaly detection. These published resources offer benchmark datasets and baselines for future research. 

  \item \textbf{A new taxonomy.} We have organized this survey with regard to different types of anomalies (\ie nodes, edges, sub-graphs, and graphs) existing in graphs or graph databases. We also pinpoint the differences and similarities between different types of graph anomalies.

\end{itemize}

The rest of this survey is organized as follows. In Section~\ref{sec:preliminaries}, we provide preliminaries about the different types of settings. 
From Section~\ref{sec:AND} to Section~\ref{sec:anosgd:dynamic}, we review existing techniques for detecting anomalous nodes, edges, sub-graphs and graphs, respectively.
In Section~\ref{sec:resources}, we first provide a collection of published graph anomaly detection algorithms and datasets and then summarize commonly used evaluation metrics and synthetic data generation strategies.
We highlight 12 future directions concerning deep learning in graph anomaly detection in Section~\ref{sec:futures} and summarize our survey in Section~\ref{sec:conclusion}.
A concrete taxonomy of our survey is given in Appendix~\ref{appendix:taxonomy}.
\vspace{-0.1cm}

\section{Preliminaries} \label{sec:preliminaries}
In this section, we provide definitions of different types of graphs mostly used in node/edge/sub-graph-level anomaly detection (Section~\ref{sec:AND} to Section~\ref{sec:subgraph}).
For consistency, we have followed the conventional categorization of graphs as in existing works~\cite{akoglu2015graph,ranshous2015anomaly,kwon2019survey} and categorize them as static graphs, dynamic graphs, and graph databases. Unless otherwise specified, all graphs mentioned in the following sections are static. Meanwhile, as graph-level anomaly detection is discussed far away on page 13, to enhance readability, the definition for the graph database is given closer to the material in Section~\ref{sec:anosgd:db}.

\textbf{\textit{Definition 1 (Plain Graph)}}. A static plain graph $G = \{V, E\}$ comprises a node set $V = \{ v_{i} \}_{1}^{n}$ and an edge set $E = \{e_{i,j}\}$ where $n$ is the number of nodes and $e_{i,j} = (v_i,v_j)$ denotes an edge between nodes $v_{i}$ and $v_{j}$. The adjacency matrix $A = [a_{i,j}]_{n\times n}$ restores the graph structure, where $a_{i,j} = 1$ if node $v_{i}$ and $v_{j}$ is connected, otherwise $a_{i,j} = 0$.

\textbf{\textit{Definition 2 (Attributed Graph)}}. A static attributed graph $G = \{V, E, X\}$ comprises a node set $V$, an edge set $E$ and an attribute set $X$. In an attributed graph, the graph structure follows the definition in Definition 1. The attribute matrix $X = [\mathbf{x}_{i}]_{n\times k}$ consists of nodes' attribute vectors, where $\mathbf{x}_{i}$ is the attribute vector associated with node $v_{i}$ and $k$ is the vector's dimension. Hereafter, the terms attribute and feature are used interchangeably.

\textbf{\textit{Definition 3 (Dynamic Graph)}}. A dynamic graph $G(t) = \{V(t), E(t), X_{v}(t), X_{e}(t) \}$ comprises nodes and edges changing overtime. $V(t)$ is the nodes set in the graph at a specific time step $t$, $E(t)$ is the corresponding edge set, $X_{v}(t)$ and $X_{e}(t)$ are the node attribute matrix and edge attribute matrix at time step $t$ in the graph if existed.

In reality, the nodes or edges might also be associated with numerical or categorical labels to indicate their classes (\eg normal or abnormal). When label information is available/partially-available, supervised/semi-supervised detection models could be effectively trained.

\section{Anomalous node detection (ANOS ND)} \label{sec:AND}

Anomalous nodes are commonly recognized as individual nodes that are significantly different from others.
In real-world applications, these nodes often represent abnormal objects that appear individually, such as a single network intruder in computer networks, an independent fraudulent user in online social networks or a specific fake news on social media.
In this section, we specifically focus on anomalous node detection in static graphs.
The reviews on dynamic graphs can be found in Section~\ref{sec:node:dg}. Table~\ref{table:ANOSND} at the end of Section~\ref{sec:node:dg} provides a summary of techniques reviewed for ANOS ND.

When detecting anomalous nodes in static graphs, the differences between anomalies and regular nodes are mainly drawn from the graph structural information and nodes/edges' attributes~\cite{li2017radar,bojchevski2018bayesian,zhu2020mixedad,perozzi2014focused}.
Given prior knowledge (\ie community structure, attributes) about a static graph, anomalous nodes can be further categorized into the following three types:
\begin{itemize}
    \item \textbf{Global anomalies} only consider the node attributes. They are nodes that have attributes significantly different from all other nodes in the graph. 
    \item \textbf{Structural anomalies} only consider the graph structural information. They are abnormal nodes that have different connection patterns (\eg connecting different communities, forming dense links with others).
    \item \textbf{Community anomalies} consider both node attributes and graph structural information. They are defined as nodes that have different attribute values compared to other nodes in the same community.
\end{itemize}

\begin{figure}[!t]
   \setlength{\belowcaptionskip}{-0.5cm}  
    \centerline{\includegraphics[scale=0.47]{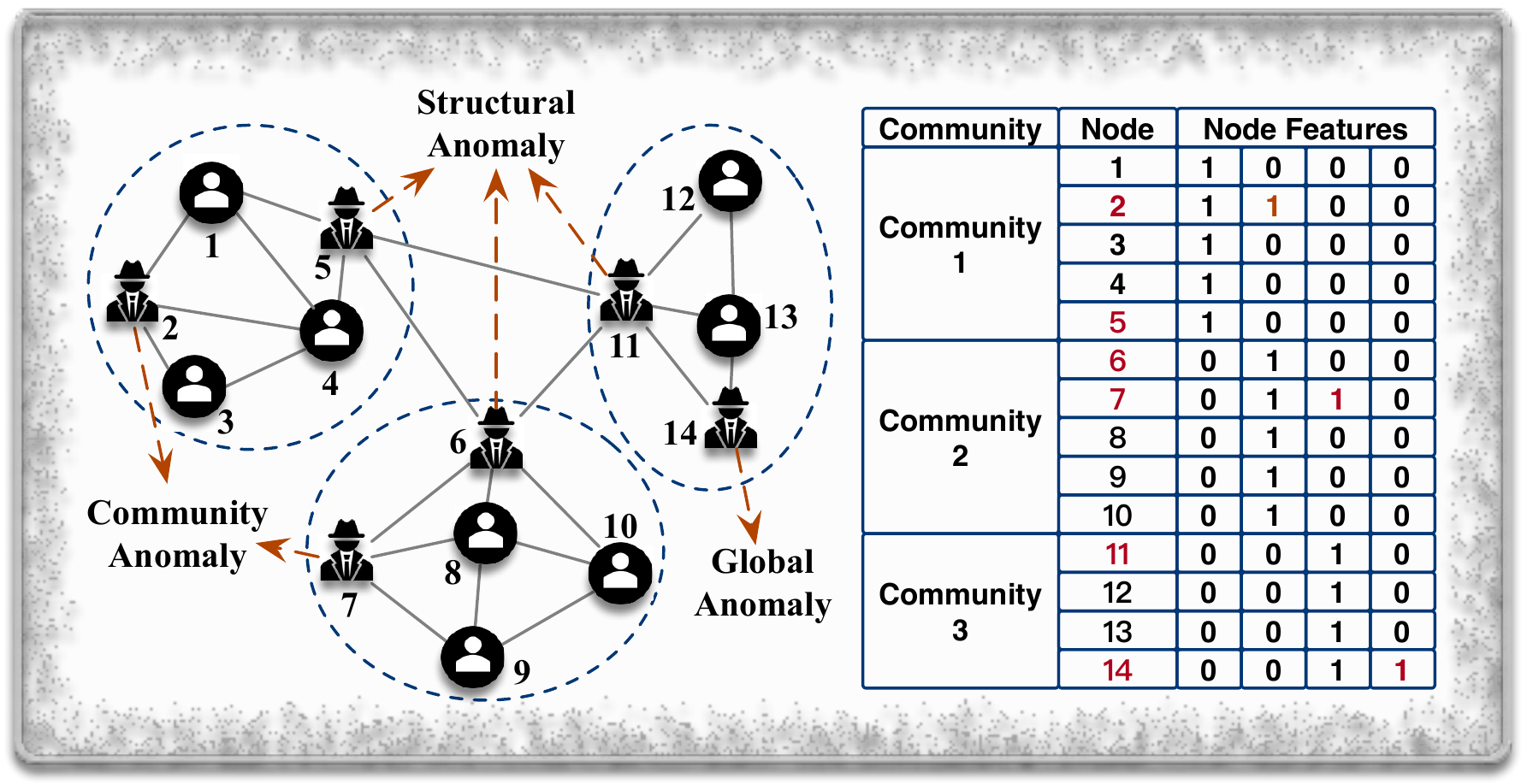}}
    \caption{Three Types of Anomalous Nodes: Structural Anomalies, Community Anomalies and Global Anomalies.}
    \label{pic:node_classes}
\end{figure}

In Fig.~\ref{pic:node_classes}, node 14 is a global anomaly because its 4th feature value is 1 while all other nodes in the graph have the value of 0 for the corresponding feature. Nodes 5, 6, and 11 are identified as structural anomalies because they have links with other communities while other nodes in their community do not form cross-community links. Nodes 2 and 7 are community anomalies because their feature values are different from others in the communities they belong to. 

\subsection{ANOS ND on Plain Graphs} \label{sec:node:spg}

Plain graphs are dedicated to representing the structural information in real-world networks.
To detect anomalous nodes in plain graphs, the graph structure has been extensively exploited from various angles. 
Here, we first summarize the representative traditional non-deep learning approaches, followed by a more recent, advanced detection technique based on representation learning.

\begin{figure*}[!t]
    \setlength{\belowcaptionskip}{-0.25cm}  
    \centerline{\includegraphics[width=0.98\textwidth]{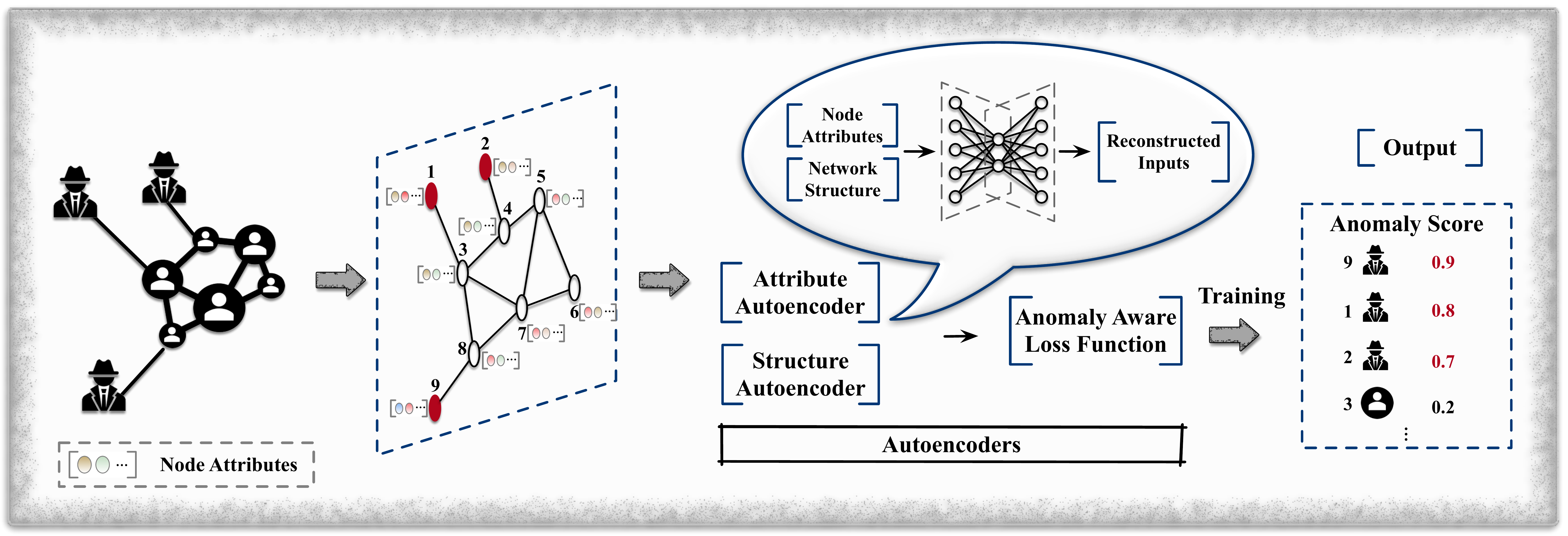}}
    \caption{ANOS ND on attributed graphs -- Deep NN based approaches. As an example, the autoencoder is used to capture the graph structure and node attributes. With a specially-designed anomaly aware loss function, anomaly scores will be assigned to every node, and the top-k nodes are anomalies (\eg nodes 9, 1, and 2 at top-3).}
    \label{pic:conventional_autoencoder_framework}
\end{figure*}

\subsubsection{Traditional Non-Deep Learning Techniques} \label{sec:anosnd:tndp}
Prior to the recent advances in deep learning and other state-of-the-art data mining technologies, traditional non-deep learning techniques have been widely used in many real-world networks to identify anomalous entities.
A key idea behind these techniques was to transform the graph anomaly detection into a traditional anomaly detection problem, because the graph data with rich structure information can not be handled by the traditional detection techniques (for tabular data only) directly. 
To bridge the gap, many approaches~\cite{akoglu2010oddball,DBLP:conf/kdd/DingKBKC12,hooi2016fraudar} used the statistical features associated with each node, such as in/out degree, to detect anomalous nodes. 

For instance, OddBall~\cite{akoglu2010oddball} employs the statistical features (\eg the number of 1-hop neighbors and edges, the total weight of edges) extracted from each node and its 1-hop neighbors to detect particular structural anomalies that: 1) form local structures in shape of near-cliques or stars; 2) have heavy links with neighbors such that the total weight is extremely large; or 3) have a single dominant heavy link with one of the neighbors. 

With properly selected statistical features, anomalous nodes can be identified with respect to their deviating feature patterns.
But, in real scenarios, it is very hard to choose the most suitable features from a large number of candidates, and domain experts can always design new statistics, \eg the maximum/minimum weight of edges.
As a result, these techniques often carry prohibitive cost for assessing the most significant features and do not effectively capture the structural information.

\subsubsection{Network Representation Based Techniques} \label{sec:spg:nr}
To capture more valuable information from the graph structure for anomaly detection, network representation techniques have been widely exploited. 
Typically, these techniques encode the graph structure into an embedded vector space and identify anomalous nodes through further analysis. Hu \etal~\cite{hu2016embedding}, for example, proposed an effective embedding method to detect structural anomalies that are connecting with many communities. 
It first adopts a graph partitioning algorithm (\eg METIS~\cite{DBLP:journals/siamsc/KarypisK98}) to group nodes into $d$ communities ($d$ is a user-specified number). 
Then, the method employs a specially designed embedding procedure to learn node embeddings that could capture the link information between each node and $d$ communities.
Denoting the embedding for node $i$ as $Z_i = \{z_{i}^{1},\cdots,z_{i}^{d}\}$, the procedure initializes each $z_{i}^{c} \in Z_i$ with regard to the membership of node $i$ to community $c$ (if node $i$ belongs to the community, then $z_{i}^{c} = \frac{1}{\sqrt{2}}$; otherwise, 0.) and optimizes node embeddings such that directly linked nodes have similar embeddings and unconnected nodes are dissimilar.

After generating the node embeddings, the link information between node $i$ and $d$ communities is quantified for further anomaly detection analysis.
For a given node $i$, such information is represented as:
\begin{equation} \label{eq:hu:nbi}
  \overline{NB(i)} = (y_{i}^{1},...,y_{i}^{d}) = \mathop{\sum}\limits_{j\in NB(i)}(1- \|Z_{i} - Z_{j}\|)\cdot Z_{j},
\end{equation}
where $NB(i)$ comprises node $i$'s neighbors.
If $i$ has many links with community $c$, then the value in the corresponding dimension $y_{i}^{c}$ will be large.

In the last step, Hu \etal~\cite{hu2016embedding} formulate a scoring function to assign anomalousness scores, calculated as:
\begin{equation} \label{eq:hu:as}
  AScore(i) = \mathop{\sum}\limits_{k=1}^{d}\frac{y_{i}^{k}}{y_{i}^{*}}, y_{i}^{*}=\max\{y_{i}^{1},...,y_{i}^{d}\}.
\end{equation}

As expected, structural anomalies receive higher scores as they connect to different communities.
Indeed, given a predefined threshold, nodes with above-threshold scores are identified as anomalies.

To date, many plain network representation methods such as Deepwalk~\cite{perozzi2014deepwalk}, Node2Vec~\cite{grover2016node2vec} and LINE~\cite{tang2015line} have shown their effectiveness in generating node representations and been used for anomaly detection performance validation~\cite{bandyopadhyay2020outlier,bandyopadhyay2019outlier,yu2018netwalk,cai2020structural}.
By pairing the conventional anomaly detection techniques such as density-based techniques~\cite{breunig2000lof} and distance-based techniques~\cite{aggarwal2001outlier} with node embedding techniques, anomalous nodes can be identified with regard to their distinguishable locations (\ie low-density areas or far away from the majorities) in the embedding space.

\subsubsection{Reinforcement Learning Based Techniques}

The success of reinforcement learning (RL) in tackling real-world decision making problems has attracted substantial interests from the anomaly detection community. Detecting anomalous nodes can be naturally regarded as a problem of deciding which class a node belongs to - anomalous or benign. As a special scenario of the general selective harvesting task, the anomalous node detection problem can be approached by a recent work in~\cite{morales2021selective} that intuitively combines reinforcement learning and network embedding techniques for selective harvesting. The proposed model, NAC, is trained with labeled data without any human intervention. Specifically, it first selects a seed network consisting of partially observed nodes and edges. Then, starting from the seed network, NAC adopts reinforcement learning to learn a node selection plan such that anomalous nodes in the undiscovered area can be identified. This is achieved by rewarding selection plans that can choose labeled anomalies with higher gains. Through offline training, NAC will learn an optimal/suboptimal anomalous node selection strategy and discover potential anomalies in the undiscovered graph step by step.

\subsection{ANOS ND on Attributed Graphs} \label{lb:node:sag}

In addition to the structural information, real-world networks also contain rich attribute information affiliated with nodes~\cite{hamilton2017inductive,hamilton2017representation}. These attributes provide complementary information about real objects and together with graph structure, more hidden anomalies that are non-trivial can now be detected. 

\begin{figure*}[!t]
    \setlength{\belowcaptionskip}{-0.25cm} 
    \centerline{\includegraphics[width=0.98\textwidth]{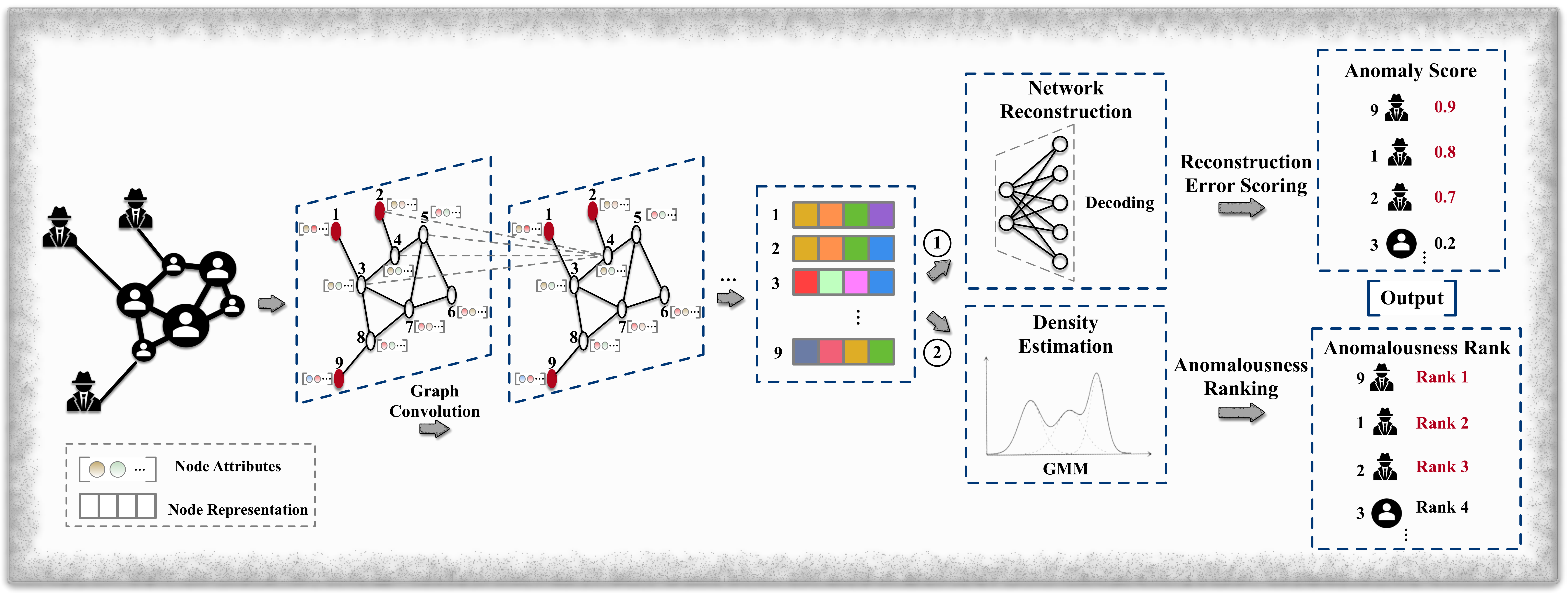}}
    \caption{ANOS ND on attributed graphs -- GCN based approaches. Node representations are generated through GCN layers. Anomalies are then detected according to their reconstruction loss (\ding{172}) or embedding distribution in the embedding space (\ding{173}).}
    \label{pic:GCNbased_framework}
\end{figure*}

For clarity, we distinguish between deep neural networks and graph neural networks in this survey. 
We review deep neural network (Deep NN) based techniques, GCN based techniques, and reinforcement learning based techniques for ANOS ND as follows. 
Due to page limitations, other existing works including traditional non-deep learning techniques, GAT~\cite{velivckovic2017graph} based techniques, GAN based techniques, and network representation based techniques are surveyed in Appendix~\ref{appendix:node:static}.

\subsubsection{Deep NN Based Techniques}

The deep learning models such as autoencoder and deep neural networks provide solid basis for learning data representations. 
Adopting these models for more effective anomalous node detection have drawn substantial interest recently.

For example, Bandyopadhyay \etal~\cite{bandyopadhyay2020outlier} developed an unsupervised deep model, DONE, to detect global anomalies, structural anomalies and community anomalies in attributed graphs.
Specifically, this work measures three anomaly scores for each node that indicate the likelihood of the situations where 1) it has similar attributes with nodes in different communities ($o_i^a$); or 2) it connects with other communities ($o_i^s$); or 3) it belongs to one community structurally but the attributes follow the pattern of another community ($o_i^{com}$).
If a particular node exhibits any of these characteristics, then it is assigned a higher score and is anomalous.

To acquire these scores, DONE adopts two separate autoencoders (AE), \ie a structure AE and an attribute AE, as shown in Fig.~\ref{pic:conventional_autoencoder_framework}. 
Both are trained by minimizing the reconstruction errors and preserving the homophily that assumes connected nodes have similar representations in the graph. 
When training the AEs, nodes exhibiting the predefined characteristics are hard to reconstruct and therefore introduce more reconstruction errors because their structure or attribute patterns do not conform to the standard behavior.
Hence, the adverse impact of anomalies should be alleviated to achieve the minimized error.
Accordingly, DONE specially designs an anomaly-aware loss function with five terms: $\mathcal{L}_{str}^{Recs}$, $\mathcal{L}_{attr}^{Recs}$, $\mathcal{L}_{str}^{Hom}$, $\mathcal{L}_{attr}^{Hom}$, and $\mathcal{L}^{Com}$.
$\mathcal{L}_{str}^{Resc}$ and $\mathcal{L}_{attr}^{Resc}$ are the structure reconstruction error and attribute reconstruction error that can be written as: 
\begin{equation} \label{DONE:ls1}
    \begin{split}
        \mathcal{L}_{str}^{Recs} = \frac{1}{N}\sum_{i=1}^{N}\log(\frac{1}{o_{i}^{s}})\|\mathbf{t}_{i} - \mathbf{\hat{t}}_{i}\|_{2}^{2},
    \end{split}
\end{equation}
and 
\begin{equation} \label{DONE:ls2}
    \begin{split}
        \mathcal{L}_{attr}^{Recs} \frac{1}{N}\sum_{i=1}^{N}\log(\frac{1}{o_{i}^{a}})\|\mathbf{x}_{i} - \mathbf{\hat{x}}_{i}\|_{2}^{2},
    \end{split}
\end{equation}
where $N$ is the number of nodes, $\mathbf{t}_{i}$ and $\mathbf{x}_{i}$ store the structure information and attributes of node $i$, $\mathbf{\hat{t}}_{i}$ and $\mathbf{\hat{x}}_{i}$ are the reconstructed vectors. 
$\mathcal{L}_{str}^{Hom}$ and $\mathcal{L}_{attr}^{Hom}$ are proposed to maintain the homophily and they are formulated as:
\begin{equation} \label{DONE:ls3}
    \begin{split}
        \mathcal{L}_{str}^{Hom} =  \frac{1}{N}\sum_{i=1}^{N}\log(\frac{1}{o_{i}^{s}})\frac{1}{|N(i)|}\sum_{j\in N(i)}||\mathbf{h}_{i}^{s} - \mathbf{h}_{j}^{s}||_{2}^{2},
    \end{split}
\end{equation}
and 
\begin{equation} \label{DONE:ls4}
    \begin{split}
        \mathcal{L}_{attr}^{Hom} =  \frac{1}{N}\sum_{i=1}^{N}\log(\frac{1}{o_{i}^{a}})\frac{1}{|N(i)|}\sum_{j\in N(i)}||\mathbf{h}_{i}^{a} - \mathbf{h}_{j}^{a}||_{2}^{2},
    \end{split}
\end{equation}
where $\mathbf{h}_{i}^{s}$ and $\mathbf{h}_{i}^{a}$ are the learned latent representations from the structure AE and attribute AE, respectively.
$\mathcal{L}^{Com}$ poses further restrictions on the generated representations for each node by the two AEs such that the graph structure and node attributes complement each other. It is formulated as:
\begin{equation} \label{DONE:ls5}
    \begin{split}
        \mathcal{L}^{Com} = \frac{1}{N}\sum_{i=1}^{N}\log(\frac{1}{o_{i}^{com}})||\mathbf{h}_{i}^{s} - \mathbf{h}_{i}^{a}||_{2}^{2},
    \end{split}
\end{equation}
By minimizing the sum of these loss functions, the anomaly scores of each node are quantified, and the top-k nodes with higher scores are identified as anomalies.

\subsubsection{GCN Based Techniques} \label{sec:AND:GCN}

Graph convolutional neural networks (GCNs)~\cite{kipf2016gcn} have accomplished decent success in many graph data mining tasks (\eg link prediction, node classification, and recommendation) owing to its capability of capturing comprehensive information in the graph structure and node attributes.
Therefore, many anomalous node detection techniques start to investigate GCNs. 
Fig.~\ref{pic:GCNbased_framework} illustrates a general framework of existing works in this line.

In~\cite{ding2019deep}, Ding \etal measured an anomaly score for each node using the network reconstruction errors of both the structure and attribute.
The proposed method, DOMINANT, comprises three parts, namely, the graph convolutional encoder, the structure reconstruction decoder, and the attribute reconstruction decoder. The graph convolutional encoder generates node embeddings through multiple graph convolutional layers. The structure reconstruction decoder tends to reconstruct the network structure from the learned node embeddings, while the attribute reconstruction decoder reconstructs the node attribute matrix.
The whole neural network is trained to minimize the following loss function:
\begin{equation} \label{eq:ad}
    \begin{split}
    \mathcal{L}_{DOMINANT} &= (1-\alpha)\mathcal{R}_{S} + \alpha \mathcal{R}_{A}\\
                &= (1-\alpha)||A - \hat{A}||_{F}^{2} + \alpha||X - \hat{X}||_{F}^{2},
    \end{split}
\end{equation}
where $\alpha$ is the coefficient, $A$ depicts the adjacency matrix of the graph, $\mathcal{R}_{S}$ and $\mathcal{R}_{A}$ quantify reconstruction errors with regard to the graph structure and node attributes, respectively.
When the training is finished, an anomaly score is then assigned to each node according to its contribution to the total reconstruction error, which is calculated by:
\begin{equation} \label{eq:nodescore}
  \textit{score}(i) 
      = (1-\alpha)||\mathbf{a}_{i} - \mathbf{\hat{a}}_{i}||_{2} + \alpha||\mathbf{x}_{i} - \mathbf{\hat{x}}_{i}||_{2},
\end{equation}
where $\mathbf{a}_{i}$ and $\mathbf{x}_{i}$ are the structure vector and attribute vector of node $i$, $\mathbf{\hat{a}}_{i}$ and $\mathbf{\hat{x}}_{i}$ are their corresponding reconstructed vectors. 
The nodes are then ranked according to their anomaly scores in descending order, and the top-k nodes are recognized as anomalies.

To enhance the performance of anomalous node detection, later work by Peng \etal~\cite{peng2020deep} further explores node attributes from multiple attributed views to detect anomalies.
The multiple attributed views are employed to describe different perspectives of the objects~\cite{sheng2019multi,6848779,wu2013multi}.
For example, in online social networks, user's demographic information and posted contents are two different attributed views, and they characterize the personal information and social activities, respectively.
The underlying intuition of investigating different views is that anomalies might appear to be normal in one view but abnormal in another view.

For the purpose of capturing these signals, the proposed method, ALARM, applies multiple GCNs to encode information in different views and adopts a weighted aggregation of them to generate node representations.
This model's training strategy is similar to DOMINANT~\cite{ding2019deep} in that it aims to minimize the network reconstruction loss and attribute reconstruction loss and can be formulated as:
\begin{equation} \label{eq:alarmloss}
\begin{split}
\mathcal{L}_{ALARM} = & \sum_{i=1}^n \sum_{j=1}^n - [\gamma A_{ij}\log\hat{A}_{ij} + (1-A_{ij})\log(1-\hat{A}_{ij})] \\
                      & + ||X - \tilde{X}||^{F}_{2},
\end{split}
\end{equation}
where $\gamma$ is coefficient to balance the errors, $A_{ij}$ is the element at coordinate $(i,j)$ in the adjacency matrix $A$, $\hat{A}_{ij}$ is the corresponding element in the reconstructed adjacency matrix $\hat{A}$, $X$ is the original node feature matrix and $\tilde{X}$ is the reconstructed node feature matrix.
Lastly, ALARM adopts the same scoring function as~\cite{ding2019deep}, and nodes with top-k highest scores are anomalous.

Instead of spotting unexpected nodes using their reconstruction errors, Li \etal~\cite{li2019specae} proposed SpecAE to detect global anomalies and community anomalies via a density estimation approach, Gaussian Mixture Model (GMM).
Global anomalies can be identified by only considering the node attributes. 
For community anomalies, the structure and attributes need to be jointly considered because of their distinctive attributes to the neighbors.
Accordingly, SpecAE investigates a graph convolutional encoder to learn node representations and reconstruct the nodal attributes through a deconvolution decoder.
The parameters in the GMM are then estimated using the node representations. 
Due to the deviating attribute patterns of global and community anomalies, normal nodes are expected to exhibit greater energies in GMM, and the k nodes with the lowest probabilities are deemed to be anomalies.

In~\cite{wang2019fdgars}, Wang \etal developed a novel detection model that identify fraudsters using their relations and features. Their proposed method, Fdgars, first models online users' reviews and visited items as their features, and then identifies a small portion of significant fraudsters based on these features. In the last step, a GCN is trained in a semi-supervised manner by using the user-user network, user features, and labeled users. After training, the model can directly label unseen users.

A more recent work, GraphRfi~\cite{GraphRfi}, also explores the potential of combining anomaly detection with other downstream graph analysis tasks.
It targets on leveraging anomaly detection to identify malicious users and provide more accurate recommendations to service benign users by alleviating the impact of these untrustworthy users. 
Specifically, a GCN framework is deployed to encode users and items into a shared embedding space for recommendation and users are classified as fraudsters or normal users through an additional neural random forest using their embeddings. 
For rating prediction between users and items, the framework reduces the corresponding impact of suspicious users by assigning less weights to their training loss. 
At the same time, the rating behavior of users also provides auxiliary information for fraudster detection. 
The mutually beneficial relationship between these two applications (anomaly detection and recommendation) indicates the potential of information sharing among multiple graph learning tasks.

\subsubsection{Reinforcement Learning Based Techniques}

In contrast to NAC, Ding \etal~\cite{ding2019interactive} investigated to the use of reinforcement learning for anomalous node detection in attributed graphs.
Their proposed algorithm, GraphUCB, models both attribute information and structural information, and inherits the merits of the contextual multi-armed bandit technology~\cite{langford2008epoch} to output potential anomalies.
By grouping nodes into $k$ clusters based on their features, GraphUCB forms a $k$-armed bandit model and measures the payoff of selecting a specific node as a potential anomaly for expert evaluation.
With experts' feedback on the predicted anomalies, the decision-making strategy is continuously optimized.
Eventually, the most potential anomalies can be selected. 

\begin{figure*}[!t]
\setlength{\belowcaptionskip}{-0.25cm}
\centerline{\includegraphics[width=0.98\textwidth]{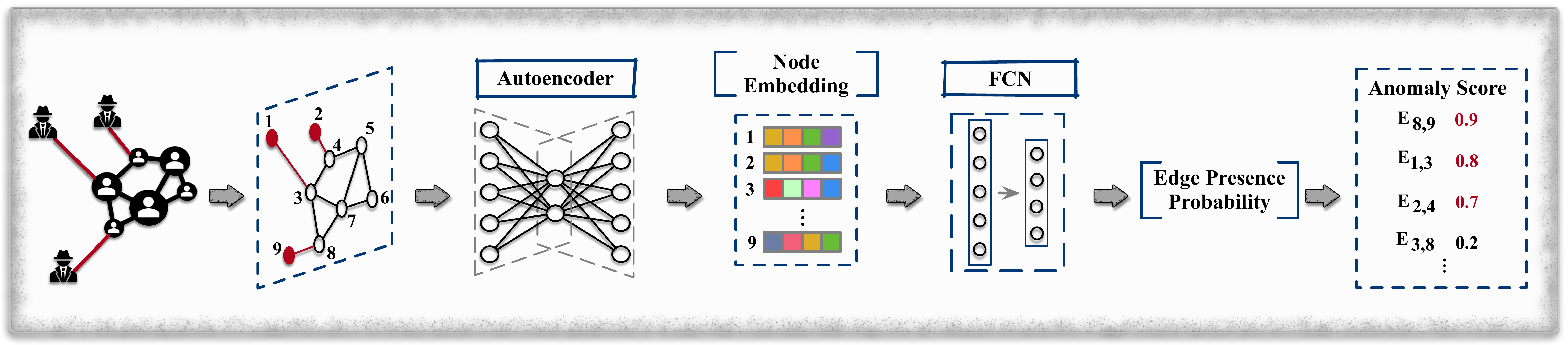}}
\caption{ANOS ED on static graphs -- Deep NN based approaches. For example, the detection technique employs the autoencoder and fully connected network to learn the presence probability of each edge. Anomaly scores are assigned with regard to the probabilities, and the top-k edges are anomalous (\eg $E_{8,9}$, $E_{1,3}$, and $E_{2,4}$ at top-3).}
\label{pic:UGED}
\end{figure*}

\section{ANOS ND on Dynamic Graphs} \label{sec:node:dg}

Real-world networks can be modeled as dynamic graphs to represent evolving objects and the relationships among them.
In addition to structural information and node attributes, dynamic graphs also contain rich temporal signals~\cite{DBLP:conf/wsdm/RossiGNH13}, \eg the evolving patterns of the graph structure and node attributes. 
On the one hand, these information inherently makes anomalous node detection on dynamic graphs more challenging. 
This is because dynamic graphs usually introduce large volume of data and temporal signals should also be captured for anomaly detection.
But, on the other hand, they could provide more details about anomalies~\cite{ranshous2015anomaly,akoglu2015graph,wang2019detecting}. 
In fact, some anomalies might appear to be normal in the graph snapshot at each time stamp, and, only when the changes in a graph's structure are considered, do they become noticeable.

In this section, we review the network representation based techniques and GAN based techniques as follows. Relevant techniques from traditional non-deep learning approaches are reviewed in Appendix~\ref{appendix:node:dynamic}.

\begin{table*}[!t]
\centering
\footnotesize
\renewcommand\arraystretch{1}
\setlength{\tabcolsep}{2.8mm}
\caption{Summary of Anomalous Node Detection Techniques.}
\resizebox{0.96\textwidth}{!}{
\begin{tabular}{m{2.1cm}<{\centering}|m{2.3cm}<{\centering}|m{1.1cm}<{\centering}|m{4cm}<{\centering}|m{1.8cm}<{\centering}|m{2.9cm}<{\centering}}
\toprule[1 pt]
\textbf{Graph Type} & \textbf{Approach} & \textbf{Category} & \textbf{Objective Function} & \textbf{Measurement} & \textbf{Outputs}  \\ 
\midrule[1 pt]
\multirow{3}{*}{\shortstack{Static Graph - \\ Plain}} & ~\cite{hu2016embedding} & NR & $\mathop{\sum}\limits_{(i,j)\in E} \|\mathbf{Z_{i}} - \mathbf{Z_{j}}\|^{2} + \alpha \mathop{\sum}\limits_{(i,j)\notin E}(\|\mathbf{Z_{i}} - \mathbf{Z_{j}}\| -1)^{2}$  & Anomaly Score  & $ \mathop{\sum}\limits_{k=1}^{d}\frac{y_{i}^{k}}{y_{i}^{*}}$ \\ \cline{2-6}
& DCI~\cite{wang2021decoupling} & NR &$\frac{1}{K}\mathop{\sum}\limits_{k=1}^K\mathcal{L}_{DCI}^{k}$ & Anomaly Prediction & Predicted Label \\\cline{2-6}

&NAC~\cite{morales2021selective} & RL & Cumulative reward  & -  & Anomalies \\

\midrule[1 pt]

\multirow{24}{*}{\shortstack{Static Graph - \\ Attributed}} &ALAD~\cite{liu2017accelerated} & Non-DP & $\min\limits_{W,H} \|A-WW^{T}\|_{F}^{2} + \alpha \|X-WH\|_{F}^{2} + \gamma (\|W\|_{F}^{2} + \|H\|_{F}^{2})$  & Anomaly Score & $\frac{\mathbf{W}_{n,c}}{\sum_{c}\mathbf{W}_{n,c}} cos(\mathbf{A}_{n*}, \mathbf{H}_{c*})$ \\   \cline{2-6}

&Radar~\cite{li2017radar} & Non-DP & $\min\limits_{W,R}\|X - W^{T} X - R\|_{F}^{2}+ \alpha \|W\|_{2,1} +\beta \|R\|_{2,1} +\gamma tr(R^{T}LR)$  & Residual Analysis & Residual Value \\  \cline{2-6}

&ANOMALOUS~\cite{peng2018anomalous} & Non-DP & $\min\limits_{W,\tilde{R}}||X - XWX - \tilde{R}||_{F}^{2} + \alpha ||W||_{2,1} + \beta ||W^{T}||_{2,1} +  \gamma ||\tilde{R}^{T}||_{2,1} + \varphi tr(\tilde{R}L\tilde{R}^{T})$ & Residual Analysis & Residual Value   \\  \cline{2-6}

&SGASD~\cite{wu2017adaptive} & Non-DP & $\min\limits_{\mathbf{w,c}}\frac{1}{2}\sum_{i=1}^{m}\mathbf{c_{i}}(V_{i,*}\mathbf{w} - y_i)^2 + \frac{\lambda_{1}}{2}||\mathbf{w}||_{2}^{2} + \lambda_{2} \sum_{i=0}^{d}\sum_{j=1}^{n_{i}}||\mathbf{c}_{G_{j}^{i}}||_{2}$  & Anomaly Prediction & Predicted Label \\  \cline{2-6}

&DONE~\cite{bandyopadhyay2020outlier} & DNN & $\alpha_{1}\mathcal{L}_{str}^{Recs} + \alpha_{2}\mathcal{L}_{attr}^{Recs} + \alpha_{3}\mathcal{L}_{str}^{Hom} + \alpha_{4}\mathcal{L}_{attr}^{Hom} + \alpha_{5}\mathcal{L}^{Com}$  & Anomaly Scores & $o_{i}^{s},o_{i}^{a},o_{i}^{com}$   \\  \cline{2-6}

&DOMINANT~\cite{ding2019deep} & GCN & $(1-\alpha)\mathcal{R}_{S} + \alpha \mathcal{R}_{A}$  & Anomaly Score & $(1-\alpha)||\mathbf{a}_{i} - \mathbf{\hat{a}}_{i}||_{2} + \alpha||\mathbf{x}_{i} - \mathbf{\hat{x}}_{i}||_{2}$ \\  \cline{2-6}

&ALARM~\cite{peng2020deep} & GCN & $\sum_{i=1}^n \sum_{j=1}^n - [\gamma A_{ij}\log\hat{A}_{ij} + (1-A_{ij})\log(1-\hat{A}_{ij})] + \mathcal{L}_a$ & Anomaly Score & $(1-\alpha)||\mathbf{a}_{i} - \mathbf{\hat{a}}_{i}||^2_{2} + \alpha||\mathbf{x}_{i} - \mathbf{\hat{x}}_{i}||^2_{2}$ \\  \cline{2-6}

&SpecAE~\cite{li2019specae} & GCN & $\mathbb{E}[dis(X,\hat{X})] + \mathbb{E}[dis(X,\tilde{X})] + \lambda_{1}\mathbb{E}(E(Z)) + \lambda_{2}KL$ & Density Estimation & Anomalousness Rank  \\  \cline{2-6}

&Fdgars~\cite{wang2019fdgars} & GCN & $\mathcal{L}_{GCN}$ & Anomaly Prediction & Predicted Label  \\  \cline{2-6}

&GraphRfi~\cite{GraphRfi} & GCN & $\mathcal{L}_{rating} + \lambda \mathcal{L}_{fraudster}$ & Anomaly Prediction & Predicted Label  \\  \cline{2-6}

&ResGCN~\cite{pei2021resgcn} & GCN & $(1-\alpha)||A - \hat{A}||^2_{F} + \alpha||X - \hat{X} - \lambda R||^2_{F}$ & Anomaly Score & $||R_{i,:}||_{2}$  \\  \cline{2-6}

&GraphUCB~\cite{ding2019interactive} & RL & Expert Judgment  & - & Anomalies \\  \cline{2-6}

&AnomalyDAE~\cite{fan2020anomalydae} & GAT & $\alpha ||(A - \hat{A})\odot \bm{\theta}||_{F}^{2} + (1-\alpha)||(X - \hat{X})\odot \bm{\eta}||_{F}^{2}$ & Reconstruction Loss & Anomalousness Rank \\  \cline{2-6}

&SemiGNN~\cite{SemiGNN} & GAT & $\alpha \mathcal{L}_{sup} + (1- \alpha) \mathcal{L}_{unsup} + \lambda \mathcal{L}_{reg}$ & Anomaly Prediction & Predicted Label \\  \cline{2-6}

&AEGIS~\cite{ding2020inductive} & GAN & $\mathcal{L}_{AE} + \mathcal{L}_{GAN}$ & Anomaly Score & $1 - D(\mathbf{z}_i)$ \\  \cline{2-6}

&REMAD~\cite{zhang2019robust} & NR & $\mathcal{L}_{res} + \beta\| R^{T}\|_{2,1}$  & Residual Analysis & Residual Value \\  \cline{2-6}

&CARE-GNN~\cite{CARE-GNN} & NR & $\mathcal{L}_{GNN} + \lambda_{1}\mathcal{L}_{Simi}^{(1)} + \lambda_{2}\mathcal{L}_{reg}$  & Anomaly Prediction & Predicted Label \\  \cline{2-6}

&SEANO~\cite{liang2018semi} & NR & $-\sum_{i \in V_{L}} \log p(y_{i}|\mathbf{x}_{i},\Bar{\mathbf{x}}_{N_{i}}) - \sum_{i \in V}\sum_{v\prime \in C_i} \log p(v\prime|\mathbf{x}_{i},\Bar{\mathbf{x}}_{N_{i}})$  & Anomaly Score & Discriminator's Output \\ \cline{2-6}

&OCGNN~\cite{wang2020ocgnn} & NR & $\frac{1}{\beta K}\sum\limits_{v_{i}\in\mathbf{V}_{tr}}[||g(X,A;\mathcal{W})_{v_i}-c||^2 -r^2]^{+}+ r^2 +\frac{\lambda}{2}\sum\limits_{l=1}^{L}||W^{(l)}||^2$  & Location in Embedding Space & Distance to Hypersphere Center \\ \cline{2-6}

&GAL~\cite{GAL} & NR &  $\max\{0,\max\limits_{y_{v^{\prime}} \neq y_{u}}g(u,v^\prime) - \min\limits_{y_{v} = y_{u}}g(u,v) + \Delta_{y_u}\}$  & Anomaly Prediction & Predicted Label \\ \cline{2-6}
 
&CoLA~\cite{liu2021anomaly}& NR & $-\sum\limits_{i=1}^{N}y_{i}\log(CLM(v_i,\mathcal{G}_{i})) + (1-y_i)\log(1-CLM(v_i,\mathcal{G}_{i}))$  & Anomaly Score & $\frac{\sum\limits_{r=1}^{R}(s_{i,r}^{(-)}-s_{i,r}^{(+)})}{R}$ \\ \cline{2-6}

&COMMANDER~\cite{ding2021cross} & NR & $-\mathcal{L}_{D} + \mathcal{L}_{C} + \mathcal{L}_{R}$  & Anomaly Score & $\bar{y_i}||\mathbf{\tilde{x}_i} - \mathbf{x}_i||_2^2$ \\ \cline{2-6}

&FRAUDRE~\cite{zhangge1}& NR & $\sum\limits_{i=1}^{n}f^{*}(y_i,\mathbf{h}_{i}^{(final)}\mathbf{W}_2)$  & Anomaly Prediction & Predicted Label \\ \cline{2-6}

&Meta-GDN~\cite{ding2021few} & NR & $(1-y_i)\cdot|dev(v_i)| + y_i\cdot\max(0,dev(v_i))$  & Anomaly Score & $\mathbf{u}_{s}^{T}\mathbf{o}_{i} + b_s$ \\ 

\midrule[1 pt]

Dynamic Graph - Plain & NetWalk~\cite{yu2018netwalk} & DNN & $\gamma\mathcal{L}_{AE} + \mathcal{L}_{Clique} + \lambda\|W\|_{F}^{2} + \beta KL$ & Anomaly Score & Nearest Distance to Cluster Centers \\ [2ex]
\midrule[1 pt]

\multirow{2}{*}{\shortstack{Dynamic Graph - \\ Attributed}} &MTHL~\cite{teng2017anomaly} & Non-DP & $\mathop{min}_{\mathcal{P}}f(\mathcal{P})$  & Anomaly Score & Distance to Hypersphere Centroid \\  \cline{2-6}
&OCAN~\cite{zheng2019one} & GAN & $\mathcal{L}_{LSTM-AE} + \mathcal{L}_{GAN}$  & Anomaly Score & Discriminator's Output  \\

\bottomrule[1 pt]
\multicolumn{6}{l}{* Non-DP: Non-Deep Learning Techniques, DNN: Deep NN Based Techniques, GCN: GCN Based Techniques, RL: Reinforcement Learning Based Techniques.} \\
\multicolumn{6}{l}{* GAT: GAT Based Techniques, NR: Network Representation Based Techniques, GAN: Generative Adversarial Network Based Techniques.}
\end{tabular}
}
\label{table:ANOSND}
\end{table*}

\begin{figure*}[!t]
\setlength{\belowcaptionskip}{-0.25cm} 
\centerline{\includegraphics[width=0.98\textwidth]{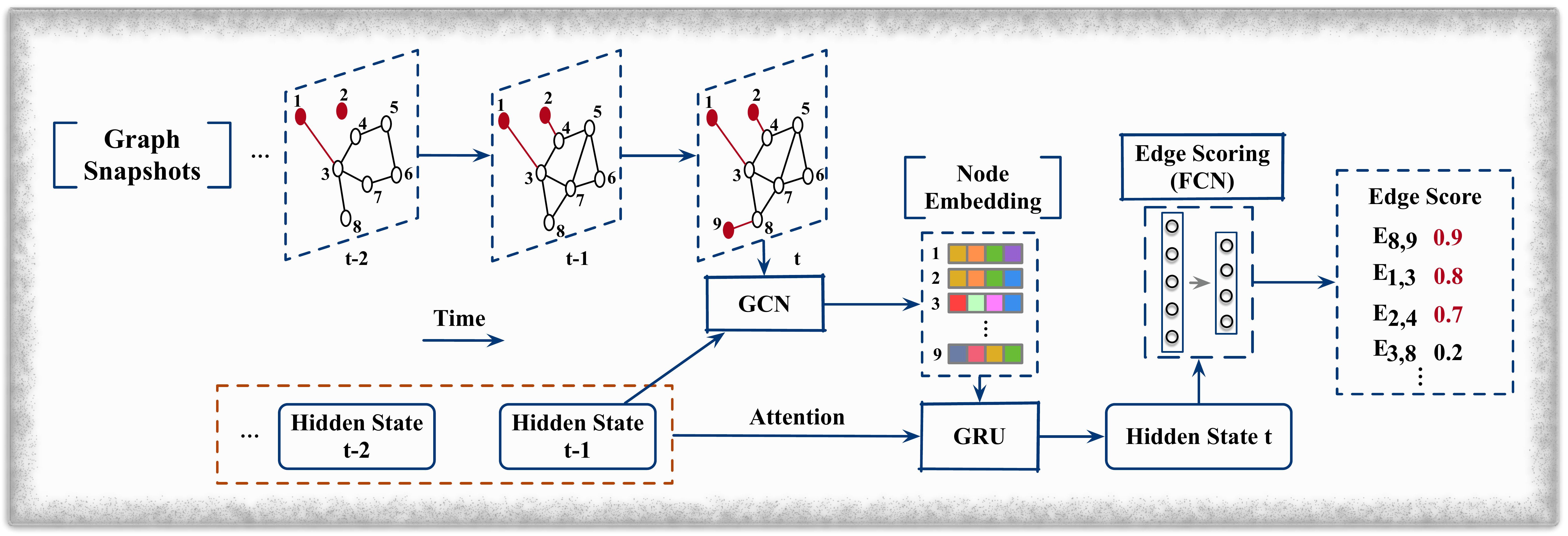}}
\caption{ANOS ED on dynamic graphs -- GCN based approaches. GCN is employed to learn node embeddings from the temporal graph at each timestamp. The attention-based GRU generates the current hidden state using the node embeddings and previous hidden states. The edge scoring function, such as a FCN, is learned to assign anomaly scores, and the top-k edges are depicted as anomalies.}
\label{pic:Dynamic_Edge_Comparison}
\end{figure*}

\subsection{Network Representation Based Techniques} \label{sec:ND:NR}

Following the research line of encoding a graph into an embedding space, after which anomaly detection is performed, dynamic network representation techniques have been investigated in the more recent works.
Specifically, in~\cite{yu2018netwalk}, Yu \etal presented a flexible deep representation technique, called NetWalk, for detecting anomalous nodes in dynamic (plain) graphs using only the structure information.
It adopts an autoencoder to learn node representations on the initial graph and incrementally updates them when new edges are added or existing edges are deleted.
To detect anomalies, NetWalk first executes the streaming $k$-means clustering algorithm~\cite{ailon2009streaming} to group existing nodes in the current time stamp into different clusters. 
Then, each node's anomaly score is measured with regard to its closest distance to the $k$ clusters. 
When the node representations are updated, the cluster centers and anomaly scores are recalculated accordingly.

\subsection{GAN Based Techniques} \label{sec:ND:GAN}
In practice, anomaly detection is facing great challenges from the shortage of ground-truth anomalies.
Consequently, many research efforts have been invested in modeling the features of anomalies or regular objects such that anomalies can be identified effectively.
Among these techniques, generative adversarial networks (GAN)~\cite{goodfellow2014generative} have received extensive attention because of its impressive performance in capturing real data distribution and generating simulated data.

Motivated by the recent advances in ``bad" GAN~\cite{dai2017good}, Zheng \etal~\cite{zheng2019one} circumvented the fraudster detection problem using only the observed benign users' attributes.
The basic idea is to seize the normal activity patterns and detect anomalies that behave significantly differently.
The proposed method, OCAN, starts by extracting the benign users' content features using their historical social behaviors (\eg historical posts, posts' URL), for which this method is classified into the dynamic category.
A long short-term memory (LSTM) based autoencoder~\cite{DBLP:conf/icml/SrivastavaMS15} is employed to achieve this and as assumed, benign users and malicious users are in separate regions in the feature space.
Next, a novel one-class adversarial net comprising a generator and a discriminator is trained.
Specifically, the generator produces complementary data points that locate in the relatively low density areas of benign users.
The discriminator, accordingly, aims to distinguish the generated samples from the benign users.  
After training, benign users' regions are learned by the discriminator and anomalies can hence be identified with regard to their locations.

Both NetWalk~\cite{yu2018netwalk} and OCAN~\cite{zheng2019one} approach the anomalous node detection problem promisingly, however, they respectively only consider the structure or attributes.
By the success of static graph anomaly detection techniques that analyze both aspects, when the structure and attribute information in dynamic graphs are jointly considered, an enhanced detection performance can be foreseen.
We therefore highlight this unexplored area for future works in Section~\ref{sec:futures}.

\section{Anomalous edge detection (ANOS ED) } \label{sec:edge:static}

In contrast to anomalous node detection, which targets individual nodes, ANOS ED aims to identify abnormal links.
These links often inform the unexpected or unusual relationships between real objects~\cite{chang2021f}, such as the abnormal interactions between fraudsters and benign users shown in Fig.~\ref{Toy}, or suspicious interactions between attacker nodes and benign user machines in computer networks. Following the previous taxonomy, in this section, we review the state-of-the-art ANOS ED methods for static graphs, and Section~\ref{sec:edge:dynamic} summarizes the techniques for dynamic graphs. A summary is provided in Table~\ref{tb:edgeandsubgraph}. This section includes methods based on deep NNs, GCNs and network representations. The non-deep learning techniques are reviewed in Appendix~\ref{appendix:edge}.

\subsection{Deep NN Based Techniques}

Similar to deep NN based ANOS ND techniques, autoencoder and fully connected network (FCN) have also been used for anomalous edge detection.
As an example, Ouyang \etal~\cite{DBLP:conf/ijcnn/Ouyang0020} approached the problem by modeling the distribution of edges through deep models to identify the existing edges that are least likely to appear as anomalies (as shown in Fig.~\ref{pic:UGED}). The probability of each edge $e_{u,v}$ is decided by $P(v|u,N(u))$ and $P(u|v,N(v))$ which measure the edge probability using node $u$ with its neighbors $N(u)$ and node $v$ with its neighbors $N(v)$, respectively.
To calculate $P(v|u,N(u))$, the proposed method, UGED, first encodes each node into a lower-dimensional vector through a FCN layer and generates node $u$'s representation by a mean aggregation of itself and its neighbors' vectors.
Next, the node representations are fed into another FCN to estimate $P(v|u,N(u))$.
The prediction is expressed as $\hat{P}(v|u,N(u)) = \text{Softmax}(W\cdot H(u))|_v$, where $W$ represents the trainable parameters, and $H(u)$ is $u$'s representation.
UGED's training scheme aims to maximize the prediction of existing edges via a cross-entropy-based loss function, $\text{CE}(\hat{P}(v|u,N(u)), v)$.
After training, an anomaly score is assigned to each edge using the average of $1-P(v|u,N(u))$ and $1-P(u|v,N(v))$. 
As such, existing edges that have a lower probability will get higher scores and the top-k edges are reported as anomalous.

\subsection{GCN Based Techniques}

Following the line of modeling edge distributions, some studies leverage GCNs to better capture the graph structure information.
Duan \etal~\cite{AANE} demonstrated that the existence of anomalous edges in the training data prevents traditional GCN based models from capturing real edge distributions, which leads to sub-optimal detection performance.
This inherently raises a problem: to achieve better detection performance, the node embedding process should alleviate the negative impact of anomalous edges, but these edges are detected using the learned embeddings.
To tackle this, the proposed method, AANE, jointly considers these two issues by iteratively updating the embeddings and detection results during training.

In each training iteration, AANE generates node embeddings $Z$ through GCN layers and learns an indicator matrix $I$ to spot potential anomalous edges.
Given an input graph $G$ with adjacency matrix $A$, each term $I_{uv}$ in $I$ is 1 if $\hat{A}_{uv} < \text{mean}_{v^{\prime}\in N_{u}} \hat{A}_{uv^{\prime}} - \mu\cdot\text{std}_{v^{\prime}\in N_{u}} \hat{A}_{uv^{\prime}}$, and 0 otherwise.
Here, $\hat{A}_{uv}$ is the predicted link probability between nodes $u$ and $v$, which is calculated as the hyperbolic tangent of $u$ and $v$'s embeddings, and $\mu$ is a predefined threshold.
By this, an edge $uv$ is identified as anomalous when its predicted probability is less than the average of all links associated with the node $u$ by a predefined threshold.

The total loss function of AANE contains two parts: an anomaly-aware loss ($\mathcal{L}_{aal}$) and an adjusted fitting loss ($\mathcal{L}_{afl}$). 
$\mathcal{L}_{aal}$ is proposed to penalize the link prediction results and the indicator matrix $I$ such that anomalous edges will have lower prediction probabilities when they are marked as 1 in $I$. This is formulated as:
\begin{equation}
    \mathcal{L}_{aal} = \sqrt{\sum_{u\in V} \sum_{v\in N(u)}((1-\hat{A}^2_{uv})(1-I_{uv}) + \hat{A}^2_{uv}I_{uv})},
\end{equation}
where $V$ is the node set, $N(u)$ is the set of $u$'s neighbors.
$\mathcal{L}_{afl}$ quantifies the reconstruction loss with regard to the removal of potential anomalous edges, denoted as:
\begin{equation}
    \mathcal{L}_{afl} = \| B - \hat{A}\|^{2}_{2},
\end{equation}
where $B$ is an adjusted adjacency matrix that removes all predicted anomalies from the input adjacency matrix $A$.
By minimizing these two losses, AANE identifies the top-k edges with lowest probabilities as anomalies.

\subsection{Network Representation Based Techniques}
Instead of using node embeddings for ANOS ED, edge representations learned directly from the graph are also feasible for distinguishing anomalies.
If the edge representations well-preserve the graph structure and interaction content (\eg messages in online social networks, co-authored papers in citation networks) between pairs of nodes, an enhanced detection performance can then be expected.
To date, several studies, such as Xu \etal~\cite{DBLP:journals/ijdsa/XuWCY20}, have shown promising results in generating edge representations.
Although they are not specifically designed for graph anomaly detection, they pinpoint a potential approach to ANOS ED. This is highlighted as a potential future direction in Section~\ref{sec:future:ED}.

\section{ANOS ED on Dynamic Graphs} \label{sec:edge:dynamic}

\begin{figure*}[!t]
\setlength{\belowcaptionskip}{-0.25cm}
\centerline{\includegraphics[width=0.98\textwidth]{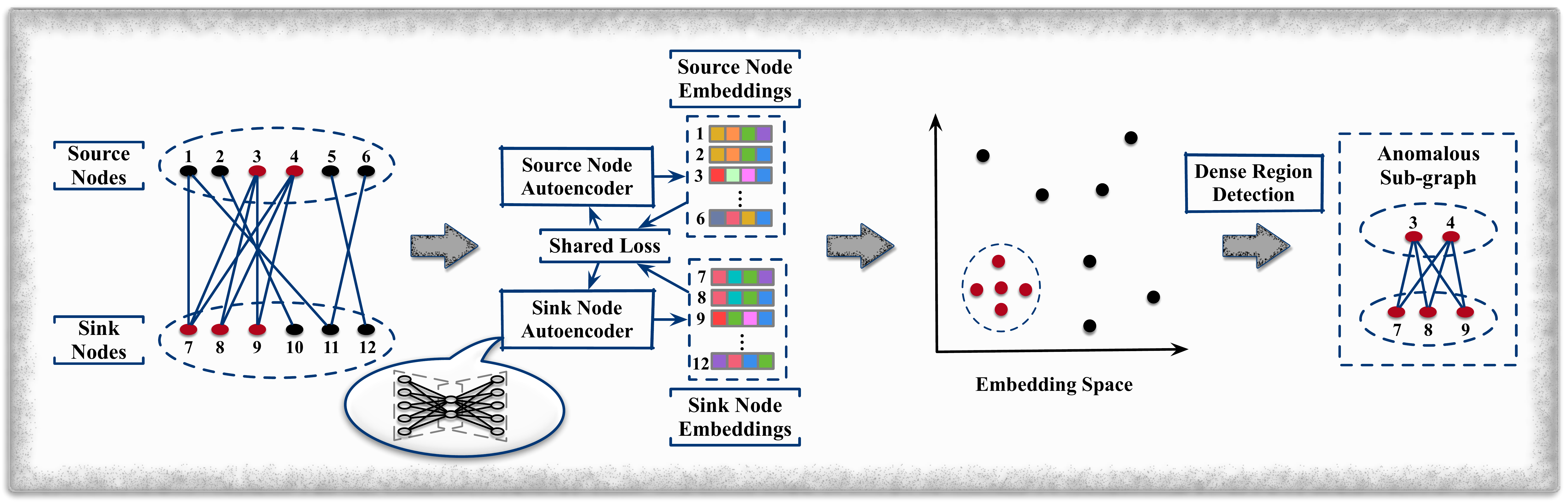}}
\caption{ANOS SGD. Real-world networks are usually represented as bipartite graphs for reflecting the interactions between two different types of objects. To detect ANOS SGD, source nodes and sink nodes are embedded using two autoencoders (linked by a shared loss function), respectively. Anomalous sub-graphs are identified by applying dense region detection algorithms in the embedding space.}
\label{pic:Staic_SG}
\end{figure*}

Dynamic graphs are powerful in reflecting the appearance/disappearance of edges over time~\cite{ranshous2016scalable}.
Anomalous edges can be distinguished by modeling the changes in graph structure and capturing the edge distributions at each time step.
Recent approaches to ANOS ED on dynamic graphs are reviewed in this section.

\subsection{Network Representation Based Techniques}

The intuition of network representation based techniques is to encode the dynamic graph structure information into edge representations and apply the aforementioned traditional anomaly detection techniques to spot irregular edges.
This is quite straightforward, but there remain vital challenges in generating/updating informative edge representations when the graph structure evolves.
To mitigate this challenge, the ANOS ND model NetWalk~\cite{yu2018netwalk} is also capable of detecting anomalous edges in dynamic graphs. 
Following the line of distance-based anomaly detection, NetWalk encodes edges into a shared latent space using node embeddings, and anomalies are identified based on their distances to the nearest edge-cluster centers in the latent space.
Practically, Netwalk generates edge representations as the Hadamard product of the source and destination nodes' representations, denoted as: $\mathbf{z}_{u,v} = \mathbf{z}_u \odot \mathbf{z}_v$.
When new edges arrive or existing edges disappear, the node and edge representations are updated from random walks in the temporary graphs at each time stamp, after which the edge-cluster centers and edge anomaly scores are recalculated. 
Finally, the top-k farthest edges to the edge-clusters are reported as anomalies.

\subsection{GCN Based Techniques}

Although NetWalk is capable of detecting anomalies in dynamic graphs, it simply updates edge representations without considering the evolving patterns of long/short-term nodes and the graph's structure.
For more effective ANOS ED, Zheng \etal~\cite{zheng2019addgraph} intuitively combined temporal, structural and attribute information to measure the anomalousness of edges in dynamic graphs. 
They propose a semi-supervised model, AddGraph, which comprises a GCN and Gated Recurrent Units (GRU) with attention~\cite{cui2019hierarchical} to capture more representative structural information from the temporal graph in each time stamp and dependencies between them, respectively.

At each time stamp $t$, GCN takes the output hidden state ($H^{t-1}$) at time $t-1$ to generate node embeddings, after which the GRU learns the current hidden state $H^{t}$ from the node embeddings and attentions on previous hidden states (as shown in Fig.~\ref{pic:Dynamic_Edge_Comparison}).
After getting the hidden state $H^{t}$ of all nodes, AddGraph assigns an anomaly score to each edge in the temporal graph based on the nodes associated with it. The proposed anomaly scoring function is formulated as:
\begin{equation}\label{eq:AddGraph:scoring}
    f(u,v,w) = w\cdot\sigma(\beta\cdot (||\mathbf{a}\odot \mathbf{h}_{u} + \mathbf{b}\odot \mathbf{h}_{v}||^{2}_{2} - \mu)),
\end{equation}
where $u$ and $v$ are the corresponding nodes, $w$ is the weight of the edge, $\mathbf{a}$ and $\mathbf{b}$ are trainable parameters, $\beta$ and $\mu$ are hyper-parameters, and $\sigma(\cdot)$ is the non-linear activation function.
To learn $\mathbf{a}$ and $\mathbf{b}$, Zheng \etal further assumed that all existing edges in the dynamic graph are normal in the training stage, and sampled non-existing edges as anomalies. Specifically, they form the loss function as:
\begin{equation}\label{eq:AddGraph:loss}
  \begin{split}
      \mathcal{L}_{AddGraph} = &\min\sum_{{(u,v,w)\in \varepsilon^{t}}}\sum_{{(u^{\prime},v^{\prime},w)\notin \varepsilon^{t}}} \\
      & \max\{0, \gamma + f(u,v,w) - f(u^{\prime},v^{\prime},w)\} + \lambda\mathcal{L}_{reg},
  \end{split}
\end{equation}
where $\varepsilon^{t}$ is the edge set, $(u^{\prime},v^{\prime})$ are sampled non-existing edges at time stamp $t$, $\lambda$ is a hyper-parameter, and $\mathcal{L}_{reg}$ regularizes all trainable parameters in the model. 
After training, the scoring function identifies anomalous edges in the test data by assigning higher anomaly scores to them based on Eq.~\ref{eq:AddGraph:scoring}.

\section{Anomalous sub-graph detection (ANOS SGD)} \label{sec:subgraph}

In real life, anomalies might also collude and behave collectively with others to garner benefits.
For instance, fraudulent user groups in an online review network, as shown in Fig.~\ref{Toy}, may post misleading reviews to promote or besmirch certain merchandise. When these data are represented as graphs, anomalies and their interactions usually form suspicious sub-graphs, and ANOS SGD is proposed to distinguish them from the benign.

\begin{figure*}[!t]
\setlength{\belowcaptionskip}{-0.25cm}
\centerline{\includegraphics[width=0.98\textwidth]{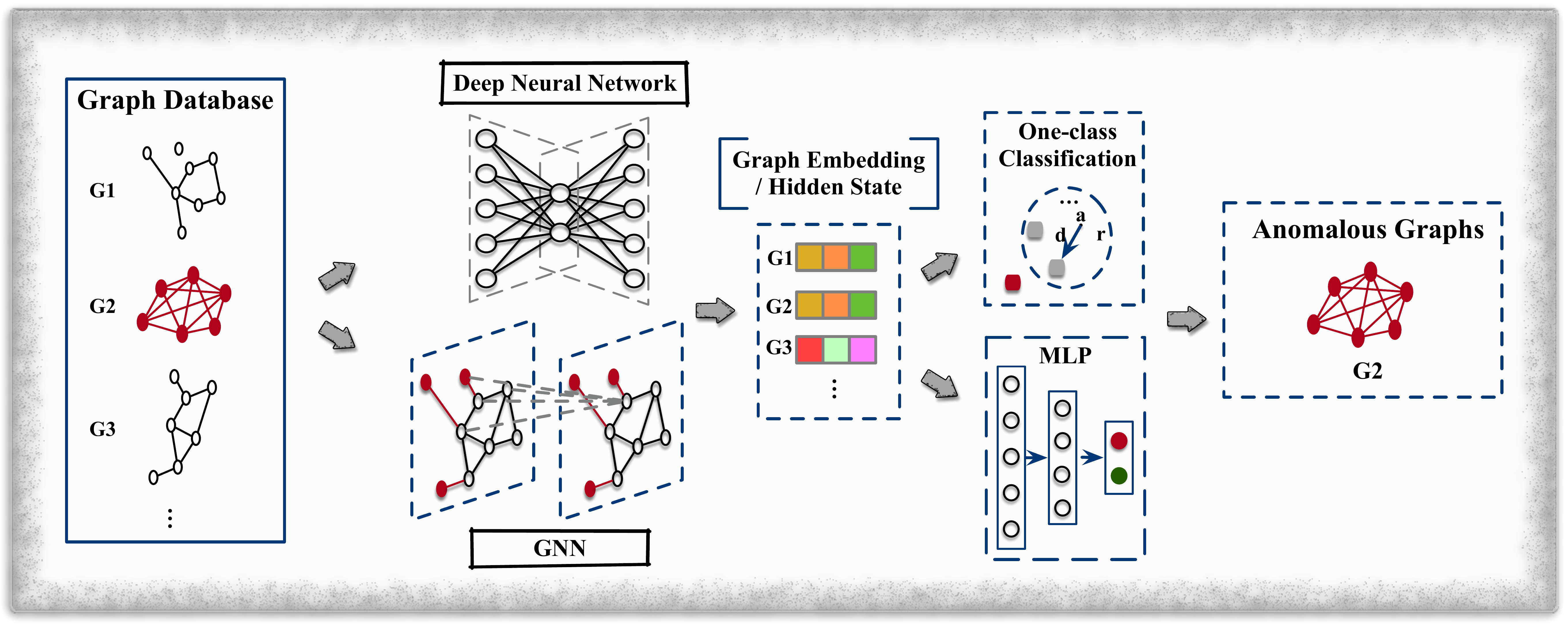}}
\caption{ANOS GD on Graph Database. Generally, the graph-level anomaly detection techniques take a graph database as input. By generating embeddings/hidden state for each single graph in the database through D(G)NNs, anomalous graphs can be depicted by One-class classifiers or MLP.}
\label{pic:graph_level}
\end{figure*}

Unlike individual and independent graph anomalies, \ie single nodes or edges, each node and edge in a suspicious sub-graph might be normal.
However, when considered as a collection, they turn out to be anomalous.
Moreover, these sub-graphs also vary in size and inner structure, making anomalous sub-graph detection more challenging than ANOS ND/ED~\cite{dGraphScan}.
Although extensive effort has been placed on circumventing this problem, deep-learning techniques have only begun to address this problem in the last five years.
For reference, traditional non-deep learning based techniques are briefly introduced in Appendix~\ref{appendix:subgraph}, and a summary of techniques reviewed for ANOS SGD is provided in Table~\ref{tb:edgeandsubgraph} at the end of Section~\ref{sec:anosgd:dynamic}.

Due to the flexibility of heterogeneous graphs in representing the complex relationships between different kinds of real objects, several recent works have taken advantage of deep network representation techniques to detect real-world anomalies through ANOS SGD.
For instance, Wang \etal~\cite{wang2018deep} represented online shopping networks as bipartite graphs (a specific type of heterogeneous graph that has two types of nodes and one type of edge), in which users are source nodes and items are sink nodes.
Fraudulent groups are then detected based on suspicious dense blocks that form in these graphs.

Wang \etal~\cite{wang2018deep} aimed to learn anomaly-aware representations of users such that suspicious users in the same group will be located closely in the vector space, while benign users will be far away (as shown in the embedding space in Fig.~\ref{pic:Staic_SG}). 
According to the observation that user nodes belonging to one fraudulent group are more likely to connect with the same item nodes, the developed model, DeepFD, measures similarities in the behavior of two users, $sim_{ij}$, as the percentage of items shared among all the items they have reviewed.
User representations are then generated through a traditional autoencoder, which is trained using three losses and follows the encoding-decoding process.
The first loss is the reconstruction loss $\mathcal{L}_{res}$ that ensures the bipartite graph structure can be reconstructed properly using the learned user representations and item representations.
The second term $\mathcal{L}_{sim}$ preserves the user similarity information in the learned user representations. That is, if two users have similar behaviors, their representations should also be similar. This loss is formulated as:
\begin{equation}\label{eq:DeepFD:simloss}
    \mathcal{L}_{sim} = \sum_{i,j=1}^{m} sim_{ij} \cdot \| \widehat{sim}_{ij} - sim_{ij} \|_{2}^{2},
\end{equation}
where $m$ is the number of user nodes, $\widehat{sim}_{ij}$ measures the similarity of user $i$ and $j$'s representations using an RBF kernel or other alternative.
The third loss $\mathcal{L}_{reg}$ regularizes all trainable parameters.
Finally, the suspicious dense blocks, which are expected to form dense regions in the vector space, are detected using DBSCAN~\cite{DBSCAN}.

Another work, FraudNE~\cite{FraudNE}, also models online review networks as bipartite graphs and further detects both malicious users and associated manipulated items following the dense block detection principle.
Unlike DeepFD, FraudNE aspires to encode both types of nodes into a shared latent space where suspicious users and items belonging to the same dense block are very close to each other while others distribute uniformly (as shown in Fig.~\ref{pic:Staic_SG}).
FraudNE adopts two traditional autoencoders, namely, a source node autoencoder and a sink node autoencoder, to learn user representations and item representations, respectively.
Both autoencoders are trained to jointly minimize their corresponding reconstruction losses and a shared loss function, and the total loss can be formulated as:
\begin{equation}
    \mathcal{L}_{FraudNE} = \mathcal{L}_{res}^{source} + \mathcal{L}_{res}^{sink} + \alpha \mathcal{L}_{share} + \eta \mathcal{L}_{reg},
\end{equation}
where $\alpha$ and $\eta$ are hyperparameters, and $\mathcal{L}_{reg}$ regularizes all trainable parameters.
Specifically, the reconstruction losses (\ie $\mathcal{L}_{res}^{source}$ and $\mathcal{L}_{res}^{sink}$) measure the gap between the input user/item features (extracted from the graph structure) and their decoded features. 
The shared loss function is proposed to restrict the representation learning process such that each linked pair of users and items get similar representations.
As the DBSCAN~\cite{DBSCAN} algorithm is convenient to apply for dense region detection, FraudNE also uses it to distinguish the dense sub-graphs formed by suspicious users and items.

To date, only a few works have put their efforts into using deep learning techniques for ANOS SGD. However, with intensifying research interest in sub-graph representation learning, we encourage more studies on ANOS SGD and highlight this as a potential future in Section~\ref{sec:future:ED}.

\begin{figure*}[!t]
\setlength{\belowcaptionskip}{-0.25cm}
    \centerline{\includegraphics[width=0.98\textwidth]{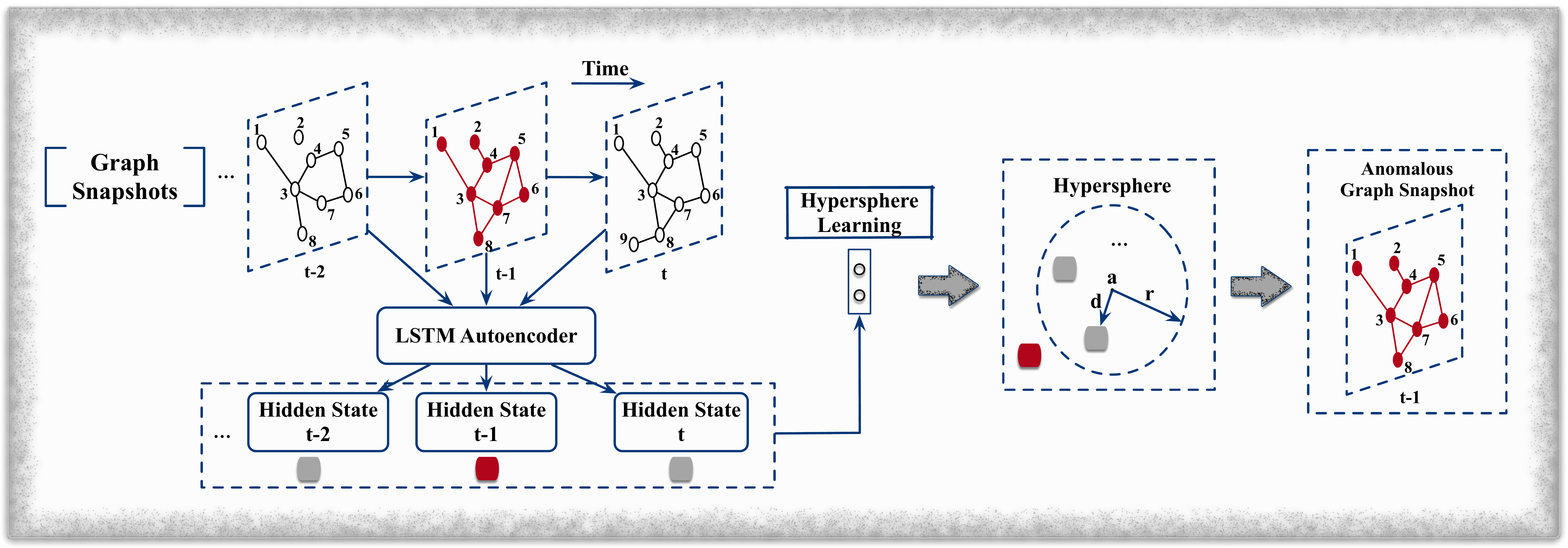}}
    \caption{ANOS GD on dynamic graphs. For each graph snapshot in the dynamic graph, the LSTM autoencoder generates its hidden state using its adjacency matrix and previous hidden states. Through hypersphere learning, a hypersphere with centroid $a$ and radius $r$ is learned such that anomalous snapshots lay outside.}
    \label{pic:GDdynamic}
\end{figure*}

\section{Anomalous graph detection (ANOS GD)} \label{sec:anosgd:db}

Beyond anomalous node, edge, and sub-graph, graph anomalies might also appear as abnormal graphs in a set/database of graphs.
Typically, a graph database is defined as:

\textbf{\textit{Definition 4 (Graph Database)}}. A graph database $\mathcal{G} = \{ G_i = (V_i,E_i,X_v(i), X_e(i)) \}^N_{i=1}$ contains $N$ individual graphs. Here, each graph $G_i$ is comprised of a node set $V_i$ and an edge set $E_i$. $X_v(i)$ and $X_e(i)$ are the node attribute matrix and edge attribute matrix of $G_i$ if it is an attributed graph.

This graph-level ANOS GD aims to detect individual graphs that deviate significantly from the others.
A concrete example of ANOS GD is unusual molecule detection.
When chemical compounds are represented as molecular/chemical graphs where the atoms and bonds are represented as nodes and edges~\cite{6702420,sun2021sugar}, unusual molecules can be identified because their corresponding graphs have structures and/or features that deviate from the others.
Brain disorders detection is another example.
A brain disorder can be diagnosed by analyzing the dynamics of brain graphs at different stages of aging in sequence and finding an inconsistent snapshot at a specific time stamp.

The prior reviewed techniques (\ie ANOS ND/ED/SGD) are not compatible with ANOS GD because they are dedicated to detecting anomalies in a single graph, whereas ANOS GD is directed at detecting graph-level anomalies.
This problem is commonly approached by: 1) measuring the pairwise proximities of graphs using graph kernels~\cite{manzoor2016fast}; 2) detecting the appearance of anomalous graph signals created by abnormal groups of nodes~\cite{hooi2018changedar}; or 3) encoding graphs using frequent motifs~\cite{noble2003graph}.
However, none of these methods are deep learning-based.
As the time of writing, very few studies in ANOS GD with deep learning have been undertaken. As such, this is highlighted as a potential future direction in Section~\ref{sec:future:ED}.

\subsection{GNN Based Techniques}

Motivated by the success of GNNs in various graph classification tasks, the most recent works in ANOS GD employ GNNs to classify single graphs as normal/abnormal in the given graph database. Specifically, Dou \etal~\cite{dou2021user} transformed fake news detection into an ANOS GD problem by modeling news as tree-structured propagation graphs where the root nodes denote pieces of news, and child nodes denote users who interact with the root news. Their end-to-end framework, UPFD, extracts two embeddings for the news piece and users, respectively, via a text embedding model (e.g. word2vec, BERT) and a user engagement embedding process. For each news graph, its latent representation is a flattened concatenation of these two embeddings, which is input to train a neural classifier with the label of the news. 
Corresponding propagation graphs that are labeled as fake by the trained model are regarded as anomalous.

Another representative work by Zhao and Akoglu~\cite{zhao2020using} employed a GIN model and one-class classification (\ie DeepSVDD~\cite{ruff2018deep}) loss to train a graph-level anomaly detection framework in an end-to-end manner.
For each individual graph in the graph database, its graph-level embedding is generated by applying mean-pooling over its nodes' node-level embeddings.
A graph is eventually depicted as anomalous if it lies outside the learned hypersphere, as shown in Fig.~\ref{pic:graph_level}. 

\subsection{Network Representation Based Techniques}

It is also possible to apply general graph-level network representation techniques to ANOS GD.
With these methods, the detection problem is transformed into a conventional outlier detection problem in the embedding space.
In contrast to D(G)NN based techniques that can detect graph anomalies in an end-to-end manner, adopting these representation techniques for anomaly detection is two-staged.
First, graphs in the database are encoded into a shared latent space using graph-level representation techniques, such as Graph2Vec~\cite{narayanan2017graph2vec}, FGSD~\cite{verma2017hunt}.
Then, the anomalousness of each single graph is measured by an off-the-shelf outlier detector.
Essentially, this kind of approach involves pairing existing methods in both stages, yet, the stages are disconnected from each other and, hence, the detection performance can be subpar since the embedding similarities are not necessarily designed for the sake of anomaly detection.

\section{ANOS GD on Dynamic Graphs} \label{sec:anosgd:dynamic}

For dynamic graph environments, graph-level anomaly detection endeavors to identify abnormal graph snapshots/temporal graphs. Similar to ANOS ND and ED on dynamic graphs, given a sequence of graphs, anomalous graphs can be distinguished regarding their unusual evolving patterns, abnormal graph-level features, or other characteristics.

\begin{table*}[!h]
\centering 
\footnotesize
\renewcommand\arraystretch{1}
\setlength{\tabcolsep}{2.8mm}
\caption{Summary of Anomalous Edge, Sub-graph and Graph Detection Techniques.} 
\resizebox{0.96\textwidth}{!}{
\begin{tabular}{m{2.1cm}<{\centering}|m{2.3cm}<{\centering}|m{1.1cm}<{\centering}|m{4cm}<{\centering}|m{1.8cm}<{\centering}|m{2.9cm}<{\centering}}
\toprule[1 pt]
\textbf{Graph Type} & \textbf{Approach} & \textbf{Category} & \textbf{Objective Function} & \textbf{Measurement} & \textbf{Outputs}  \\ 
\midrule[1 pt]
\multicolumn{6}{c}{Anomalous Edge Detection Techniques} \\
\midrule[1 pt]
\multirow{2}{*}{\shortstack{Static Graph - \\ Plain}} &UGED~\cite{DBLP:conf/ijcnn/Ouyang0020} & DNN & $\text{cross-entropy}(f(u,N(u)), v)$  & Anomaly Score  & $\text{mean}(1-P(v|u,N(u)), 1-P(u|v,N(v)))$ \\  \cline{2-6}
&AANE~\cite{AANE} & GCN & $\mathcal{L} = \mathcal{L}_{afl} + \gamma\mathcal{L}_{aal}$  & Anomaly Ranking  & Edge Existing Probability \\ 
\midrule[1 pt]

Static Graph - Attributed &eFraudCom~\cite{zhangge2} & NR & $\mathcal{L}_{MAIN} - \lambda\mathcal{L}_{MI}$ & Anomaly Prediction & Predicted Label \\ 

\midrule[1 pt]

Dynamic Graph - Plain & NetWalk~\cite{yu2018netwalk} & NR & $\gamma\mathcal{L}_{AE} + \mathcal{L}_{Clique} + \lambda\|W\|_{F}^{2} + \beta KL$ & Anomaly Score & Nearest Distance to Cluster Centers \\
\midrule[1 pt]

Dynamic Graph - Attributed & AddGraph~\cite{zheng2019addgraph} & GCN & $\text{min} \sum_{{e\in \varepsilon^{t}}}\sum_{{e^{\prime}\notin \varepsilon^{t}}} \text{max}\{0, \gamma + f(i,j,w) - f(i^{\prime},j^{\prime},w)\} + \lambda\mathcal{L}_{reg}$  & Anomaly Score & $f(i,j,w) = w\cdot\sigma(\beta\cdot (||\mathbf{a} \odot \mathbf{h}_{i} + \mathbf{b} \odot \mathbf{h}_{j}||^{2}_{2} - \mu))$ \\ 

\midrule[1 pt]
\midrule[1 pt]
\multicolumn{6}{c}{Anomalous Sub-graph Detection Techniques} \\
\midrule[1 pt]
\multirow{2}{*}{\shortstack{Static Graph - \\ Plain}} &DeepFD~\cite{wang2018deep} & NR & $\mathcal{L}_{recon} +\alpha \mathcal{L}_{sim} + \gamma \mathcal{L}_{reg}$& Density-based Method (DBSCAN)  & Dense sub-graphs \\   \cline{2-6}

&FraudNE~\cite{FraudNE} & NR & $\mathcal{L}_{res}^{source} + \mathcal{L}_{res}^{sink} + \alpha \mathcal{L}_{share} + \eta \mathcal{L}_{reg}$  & Density-based Method (DBSCAN)  & Dense sub-graphs \\

\midrule[1 pt]
\midrule[1 pt]

\multicolumn{6}{c}{Anomalous Graph Detection Techniques} \\
\midrule[1 pt]
\multirow{2}{*}{\shortstack{Graph Database - \\ Attributed}}&UPFD~\cite{dou2021user} & NR & $ -(y\log(p) + (1-y)\log(1-p))$ & Anomaly Prediction & Predicted Label \\ \cline{2-6}
& OCGIN~\cite{zhao2020using} & GNN & $ \min\limits_{W}\frac{1}{N}\sum\limits_{i=1}^{N}||GIN(G_i,W)-c||^2 + \frac{\lambda}{2}\sum\limits_{l=1}^{L}||W^l||^2_{F}$ & Location in Embedding Space & Distance to Hypersphere Center \\
\midrule[1 pt]

\multirow{2}{*}{\shortstack{Dynamic Graph - \\ Plain}} & DeepSphere~\cite{teng2018deep} & DNN & $\mathcal{L} = \mathcal{L}_{h} + \lambda\mathcal{L}_{res}$ & Location in Embedding Space  & Anomalous Label  \\ \cline{2-6}
& GLAD-PAW~\cite{wan2021glad} & GNN & $\text{cross-entropy}(\mathbf{y},\mathbf{\hat{y}})$ & Anomaly Prediction  & Predicted Label  \\ 

\bottomrule[1 pt]
\multicolumn{6}{l}{* DNN: Deep NN Based techniques, GCN: GCN Based Techniques, NR: Network Representation Based Techniques.}\\
\multicolumn{6}{l}{* GNN: Graph Neural Network Based Techniques.}
\end{tabular}
\label{tb:edgeandsubgraph}
}
\end{table*}

In order to derive each graph snapshot/temporal graph’s characteristics, the commonly used GNN, LSTM and autoencoder are feasible to apply. For instance, Teng \etal~\cite{teng2018deep} applied a LSTM-autoencoder to detect abnormal graph snapshots, as shown in Fig.~\ref{pic:GDdynamic}.
In their proposed model, DeepSphere, a dynamic graph is described as a collection of three-order tensors, $\{\mathcal{X}_k, k=1,2...\}$ where each $\mathcal{X} \in \mathcal{R}^{N \times N \times T}$, and the slices along the time dimension are the adjacency matrices of graph snapshots.
To identify abnormal tensors, DeepSphere first embeds each graph snapshot into a latent space using an LSTM autoencoder, and then leverages a one-class classification objective~\cite{ruff2018deep} that learns a hypersphere such that normal snapshots are covered, and anomalous snapshots lay outside.
The LSTM autoencoder takes the adjacency matrices as input sequentially and attempts to reconstruct these input matrices through training.
The hypersphere is learned through a single neural network layer and its objective function is formulated as:
\begin{equation} \label{eq:ds:hsp}
    \mathcal{L}_{h} = r^{2} + \gamma\sum_{k=1}^{m} \epsilon_{k} + \frac{1}{m}\sum_{k=1}^{m} \|\textbf{z}_{k} - \textbf{a}\|^{2},
\end{equation}
where $\textbf{z}_{k}$ is the latent representation generated by the LSTM autoencoder, $\textbf{a}$ is the centroid of the hypersphere, $r$ is the radius, $\epsilon_{k}$ is the outlier penalty ($\epsilon_{k} = \|\textbf{z}_{k} - \textbf{a}\|^{2} - r^2$), $m$ is the number of training graph snapshots, and $\gamma$ is a hyperparameter.  
The overall objective function of DeepSphere is represented as:
\begin{equation} \label{eq:ds:obj}
    \mathcal{L} = \mathcal{L}_{h} + \lambda\mathcal{L}_{res},
\end{equation}
where $\mathcal{L}_{res}$ is the reconstruction loss of the LSTM autoencoder.
When the training is finished, DeepSphere spots a given unseen data $\mathcal{X}$ as anomalous if its embedding lies outside the learned hypersphere with a radius of $r$.

In addition to all ANOS ND, ED, SGD, and GD techniques reviewed above, it is worth mentioning that perturbed graphs, which adversarial models generate to attack graph classification algorithms or GNNs~\cite{zhang2021backdoor,dai2018adversarial,9338329}, can also be regarded as (intensional) anomalies. In a perturbed graph, the nodes and edges are modified deliberately to deviate from the others. We have not reviewed these in this survey because their main purpose is to attack a GNN model. The key idea behind these methods is the attacking/perturbation strategy, and studies in this sphere seldom focus on a detection or reasoning module to identify the perturbed graph or its sub-structures, \ie anomalous nodes, edges, sub-graphs, or graphs.

\begin{table*}[!t]
\centering 
\footnotesize
\renewcommand\arraystretch{1}
\setlength{\tabcolsep}{2.8mm}
\caption{Published Algorithms and Models} 
\resizebox{0.96\textwidth}{!}{
\begin{tabular}{c|c|c|c|l}
\toprule[1 pt]
\textbf{Model} & \textbf{Language} & \textbf{Platform} &  \textbf{Graph} & \textbf{Code Repository} \\ 
\midrule[1 pt]

AnomalyDAE~\cite{fan2020anomalydae}  & Python & Tensorflow   & Static Attributed Graph & https://github.com/haoyfan/AnomalyDAE   \\ \hline

MADAN~\cite{gutierrez2020multi}    & Python & -   & Static Attributed Graph & https://github.com/leoguti85/MADAN  \\ \hline

PAICAN~\cite{bojchevski2018bayesian}  & Python & Tensorflow  & Static Attributed Graph & http://www.kdd.in.tum.de/PAICAN/   \\ \hline

ONE~\cite{bandyopadhyay2019outlier} & Python & -  & Static Attributed Graph & https://github.com/sambaranban/ONE   \\ \hline

DONE\&AdONE~\cite{bandyopadhyay2020outlier}  & Python & Tensorflow    & Static Attributed Graph & https://bit.ly/35A2xHs \\ \hline

SLICENDICE~\cite{nilforoshan2019slicendice}  & Python & -  & Static Attributed Graph & http://github.com/hamedn/SliceNDice/ \\ \hline

FRAUDRE~\cite{zhangge1}  & Python & Pytorch  & Static Attributed Graph & https://github.com/FraudDetection/FRAUDRE \\ \hline

SemiGNN~\cite{SemiGNN}  & Python & Tensorflow  & Static Attributed Graph & https://github.com/safe-graph/DGFraud \\ \hline

CARE-GNN~\cite{CARE-GNN}  & Python & Pytorch & Static Attributed Graph & https://github.com/YingtongDou/CARE-GNN \\ \hline

GraphConsis~\cite{GraphConsis}  & Python & Tensorflow  & Static Attributed Graph & https://github.com/safe-graph/DGFraud \\ \hline

GLOD~\cite{zhao2020using}  & Python & Pytorch  & Static Attributed Graph & https://github.com/LingxiaoShawn/GLOD-Issues \\ \hline

OCAN~\cite{zheng2019one} & Python & Tensorflow  & Static Graph &https://github.com/PanpanZheng/OCAN \\ \hline

DeFrauder~\cite{DBLP:conf/ijcai/DhawanGK019} & Python & -  & Static Graph & https://github.com/LCS2-IIITD/DeFrauder \\ \hline

GCAN~\cite{lu2020gcan}  & Python & Keras & Heterogeneous Graph & https://github.com/l852888/GCAN \\ \hline

HGATRD~\cite{huang2020heterogeneous} & Python & Pytorch   & Heterogeneous Graph & https://github.com/201518018629031/HGATRD  \\\hline

GLAN~\cite{yuan2019jointly}  & Python & Pytorch  & Heterogeneous Graph & https://github.com/chunyuanY/RumorDetection \\ \hline

GEM~\cite{liu2018heterogeneous} & Python & -  & Heterogeneous Graph & https://github.com/safe-graph/DGFraud/tree/master/algorithms/GEM \\ \hline

eFraudCom~\cite{zhangge2} & Python & Pytorch & Heterogeneous Graph & https://github.com/GeZhangMQ/eFraudCom \\ \hline

DeepFD~\cite{wang2018deep} & Python & Pytorch  & Bipartite Graph & https://github.com/JiaWu-Repository/DeepFD-pyTorch \\ \hline

ANOMRANK~\cite{yoon2019fast} & C++ & - & Dynamic Graph & https://github.com/minjiyoon/anomrank \\ \hline

MIDAS~\cite{DBLP:conf/aaai/0001HYSF20} & C++ & - & Dynamic Graph & https://github.com/Stream-AD/MIDAS \\ \hline

Sedanspot~\cite{eswaran2018sedanspot}    & C++ &  -   & Dynamic Graph & https://www.github.com/dhivyaeswaran/sedanspot \\ \hline

F-FADE~\cite{chang2021f} & Python & Pytorch  & Dynamic Graph & http://snap.stanford.edu/f-fade/ \\ \hline

DeepSphere~\cite{teng2018deep}  & Python & Tensorflow   & Dynamic Graph & https://github.com/picsolab/DeepSphere   \\ \hline

Changedar~\cite{hooi2018changedar}  & Matlab & -   & Dynamic Graph & https://bhooi.github.io/changedar/  \\\hline

UPFD~\cite{dou2021user} & Python & Pytorch  & Graph Database & https://github.com/safe-graph/GNN-FakeNews \\ \hline

OCGIN~\cite{zhao2020using} & Python & Pytorch  & Graph Database & https://github.com/LingxiaoShawn/GLOD-Issues \\ \hline

DAGMM~\cite{zong2018deep} & Python & Pytorch  &  Non Graph & https://github.com/danieltan07/dagmm \\ \hline

DevNet~\cite{pang2019deep} & Python & Tensorflow  & Non Graph &https://github.com/GuansongPang/deviation-network \\ \hline

RDA~\cite{zhou2017anomaly} & Python & Tensorflow & Non Graph &https://github.com/zc8340311/RobustAutoencoder \\ \hline

GAD~\cite{GAD} & Python & Tensorflow  & Non Graph &https://github.com/raghavchalapathy/gad \\ \hline

Deep SAD~\cite{ruff2019deep} &Python & Pytorch  & Non Graph & https://github.com/lukasruff/Deep-SAD-PyTorch \\ \hline

DATE~\cite{10.1145/3394486.3403339} &Python & Pytorch  &Non Graph & https://github.com/Roytsai27/Dual-Attentive-Tree-aware-Embedding \\ \hline

STS-NN~\cite{STS-NN} & Python & Pytorch  & Non Graph & https://github.com/JiaWu-Repository/STS-NN \\
\bottomrule[1 pt]
\multicolumn{5}{l}{* -: No Dedicated Platforms.}
\end{tabular}
\label{tb:publishedalgs}
}
\end{table*}

\begin{table*}[!t]
\centering 
\footnotesize
\renewcommand\arraystretch{1}
\setlength{\tabcolsep}{2.8mm}
\caption{Published Datasets.}
\resizebox{0.96\textwidth}{!}{
\begin{tabular}{m{1.6cm}<{\centering}|m{1.4cm}<{\centering}|m{0.5cm}<{\centering}|m{0.5cm}<{\centering}|m{0.5cm}<{\centering}|m{0.5cm}<{\centering}|m{0.5cm}<{\centering}|m{2cm}<{\centering}|l}
\toprule[1 pt]
\textbf{Category} & \textbf{Dataset} &  \textbf{\#G} & \textbf{\#N} & \textbf{\#E} & \textbf{\#FT} & \textbf{\#AN} & \textbf{REF} & \textbf{URL} \\ 
\midrule[1 pt]
\multirow{5}{*}{\shortstack{Citation \\ Networks}} 
& ACM & 1 & 16K & 71K & 8.3K & - &\cite{ding2019deep,ding2019interactive,fan2020anomalydae,ding2020inductive} &\text{http://www.arnetminer.org/open-academic-graph} \\   \cline{2-9}
& Cora & 1 & 2.7K & 5.2K & 1.4K & - &\cite{li2019specae,bandyopadhyay2020outlier,bandyopadhyay2019outlier,liang2018semi,bojchevski2018bayesian} &http://linqs.cs.umd.edu/projects/projects/lbc \\    \cline{2-9}
& Citeseer & 1& 3.3K & 4.7K & 3.7K & - &\cite{bandyopadhyay2020outlier,bandyopadhyay2019outlier,liang2018semi,perozzi2016scalable} & http://linqs.cs.umd.edu/projects/projects/lbc \\    \cline{2-9}
& Pubmed & 1& 19K & 44K & 500 & - &\cite{li2019specae,bandyopadhyay2020outlier,bandyopadhyay2019outlier,liang2018semi} &http://linqs.cs.umd.edu/projects/projects/lbc \\   \cline{2-9}
& DBLP & 1& - & - & - & -  &\cite{yu2018netwalk,eswaran2018sedanspot,bojchevski2018bayesian,hu2016embedding,perozzi2016scalable} &http://www.informatik.uni-trier.de/˜ley/db/ \\
\midrule[1 pt]
\multirow{10}{*}{\shortstack{Social \\ Networks}}
&Enron & - & 80K & - & - & - & \cite{li2017radar,peng2018anomalous,zhang2019robust,gutierrez2020multi,wang2019detecting,yoon2019fast,eswaran2018sedanspot,eswaran2018spotlight,rayana2016less}  & http://odds.cs.stonybrook.edu/\#table2 \\    \cline{2-9}
&UCI Message & 1& 5K & - & - & -  &\cite{yu2018netwalk,cai2020structural,zheng2019addgraph} &http://archive.ics.uci.edu/ml \\  \cline{2-9}
&Google+ & 4 & 75M & 11G & - & -  & -  & https://wangbinghui.net/dataset.html \\ \cline{2-9}
&Twitter Sybil & 3 & 41M & - & - & 100K  & -  & https://wangbinghui.net/dataset.html \\ \cline{2-9}
&Twitter WorldCup2014 & - & 54K & - & - & -  & \cite{rayana2016less}  & http://shebuti.com/SelectiveAnomalyEnsemble/ \\ \cline{2-9}
&Twitter Security2014 & - & 130K & - & - & -  & \cite{rayana2016less}  & http://shebuti.com/SelectiveAnomalyEnsemble/ \\ \cline{2-9}
&Reality Mining & - & 9.1K & - & - & -  & \cite{rayana2016less}  & http://shebuti.com/SelectiveAnomalyEnsemble/ \\ \cline{2-9}
&NYTNews & - & 320K & - & - & -  & \cite{rayana2016less}  & http://shebuti.com/SelectiveAnomalyEnsemble/ \\ \cline{2-9}
&Politifact & 314 & 41K & 40K & - & 157  &\cite{dou2021user}  &https://github.com/safe-graph/GNN-FakeNews \\ \cline{2-9}
&Gossipcop & 5.4K & 314K & 308K & - & 2.7K & \cite{dou2021user} &https://github.com/safe-graph/GNN-FakeNews \\ 
\midrule[1 pt]
\multirow{5}{*}{\shortstack{Co-purchasing \\ Networks}}
& Disney & 1 & 124 & 334 & 30 & 6  &\cite{li2017radar,peng2018anomalous,liu2017accelerated,zhang2019robust,gutierrez2020multi} &https://www.ipd.kit.edu/mitarbeiter/muellere/consub/ \\  \cline{2-9}
&Amazon-v1 & 1 & 314K & 882K & 28 & 6.2K  &\cite{peng2018anomalous,hooi2017graph,zhu2020mixedad,kumar2018rev2,bojchevski2018bayesian,hu2016embedding,hooi2016fraudar} &https://www.ipd.kit.edu/mitarbeiter/muellere/consub/ \\ \cline{2-9}
&Amazon-v2  & 1 & 11K & - & 25 & 821  & -&https://github.com/dmlc/dgl/blob/master/python/dgl/data/fraud.py \\ \cline{2-9}
&Elliptic & 1 & 203K & 234K & 166 & 4.5K  & - &https://www.kaggle.com/ellipticco/elliptic-data-set \\\cline{2-9}
&Yelp  & 1 & 45K & - & 32 & 6.6K  & -&https://github.com/dmlc/dgl/blob/master/python/dgl/data/fraud.py \\ 

\midrule[1 pt]
Transportation Networks &  New York City Taxi& - & - & - & - & - &\cite{teng2018deep,teng2017anomaly,eswaran2018spotlight} &http://www.nyc.gov/html/tlc/html/about/triprecorddata.shtml \\

\bottomrule[1 pt]
\multicolumn{9}{l}{* -: Not Given, \#G: Number of Graphs, \#N: Number of Nodes, \#E: Number of Edges, \#FT: Number of Features, \#AN: Number of Anomalies, REF: References.}
\end{tabular}
\label{tb:publisheddataset}
}
\end{table*}

\section{Published Algorithms and Datasets} \label{sec:resources}

Acquiring open-sourced implementations and real-world datasets with real-world anomalies are far from trivial in academic research on graph anomaly detection.
Here, we first list the published algorithms with publicly available implementations, then we provide a collection of public benchmark datasets and summarize the commonly used evaluation metrics. 
Lastly, due to the shortage of labeled anomalies in real-world datasets, we review three synthetic dataset generation strategies used in the existing works.
All the resources are available at: https://github.com/XiaoxiaoMa-MQ/Awesome-Deep-Graph-Anomaly-Detection/.

\subsection{Published Algorithms} 
The published implementations of algorithms and models contribute to baseline experiments. Table~\ref{tb:publishedalgs} provides a summary of published implementations outlining the language and platforms, graphs that they can admit, and URLs to code repositories.

\subsection{Published Datasets}

Table~\ref{tb:publisheddataset} summarizes the most commonly used datasets, categorizing them into different groups with regard to their application fields. Notably, there is a lack of anomaly ground truth with labeled anomalies only provided in the Enron, Twitter Sybil, Disney, Amazon, Elliptic and Yelp datasets. Details of the DBLP, UCI message, Digg, Wikipedia, and New York city taxi datasets are not given because these public datasets only contain the raw data, and in most existing works, they are further processed to build different graphs (\eg homogeneous graphs, bipartite graphs).
The well-known citation networks are often used to generate synthetic datasets by injecting anomalies into them - the number of anomalies varies from study to study.

In addition to these anomaly detection datasets, Table~\ref{tb:publisheddataset} also lists eight graph classification datasets.
As mentioned in~\ref{sec:synthetic}, through downsampling, these datasets can be used as benchmarks for evaluating the anomaly detection performance.

\subsection{Synthetic Dataset Generation} \label{sec:synthetic}

Given the rarity of ground-truth anomalies, many researchers have employed synthetic datasets to investigate the effectiveness of their proposed methods~\cite{bandyopadhyay2019outlier,DBLP:journals/aim/SenNBGGE08,DBLP:conf/icdm/SanchezMLKB13}. 
Typically, these datasets can be categorized as follows: 
\begin{itemize}
    \item \textit{Synthetic graphs with injected anomalies.} Pursuing this strategy, graphs are created to simulate real-world networks. All the nodes and edges are manually added with well-known benchmarks (\eg Lanchinetti-Fornunato-Radicchi (LFR)~\cite{lancichinetti2009community}, small-world~\cite{watts1998collective}, scale-free graphs~\cite{akoglu2009rtg}). Once built, ground-truth anomalies are planted into the network. For the feasibility of generating expected scale of networks, this strategy is mostly used by previous works to validate their underlying intuitions in anomaly detection.
    \item \textit{Real-world datasets with injected anomalies.} These datasets are built based on the real-world networks. In particular, anomalies are created either by modifying the topological structure or the attributes of existing nodes/edges/sub-graphs, or by inserting non-existent graph objects. 
    \item \textit{Downsampled graph classification datasets.} The widely-used graph classification datasets (\eg NCI1, IMDB, ENZYMES in~\cite{zhao2020using}) can be easily converted into sets suitable for anomaly detection through two steps. Firstly, one particular class and its data records are chosen to represent normal objects. Then, the other data records are downsampled as anomalies at a specified downsampling rate. By this, the generated graph anomaly detection dataset is, in fact, a subset of the original dataset. The most significant strength of this strategy is that no single data record has been modified.
\end{itemize}

\subsection{Evaluation Metrics} \label{sec:em}

To date, the most widely used metrics for evaluating anomaly detection performance evaluation include accuracy, precision, recall rate, F1-score and AUC-AP (Average Precision). Their formulas/descriptions are given in Table~\ref{tb:evluationmetrics}. However, a more dedicated analysis with some new evaluation metrics is needed for further performance examination 
because anomaly detection has different requirements for different applications~\cite{bozarth2020toward,DBLP:conf/cybersecpods/ElmrabitZL020,DBLP:conf/nss/EnglyLM20}, \eg false negative and false positive.
For instance, network intrusion prevention systems are more sensitive to false negative errors, while false positive errors are considered relatively harmless. This is because any risky connections should be shut down to prevent information leaks.
By contrast, other applications concentrate more on the false positives, \eg in auditing domain, companies often set a budget for an auditor to look at flagged anomalies, and they want high precision/small false positive rate such that the auditor's time can be best used.
Hence, when evaluating detection performance, we suggest reviewing the specific requirements of the application domain for fair and suitable comparisons.

\begin{table}
\centering 
\footnotesize
\renewcommand\arraystretch{1}
\setlength{\tabcolsep}{2.8mm}
\caption{Evaluation Metrics. $tp$: true positives; $tn$: true negatives; $fp$: false positives; $fn$: false negatives.}
\resizebox{0.46\textwidth}{!}{
\begin{tabular}{l|c}
\toprule[1 pt]
\textbf{Evaluation Metric} & \textbf{Formula/Description} \\ 
\midrule[1 pt]
Accuracy &  $\frac{tp + tn}{tp + tn + fp + fn}$ \\\hline
Precision &  $\frac{tp}{tp + fp}$ \\\hline
Recall & $\frac{tp}{tp + fn}$ \\\hline
F1 Score & $2 * \frac{Recall * Precision}{Recall + Precision}$ \\\hline
AUC-ROC & The Area Under the ROC Curve\\ \hline
AUC-AP &  The Area Under Precision-Recall Curve \\

\bottomrule[1 pt]
\end{tabular}
\label{tb:evluationmetrics}
}
\end{table}

\section{Future Directions} \label{sec:futures}

So far, we have reviewed the contemporary deep learning techniques that are devoted to graph anomaly detection.
An apparent observation from our survey is there remain many compounded challenges imposed by the complexity of anomaly detection, graph data, and the immaturity of deep learning techniques for graph data mining. Another observation is that deep learning techniques in graph anomaly detection are still confined to a relatively small number of studies, and most of these focus on anomalous node detection. For a gauge, simply compare the length of Tables~\ref{table:ANOSND} and~\ref{tb:edgeandsubgraph}. Edge, sub-graph, and graph-level anomaly detection have clearly received much less attention.
To bridge the gaps and push forward future work, we have identified 12 directions of future research for graph anomaly detection with deep learning.

\subsection{Anomalous Edge, Sub-graph, and Graph Detection} \label{sec:future:ED}
In real-world graphs, anomalies also appear as unusual relationships between objects, sub-structures formed by abnormal groups, or abnormal graphs, which are known as anomalous edges, sub-graphs, and graphs respectively.
As indicated in our review, there is a huge gap between the existing anomalous edge/sub-graph/graph detection techniques and the emerging demands for more advanced solutions in various application domains (\eg social networks, computer networks, financial networks).
When detecting anomalous edges/sub-graphs/graphs, the proposed methods should be capable of leveraging the rich information contained in graphs to find clues and characteristics that can distinguish normal objects and anomalies in specific applications.
Typically, this involves extracting edge/sub-graph/graph-level features, modeling the patterns of these features, and measuring the abnormalities accordingly.
However, current deep learning based graph anomaly detection techniques put forward very little effort in this regard.

\textit{Opportunities}: We believe more research efforts can be done on anomalous edge, sub-graph, and graph detection with regard to their significance in real-world applications. Possible solutions to this gap might to be first consider the application domain and explore domain knowledge to find complementary clues as a basis for these problems. Then, motivated by recent advances in deep learning for edge, sub-graph, and graph-level representation learning~\cite{DBLP:journals/ijdsa/XuWCY20,DBLP:conf/nips/AlsentzerFLZ20}, extensive work can be done to learn an anomaly-aware embedding space such that it is feasible to extract abnormal patterns of anomalies. Although this direction seems quite straightforward, the true challenge lies in the specific application domains. Hence, domain knowledge, anomalous pattern recognition and anomaly-aware deep learning techniques should be enforced simultaneously.

\subsection{Anomaly Detection in Dynamic Graphs}
Dynamic graphs provide powerful machinery with which to capture the evolving relationships between real objects and their attributes.
Their ever-changing structure and attribute information inherently make anomaly detection very challenging in these scenarios, leading to two primary concerns for the task.
The first is to consider the spatial and temporal information contained in each graph snapshot at different time stamps, and the second is to explore the evolving patterns of nodes, edges, sub-graphs and graphs, as well as their interaction with the node/edge attributes over time.
When these challenges have been tackled with mature solutions, detection techniques will achieve better results.

\textit{Opportunities}: From our observations, most of deep learning based dynamic graph anomaly detection techniques are built on DeepWalk~\cite{perozzi2014deepwalk}, GCN~\cite{kipf2016gcn} or other deep models that are intuitively designed for static graphs. This means other information, like evolving patterns in attributes~\cite{wang2019time,zhang2018salient}), are not adequately used in the detection task.
We can therefore identify the following directions for future studies to target.
\begin{itemize}
    \item \textit{Using dynamic graph mining tools.} As a popular research topic, deep learning for dynamic graph data mining~\cite{DBLP:conf/wsdm/SankarWGZY20,DBLP:journals/jmlr/KazemiGJKSFP20} has shown its effectiveness in supporting dynamic graph analysis, such as node clustering and graph classification~\cite{cui2018survey}. More future works can be foreseen that adopt these techniques for anomaly detection.
    \item \textit{Deriving solid evidence for anomaly detection.} The rich structural, attribute and temporal information in dynamic graphs are valuable resources for identifying anomalies. Apart from the indicators widely used in current works, such as burst of connections between node pairs or suddenly vanishing connections, we suggest exploring structural and attribute changes in depth. From such studies, we may derive additional information to enhance the detection performance, such as the appearance of abnormal attributes.
    \item \textit{Handling complex dynamics.} Real-world networks always exhibit changes in both the network structure and node attributes, but only very few studies address this circumstance. Most of the state-of-arts only consider changes in one of these aspects. Although this `double' scenario is extremely complex and detecting anomalies in this kind of dynamic graph is very challenging, it is worth studying because these graphs are highly reflective of real network data.
\end{itemize}

\subsection{Anomaly Detection in Heterogeneous Graphs}
Heterogeneous graphs are a specific type of graph that contain diverse types of nodes and edges.
For instance, Twitter can be intuitively modeled as a heterogeneous graph comprised of tweets, users, words, etc.

\textit{Opportunities}: To use the complex relationships between different types of nodes in heterogeneous graphs for anomaly detection, representative works, such as HGATRD~\cite{huang2020heterogeneous}, GCAN~\cite{lu2020gcan} and GLAN~\cite{yuan2019jointly}, typically decompose a heterogeneous graph into individual graphs according to meta-paths, \eg one with tweets and users, and another with tweets and words. They then use D(G)NNs to learn the embeddings for graph anomaly detection. Such a decomposition inherently overlooks the direct inter-relations among diverse types of nodes/edges and downgrades the effectiveness of the embeddings. A possible solution is to reveal the complex relations between different types of nodes and edges, and encode them into a unique representation for boosted detection performance.

\subsection{Anomaly Detection in Large-scale Graphs}
The scalability of methods to high-dimensional and large-scale data is an ongoing and significant challenge to anomaly detection techniques.
In face of large-scale networks, such as Facebook and Twitter that contain billions of users and friendship links, the size of data in terms of both graph size and number of node attributes is extremely high.
However, most of the existing works lack the ability to detect anomalies in such large-scale data because they are transductive models and need to take the whole graph as input for further analysis.
Computation time and memory cost increase dramatically as the network scales up, and this stops existing techniques from being used on large-scale networks.

\textit{Opportunities}: Accordingly, there is a need for scalable graph anomaly detection techniques.
One possible approach would be an inductive learning scheme that first trains a detection model on part of the whole graph and then applies the model to detect anomalies in the unseen data.
As some inductive learning models, such as GraphSAGE~\cite{GraphSAGE}, have shown their effectiveness on link prediction and node classification in large-scale graphs, this approach is expected to provide a basis for graph anomaly detection in large-scale graphs and similar techniques can be investigated in the future.

\subsection{Multi-view Graph Anomaly Detection}
In real-world networks, objects might form different kinds of relationships with others (\eg user's followership and friendship on Twiter). And their attribute information might be collected from different resources, such as user's profile, historical posts.
This results in two types of multi-view graphs: 1) multi-graph that contains more than one type of edges between two nodes~\cite{DBLP:conf/aaai/KhanB19,fan2020one2multi}; and 2) multi-attributed-view graph that stores node attributes in different attributed views~\cite{DBLP:conf/ijcai/ChengWTXG20,peng2020deep,zhang2016identifying}.

\textit{Opportunities}: These multi-views basically allow us to analyze real objects' characteristics from different perspectives.
Each view also provides complementary information to other views, and they might have different significance on anomaly detection. 
For instance, anomalies might be indistinguishable in one view but are obviously divergent from the majority in another view.
There are a variety of work in data mining on multi-view learning~\cite{xiao2015temporal,gujral2020beyond}. However, work that can accommodate multi-view graphs along with multi-view attributes on nodes for anomaly detection purposes is nascent.
Moreover, the rich information contained in multiple views and the inconsistency among them are overlooked in these works. 
To this end, we believe more research effort in this direction is needed. Digesting the relationships between views will be vital to their success, as two views might provide contrary/supplementary information for anomaly detection.

\subsection{Camouflaged/Adversarial Anomaly Detection}
The easy accessibility of online platforms has made them convenient targets for fraudsters, attackers and other malevolent agents to carry out malicious activities.
Although various anomaly detection systems have been deployed to protect benign objects, anomalies can still conceal themselves to evade detection~\cite{shah2014spotting}.
Known as camouflaged anomalies, these entities typically disguise themselves as regular objects.
If the detection techniques are not robust against such cases, \ie if they cannot quickly and effectively adapt to the evolving behavior of evasion-seeking attackers, the anomalies are simply left to cause their damage.
 
\textit{Opportunities}: In the face of camouflage, the boundary between anomalies and regular objects is blurred, making anomalies much harder to be identified.
We believe extensive effort should be placed on detecting these anomalies because, as yet, very few studies have looked at handling camouflaged anomalies in graphs~\cite{CARE-GNN,hooi2016fraudar,hooi2017graph}.
To fulfill this gap, one major direction might be to jointly analyze the attributes, co-relations, such as the triadic, tetradic, or high-order relationships between objects in hypergraphs~\cite{chen2020multi,guzzo2017malevolent,sun2021heterogeneous,silva2008hypergraph}, and other information comprised in graphs.
By this, anomalies that only camouflage their local structures or attributes can be identified effectively.
Enhancing existing techniques might be another direction. 
This involves incorporating additional detection mechanisms or function blocks particularly designed for distinguishing camouflaged anomalies with existing detection techniques.
Consequently, these techniques will bridge most existing works and camouflaged anomaly detection. 

\subsection{Imbalanced Graph Anomaly Detection}
Anomalies are rare, which means anomaly detection always occurs coexists with class imbalance in the training data. As deep learning models rely heavily on the training data, such imbalance pose great challenges to graph anomaly detection, and this remains a significant obstacle to deep learning techniques. 
Typically, the imbalanced class distributions will downgrade the detection techniques' ability to capture the differences between anomalies and non-anomalies. It might even cause over-fitting on the anomalous class because there are too few anomalies in the data.
If the detection model overlooks this critical fact and is trained improperly, detection performance will be sub-optimal.

\textit{Opportunities}: In fact, class imbalance has been widely explored in various research areas~\cite{zhangge1,ding2021cross}.
Advances such as under-sampling the majority class or modifying the algorithms shed important light on solving training problems with imbalances.
Yet, contemporary graph anomaly detection methods rarely incorporate these techniques. For more effective detection techniques, biased models that pay more attention to anomalies, such as penalizing additional training losses on misclassified anomalies, would be a possible direction to circumvent the problem.
Moreover, when adopting graph neural networks that aggregate neighboring information to the target node, like GCN or GraphSAGE, the over-smoothing between the features of connected nodes should be prevented such that the distinguishable features of the anomalies can be preserved to support anomaly detection.

\subsection{Multi-task Anomaly Detection}
Graph anomaly detection has close relations with other graph mining tasks including community detection~\cite{ijcai2020-693} and node classification~\cite{tang2016node}, and link prediction~\cite{gao2011temporal}. 
For a concrete example, when detecting community anomalies, community detection techniques are usually used to extract the community structures prior to anomaly detection. Meanwhile, the anomaly detection results can be used to optimize the community structure.
Such mutually beneficial collaborations between anomaly detection and other tasks inherently suggest an opportunity for multi-task learning that can handle diverse tasks simultaneously and share information among tasks.

\textit{Opportunities}: Multi-task learning provides effective machinery with which incorporate associated tasks~\cite{DBLP:conf/aaai/SanhWR19,DBLP:conf/aaai/HesselSE0SH19}. Its utmost advantage is that the training signal from another task could yield complementary information to distinguish anomalies from non-anomalies. The result would be enhanced detection performance. However, very few attempts focus on this at present.
Beyond current works, such as~\cite{GraphRfi} that jointly perform anomalous node detection and personalized recommendation, explorations into combining other learning tasks with graph anomaly detection are likely to emerge as a fruitful future direction.

\subsection{Graph Anomaly Interpretability}
The interpretability of anomaly detection techniques is vital to the subsequent anomaly handling process.
When applying these techniques to real applications, such as financial and insurance systems, it is essential to provide explainable and lawful evidence to support the detection results.
However, most of the existing works lack the ability to provide such evidence.
To identify the anomalies, the most commonly used metrics are top-k rankings and simple anomaly scoring functions.
These metrics are flexible enough to label objects as being either an anomaly or not an anomaly, but they cannot derive solid explanations.
Moreover, as deep learning techniques have also been criticized for their low interpretability, future works on graph anomaly detection with deep learning should pay much more attention to this~\cite{pang2020self}.

\textit{Opportunities}: To bridge this gap, integrating specially designed interpretation algorithms or mechanisms~\cite{DBLP:conf/nips/Sanchez-Lengeling20,attribution1} into the detection framework would be a possible solution, noting that this would inherently induce a higher computational cost.
Future works should therefore balance the cost of anomaly detection performance and interpretability.
Visualization-based approaches, \eg dashboards, charts, might also be feasible for showing the distinction between anomalies and non-anomalies in a human-friendly manner.
Further research in this direction will be successful if interpretable visualization results can be given~\cite{hohman2019s}. 

\subsection{Graph Anomaly Identification Strategies}
Amongst existing unsupervised graph anomaly detection techniques, anomalies are mainly identified based on residual analysis~\cite{peng2018anomalous,li2017radar}, reconstruction loss~\cite{fan2020anomalydae}, distance-based statistics~\cite{yu2018netwalk}, density-based statistics~\cite{li2019specae}, graph scan statistics~\cite{kulldorff1997spatial,neill2005detection,sharpnack2013near,berk1979goodness}, and one-class classification~\cite{wang2020ocgnn}.
The underlying intuition of these identification strategies is that anomalies have inconsistent data patterns with regular objects, and they will, therefore: 1) introduce more residual errors or be harder to reconstruct; or 2) be located in low-density areas or far away from the majority class in an anomaly-aware feature space.
Effort toward designing novel loss functions for GNNs for anomaly detection is currently quite limited~\cite{GAL}

\textit{Opportunities}: Although these strategies could capture the deviating data patterns of anomalies, they also have different limitations. Specifically, the residual analysis, one-class classification and reconstruction loss strategies are sensitive to noisy training data. Noisy nodes, edges or sub-graphs also exhibit large residuals, a large distance to the origin/hypersphere center and high reconstruction losses. Meanwhile, the distance-based and density-based strategies can only be applied when anomalies and non-anomalies are well separated in lower-dimensional space. Detection performance also downgrades dramatically if the gap between anomalies and non-anomalies is not that evident.
It calls for extensive future efforts to break these limitations and explore new anomaly identification strategies.

\subsection{Systematic Benchmarking}
Systematic benchmarking is key to evaluating the performance of graph anomaly detection techniques.
As indicated in our analysis in Section~\ref{sec:em}, recent studies are constantly raising attention to more comprehensive and effective benchmarking~\cite{bozarth2020toward,DBLP:conf/cybersecpods/ElmrabitZL020,DBLP:conf/nss/EnglyLM20,zhao2020using}. 
Typically, the benchmarking framework consists of benchmark datasets, baseline methods, evaluation metrics, and further analysis tools.
When evaluating a techniques' performance with other baselines, the evaluation dataset and metrics become very important because the performance of each model may vary depending on the setting.
The shortage of public datasets and (public available) baseline methods also imposes great challenges for effective evaluation.
Although one of the aims of our survey is to provide extensive materials for this purpose, like open-sourced implementations, datasets, and evaluation metrics, this work can only serve as a basis for future investigations into systematic benchmarking. We invite more efforts from the anomaly detection community toward this important case.
Certainly, rigorous attention to designing better frameworks for bechmarking would help to disclose the advances and shortcomings of various detection techniques and essentially track a unprejudiced and accurate progress record in this field.

\subsection{Unified Anomaly Detection Framework}
A graph anomaly can be categorized as an anomalous node, edge, and sub-graph in a single graph or an anomalous graph in a graph database.
These anomalies usually coexist in real-world datasets.
For instance, individual fraudsters, abnormal relationships, and fraud groups exist concurrently in online social network, as shown in Fig.~\ref{Toy}.
Moreover, there may be different ways to define anomalies of a certain type, such as community outliers versus anomalous communities or attribute-based versus structural anomalies.
When deploying detection techniques in real applications, it is expected that all types of anomalies can be identified while consuming the least resources and time.
A straightforward approach would be to integrate independent anomalous node, edge, and sub-graph detection techniques.
Although this is convenient to apply to relatively small networks, its high computational cost will surely prevent the approach from scaling to large networks, such as Facebook and Twitter, because the same graph data has to be loaded and processed more than once by different techniques.

\textit{Opportunities}: Unified frameworks that can detect diverse types of anomalies together~\cite{DBLP:conf/bigdataconf/BabaieCAY14,DBLP:conf/issre/LiCJHY20} may provide feasible solutions to bridge the gap.
To build such frameworks, one possible direction is to capture all the information needed by different detection techniques simultaneously so that these techniques can be applied.
The idea seems to be non-challenging, but in deep learning, methods of designing neural network layers and learning strategies that can fulfill this need will require extensive effort.

\section{Conclusion} \label{sec:conclusion}

Due to the complex relationships between real-world objects and recent advances in deep learning, especially graph neural networks, graph anomaly detection with deep learning is currently at the forefront of anomaly detection.
To the best of our knowledge, this is the first survey to present a comprehensive review dedicated to graph anomaly detection with modern deep learning techniques.
Specifically, we have reviewed and categorized contemporary deep learning techniques according to the types of graph anomalies they can detect, including: (1) anomalous node detection; (2) anomalous edge detection; (3) anomalous sub-graph detection; and, finally, (4) anomalous graph detection.
Clear summarizations and comparisons between different works are given to offer a complete and thorough picture of the current work and progress of graph anomaly detection as a field.

Moreover, to push forward future research in this area, we have provided a basis for systematic benchmarking by compiling a wide range of commonly used datasets, open-sourced implementations and synthetic dataset generation techniques.
We further highlight 12 potential directions for future work based on the survey results.
It is our firm belief that graph anomaly detection with deep learning is indeed more than a burst of temporary interest, and numerous applications from diverse domains are to surely benefit from it for the years to come.

\bibliographystyle{IEEEtran}
\bibliography{main.bib}


\clearpage
\setcounter{page}{1}
\appendices

\section{Challenges in Graph Anomaly Detection} \label{appendix:data-challenges}
Due to the complexity of anomaly detection and graph data mining, adopting deep learning technologies for graph anomaly detection faces a number of challenges:

\textbf{Data-CH1. Ground-truth is scarce.} In most cases, there is little or no prior knowledge about the features or patterns of anomalies in real applications. Ground-truth anomalies are often identified by domain experts, and this generally cost-prohibitive. As a result, labeled ground-truth anomalies are often unavailable for analysis in a wide range of disciplines.

\textbf{Data-CH2. Various types of graphs.} Different graphs model different real-world data. For instance, plain graphs contain only structural information, attributed graphs contain both structural and attribute information, and heterogeneous graphs represent the complex relations between different types of objects.
These graphs reflect the real-world data in different forms, and graph anomalies will show different deviating patterns in different types of graphs.

\textbf{Data-CH3. Various types of graph anomalies.} Given a specific type of graph, graph anomalies could appear as a specific node, edge, sub-graph, or an entire graph, and each type of these anomalies is significantly different from others. This means detection methods must involve concise definitions of anomalies and be able to identify concrete clues about the deviating patterns of anomalies.

\textbf{Data-CH4. High dimensionality and large scale.} Representing the structure information of real-world networks usually results in high dimensional and large-scale data~\cite{NIPS20201} because real-world network often contain millions or billions of nodes. Graph anomaly detection techniques, hence, should be capable of handling such high dimensional and large scale data; this includes the ability to extract anomalous patterns under the constraints of execution time and feasible computing resources. 

\textbf{Data-CH5. Interdependencies and dynamics.} The relationships between real objects reveal their interdependencies and they can no longer be treated individually for anomaly detection. That is to say, the detection techniques need to consider the deviating patterns of anomalies by assessing the pairwise, triadic, and higher relationships among objects restored in conventional graphs or hypergraphs~\cite{silva2008hypergraph,sun2021heterogeneous,guzzo2017malevolent,chen2020multi}.
In addition, the dynamic nature of real-world networks makes detection problems much more challenging.
 
\textbf{Data-CH6. Class imbalance.} As anomalies are rare occurrences, only a very small proportion of the real-world data might be anomalous. This naturally introduces a critical class imbalance problem to anomaly detection because the number of normal objects is far greater than anomalies in the training data. If no further actions are taken to tackle this challenge, learning-based anomaly detection techniques might overlook the patterns of anomalies, leading to sub-optimal results.

\textbf{Data-CH7. Unknown and camouflage of anomalies.} In reality, knowledge about anomalies mainly stems from human expertise. There are still many unknown anomalies across different application domains, and new types of anomalies might appear in the future. Nevertheless, real-world anomalies can hide or be camouflaged as benign objects to bypass existing detection systems. In graphs, anomalies might hide themselves by connecting with many normal nodes or by mimicking their attributes. Detection methods, therefore, need to be adaptive to unknown and novel anomalies and robust to camouflaged anomalies.

These data-specific challenges and technical-specific challenges (discussed in Section~\ref{sec:introdution:challenges}) are summarized in Table.~\ref{tb:challenges}, along with the corresponding articles that aim to address them.

\begin{table}
\centering 
\footnotesize
\renewcommand\arraystretch{1}
\setlength{\tabcolsep}{2.8mm}
\caption{Challenges and Methods.}
\resizebox{0.48\textwidth}{!}{
\begin{tabular}{m{1.1cm}<{\centering}|m{2cm}<{\centering}|m{2.9cm}<{\centering}}
\toprule[1 pt]
\textbf{Challenges} & \textbf{Details} &  \textbf{Methods} \\ 
\midrule[1 pt]
\multicolumn{3}{c}{\textbf{Data-specific Challenges}} \\
\midrule[1 pt]
Data-CH1 & Ground-truth is scarce              & \cite{li2017radar,ding2019interactive,bandyopadhyay2020outlier,bandyopadhyay2019outlier,yu2018netwalk,liu2017accelerated,ding2020inductive,teng2018deep,zheng2019one,peng2020deep,wang2019fdgars,SemiGNN,DBLP:conf/ijcnn/Ouyang0020,zheng2019addgraph} \\ 
\cline{1-3}
Data-CH2 & Various types of graphs             & \cite{liang2018semi,peng2018anomalous,ding2019deep,li2019specae,ding2019interactive,fan2020anomalydae,bandyopadhyay2020outlier,bandyopadhyay2019outlier,liu2017accelerated,zhang2019robust,peng2020deep,CARE-GNN,wang2018deep,FraudNE,dou2021user} \\ 
\cline{1-3}
Data-CH3 & Various types of graph anomalies    & \cite{li2019specae,ding2019interactive,peng2020deep,CARE-GNN,AANE} \\ 
\cline{1-3}
Data-CH4 & High dimensionality and large scale & \cite{zong2018deep,fan2020anomalydae,bandyopadhyay2020outlier,yu2018netwalk,liu2017accelerated,teng2018deep,hu2016embedding,wu2017adaptive} \\ 
\cline{1-3}
Data-CH5 & Interdependencies and dynamics      & \cite{peng2018anomalous,ding2019deep,li2019specae,ding2019interactive,fan2020anomalydae,bandyopadhyay2020outlier,bandyopadhyay2019outlier,yu2018netwalk,liu2017accelerated,teng2018deep,gutierrez2020multi,hu2016embedding,wu2017adaptive,peng2020deep,wang2019fdgars,SemiGNN,teng2017anomaly,DBLP:conf/ijcnn/Ouyang0020,AANE,zheng2019addgraph,wang2018deep,FraudNE,dou2021user} \\ 
\cline{1-3}
Data-CH6 & Class imbalance                     & \cite{zong2018deep,SemiGNN,GAL,zhao2020using} \\ 
\cline{1-3}
Data-CH7 & Unknown and camouflage of anomalies & \cite{li2017radar,peng2018anomalous,liu2017accelerated,hooi2017graph,ding2020inductive,teng2018deep,CARE-GNN,zheng2019addgraph,wang2018deep,FraudNE} \\ 

\midrule[1 pt]
\midrule[1 pt]
\multicolumn{3}{c}{\textbf{Techniques-specific Challenges}} \\
\midrule[1 pt]
Tech-CH1 & Anomaly-aware training objectives   & \cite{bandyopadhyay2020outlier,bandyopadhyay2019outlier,yu2018netwalk,liu2017accelerated,ding2020inductive,teng2018deep,zheng2019one,wang2019fdgars,DBLP:conf/ijcnn/Ouyang0020,zheng2019addgraph,gutierrez2020multi,hu2016embedding,wu2017adaptive,SemiGNN,teng2017anomaly,AANE,liang2018semi,peng2018anomalous,ding2019deep,li2019specae,ding2019interactive,fan2020anomalydae,zhang2019robust,peng2020deep,CARE-GNN,wang2018deep,FraudNE,dou2021user,zhu2020mixedad,GAL,zhao2020using} \\\cline{1-3}
Tech-CH2 & Anomaly interpretability            & \cite{hu2016embedding,SemiGNN} \\\cline{1-3}
Tech-CH3 & High training cost                  & \cite{yu2018netwalk,wang2019fdgars,wang2018deep,peng2020deep,chang2021f,zhu2020mixedad} \\\cline{1-3}
Tech-CH4 & Hyperparameter tuning              & \cite{zhao2020automating,GAL} \\
\bottomrule[1 pt]
\end{tabular}
\label{tb:challenges}
}
\end{table}

\section{Taxonomy} \label{appendix:taxonomy}
\begin{sidewaysfigure*}[h]
    \centering
    \includegraphics[width=\textwidth,keepaspectratio]{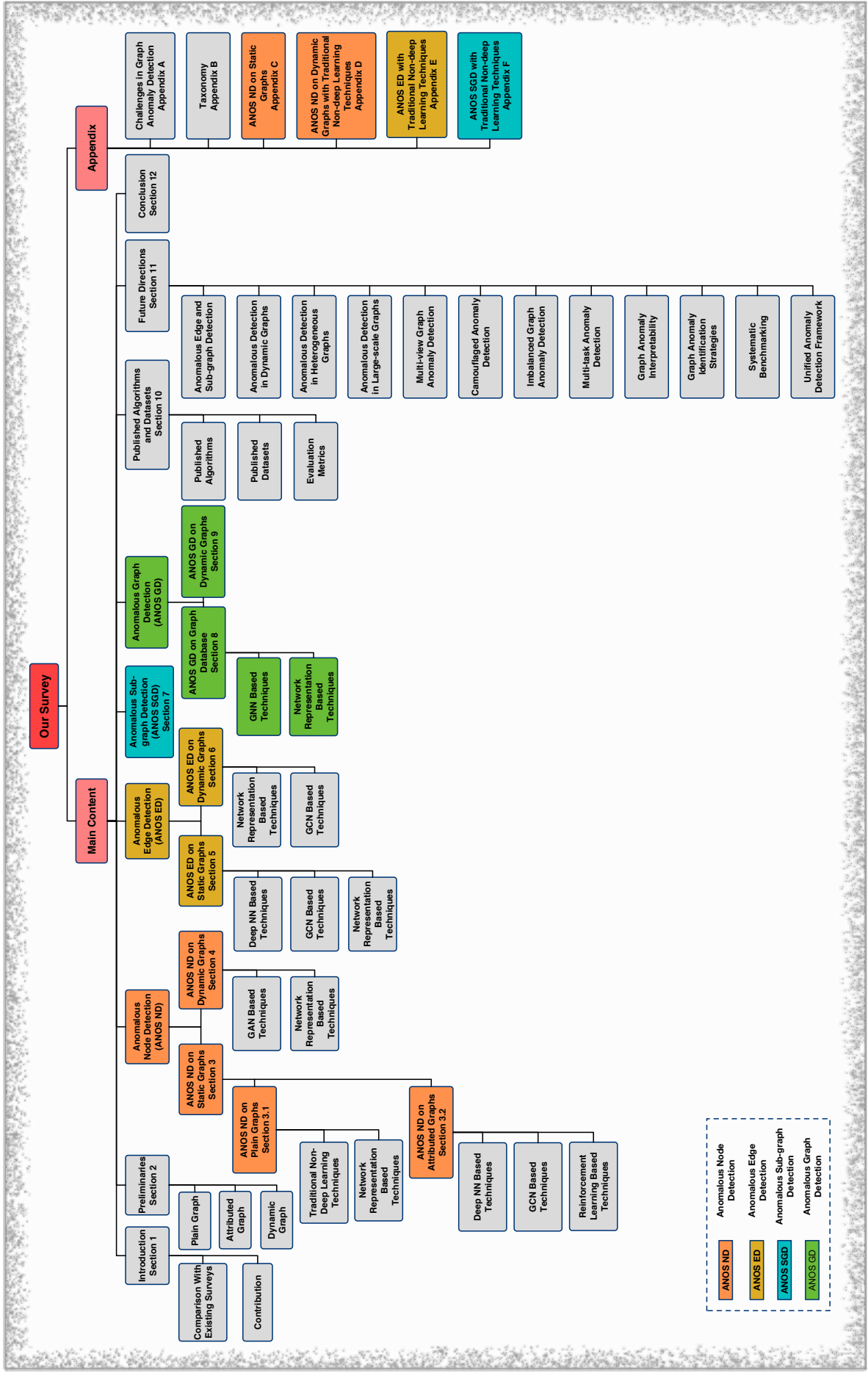}
    \captionof{figure}{The Taxonomy of Our Survey.}
    \label{pic:taxonomy}
\end{sidewaysfigure*}

The taxonomy of this survey is shown in Fig.~\ref{pic:taxonomy}.

\section{ANOS ND on Static Graphs} \label{appendix:node:static}
Following the taxonomy of Section~\ref{lb:node:sag}, this section reviews traditional non-deep learning techniques designed for ANOS ND on static attributed graphs, followed by techniques based on GATs, GANs, and network representation.

\onecolumn

\twocolumn
\subsection{Traditional Non-Deep Learning Techniques}
Traditional techniques, such as statistical models, matrix factorization, and KNN, have been widely applied to extract the structural/attribute patterns of anomalous nodes for a subsequent detection process.

\begin{figure}[!t]
\setlength{\belowcaptionskip}{-0.25cm}
  \centering
    \subfigure[The Framework of ALAD~\cite{liu2017accelerated}]{
    \label{pic:ALAD} 
    \includegraphics[width=0.48\textwidth]{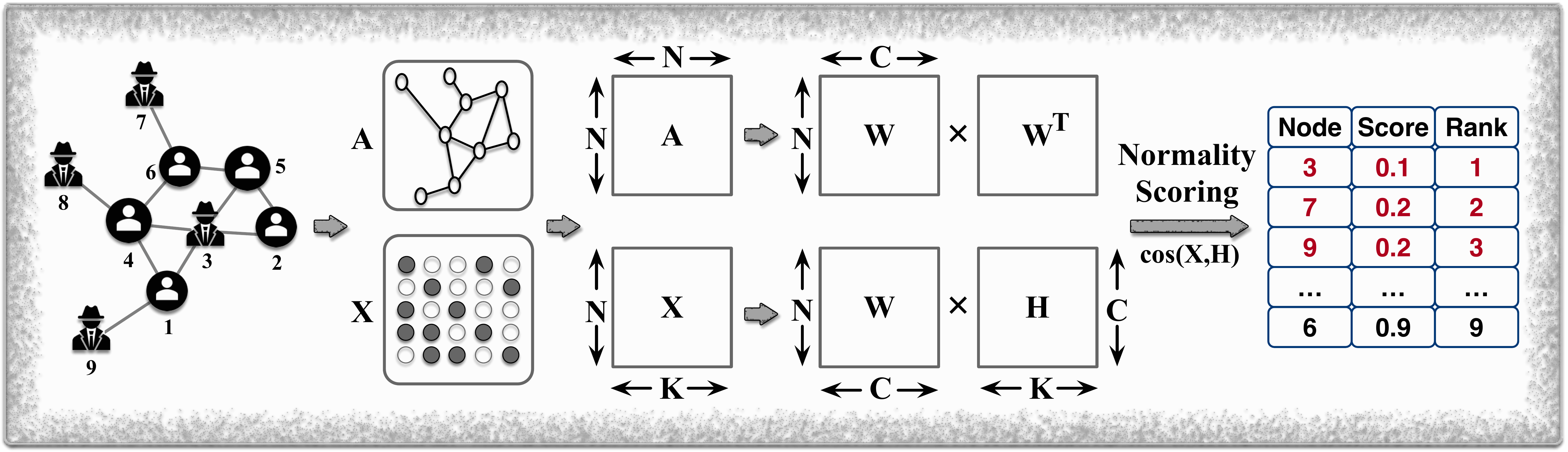}}
    \subfigure[The Framework of Radar~\cite{li2017radar}]{
    \label{pic:Radar} 
    \includegraphics[width=0.48\textwidth]{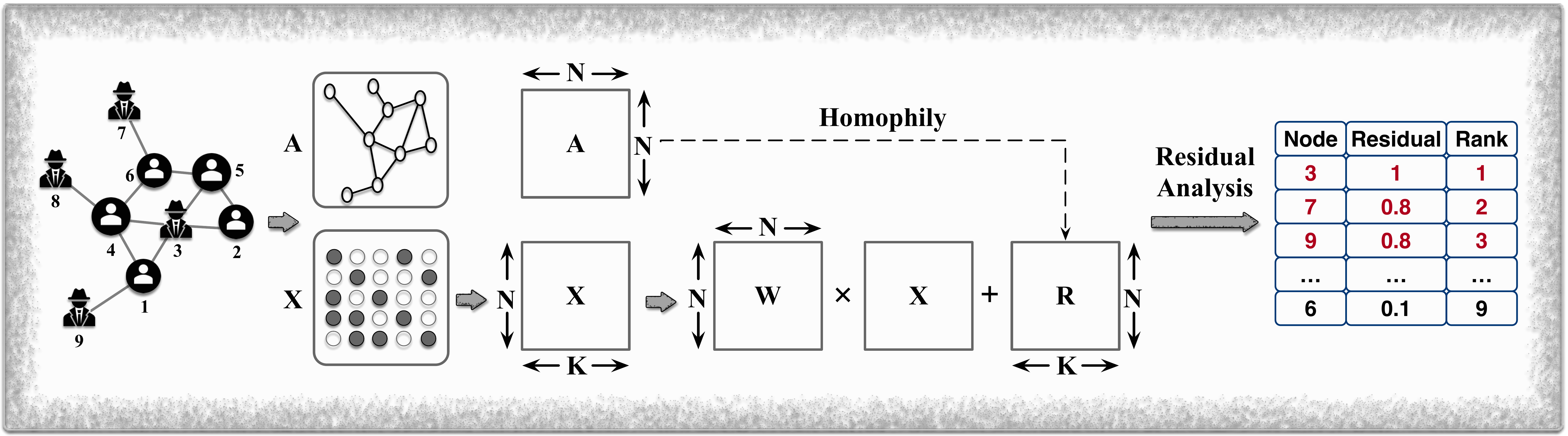}}
    \subfigure[The Framework of ANOMALOUS~\cite{peng2018anomalous}]{
    \label{pic:ANOMALOUS}
    \includegraphics[width=0.48\textwidth]{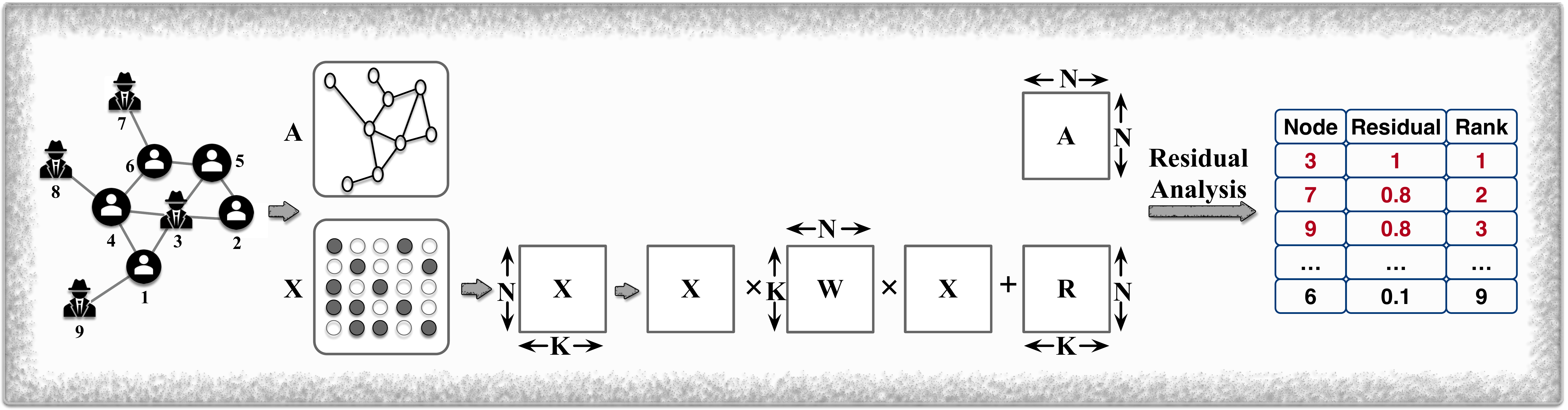}}
  \caption{ANOS ND on attributed graphs -- Matrix Factorization based approaches. The three representative MF techniques adopt different decomposition strategies to fuse the graph structural information and node attributes. Anomalous nodes are then spotted through scoring functions or residual analysis.}
  \label{pic:MFtechniques}
\end{figure}

Among these, matrix factorization (MF) based techniques have shown power at capturing both topological structures and node attributes, achieving promising detection performance.
An early attempt at this type of method was Liu \etal~\cite{liu2017accelerated}.
They aimed to detect community anomalies (defined in Section~\ref{sec:AND}) through the developed model, ALAD.
As shown in Fig.~\ref{pic:ALAD}, through non-negative matrix factorization, ALAD incorporates both the graph structure $A$ and node attributes $X$ to derive community structures $W$ and their attribute distribution vectors $H$.
When the matrices are decomposed, ALAD measures the normality of each node according to the attribute similarity between it and the community it belongs to.
By ranking the nodes' normality scores in ascending order, the top-k nodes are identified as community anomalies. 

Li \etal~\cite{li2017radar} approached ANOS ND from a different perspective using residual analysis.
As assumed, anomalies will lead to larger attribute reconstruction residual errors because they do not conform to the attribute patterns of the majorities.
Accordingly, the proposed model, Radar, learns the residual errors $R$, as shown in Fig.~\ref{pic:Radar}, by factorizing the node attributes $X$. To incorporate the structural information for obtaining these errors, Radar puts explicit restrictions on the learned residuals such that directly linked nodes will have similar residual patterns in $R$ (known as the homophily effect).
Finally, the top-k nodes with larger norms in $R$ are identified as anomalies.

Although the homophily hypothesis provides strong support for exploiting structural information, it might not always be hold.
In fact, real objects may have distinct attributes from their connected neighbors and it is non-trivial to regulate all connected objects share similar values in each dimension in the feature space. By this, Peng \etal~\cite{peng2018anomalous} indicated that there are structurally irrelevant node attributes that do not satisfy the homophily hypothesis.
Indeed, these structurally irrelevant node attributes would have adverse effects on anomaly detection techniques that are developed based on this hypothesis. 
To tackle this problem, their developed model, ANOMALOUS, uses CUR~\cite{DBLP:journals/pnas/MahoneyD09} decomposition to select attributes that are closely related to the network structure and then spot anomalies through residual analysis following Radar (as shown in Fig.~\ref{pic:ANOMALOUS}). 

Beyond matrix factorization, linear regression models have also been designed to train anomaly classifiers given labeled training data.
A representative work is Wu \etal~\cite{wu2017adaptive}.
Their supervised model, SGASD, has yielded encouraging results with identifying social spammers using the social network structure, the content information in social media and user labels.

These non-deep learning techniques are able to capture valuable information from graph topology and node attributes, but their application and generalizability to real-world networks (which are usually in large-scale) is strictly limited due to the high computational cost of matrix decomposition operations and regression models.

\subsection{GAT Based Techniques}

\begin{figure*}[!t]
\setlength{\belowcaptionskip}{-0.25cm}
    \centerline{\includegraphics[width=0.98\textwidth]{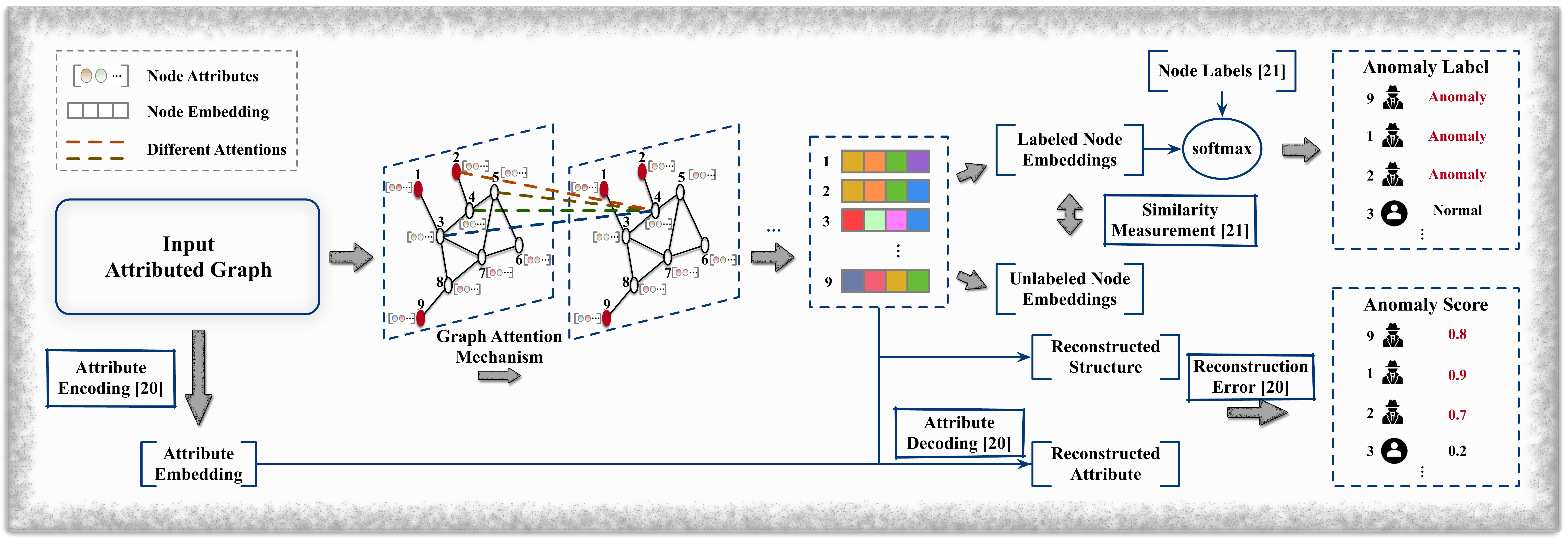}}
    \caption{ANOS ND on attributed graphs -- GAT based approaches. Given the input graph, these techniques employ graph attention neural network to learn node embeddings. The unsupervised technique, AnomalyDAE~\cite{fan2020anomalydae}, scores each node based on the reconstruction loss and mark the top-k nodes as anomalies, while the semi-supervised technique, SemiGNN~\cite{SemiGNN}, trains a classifier to predict node labels.}
    \label{pic:GAT}
\end{figure*}

Although GCN provides an effective solution to incorporating graph structure with node attributes for ANOS ND (reviewed in Section~\ref{sec:AND:GCN}), its ability to capture the most relevant information from neighboring is subpar.
This is due to the simple convolution operation that aggregates neighbor information equally to the target node.
Recently, graph attention mechanism (GAT)~\cite{velivckovic2017graph} is employed to replace the traditional graph convolution.
For instance, Fan \etal~\cite{fan2020anomalydae} applied graph attention neural network to encode the network structure information (structure encoding).
The method, AnomalyDAE, also adopts a separate attribute autoencoder to embed the node attributes (attribute encoding).
Through an unsupervised encoding-decoding process, each node is ranked according to its corresponding reconstruction loss, and the top-k nodes introducing the greatest losses are identified as anomalies.
Specifically, the attribute decoding process takes both node embeddings learned through the structure and attribute encoding processes to reconstruct node attributes, as shown in Fig.~\ref{pic:GAT}, while the graph topology is reconstructed only using the embeddings output by the GAT.
To acquire better reconstruction results, AnomalyDAE is trained to minimize the overall loss function, denoted as:
\begin{equation} \label{eq:anomalydae}
\begin{split}
    \mathcal{L}_{AnomalyDAE} = & \alpha ||(A - \hat{A})\odot \bm{\theta}||_{2}^{2} + \\
    &(1-\alpha)||(X - \hat{X})\odot \bm{\eta}||_{2}^{2},
\end{split}
\end{equation}
where $\alpha$ is the coefficient, $A$ and $X$ is the input adjacency matrix and attribute matrix, $\hat{A}$ and $\hat{X}$ are the reconstructed matrices. Each $\theta_{i,j} \in \bm{\theta}$ and $\eta_{i,j} \in \bm{\eta}$ is 1, if the corresponding element $A_{ij}$ and $X_{ij}$ equals 0, otherwise, their values are defined by hyperparameters greater than 1.

Another decent work is SemiGNN~\cite{SemiGNN}, in which Wang \etal proposed a semi-supervised attention-based graph neural network for detecting fraudulent users in online payment platforms.
This work further explores user information collected from various sources (\eg transaction information and user profiles), and represents real-networks as multi-view graphs.
Each view in the graph is modeled to reflect the relationship between users or the correlation between user attributes.
For anomaly detection, SemiGNN first generates node embedding $h_u^v$ from each view $v$ by aggregating neighbor information through a node-level attention mechanism.
It then employs view-level attention to aggregate node embeddings from each view and generates a unified representation $a_u$ for each node.
Lastly, the class of each node is predicted through a softmax classifier.
Indeed, Wang \etal designed a supervised classification loss and an unsupervised graph reconstruction loss to jointly optimize the model by fully utilizing labeled and unlabeled data.
The classification loss can be denoted as:
\begin{equation} 
    \mathcal{L}_{sup} = - \frac{1}{|U_L|}\sum_{u \in U_L} \sum_{i=1}^{k}I(y_u=i)\log\frac{\exp(a_u\cdot\theta_i)}{\sum_{j=1}^{k}\exp(a_u\cdot\theta_j)},
\end{equation}
where $U_L$ is the labeled user set and its size is $|U_L|$, $I(\cdot)$ is an indicator function, $k$ is the number of labels to be predicted (in most cases, the label is either anomalies or non-anomalies, and $k=2$), and $\theta$ represents the trainable variables.
Meanwhile, the unsupervised loss encourages unlabeled nodes (users) that can be reached by labeled nodes through random walks to obtain similar representations and vice versa.
This is achieved by negative sampling (unlabeled nodes that cannot be reached by random walks are negative samples) and the loss can be formulated as:
\begin{equation}
  \begin{split}
    \mathcal{L}_{unsup}  & = \sum_{u\in U}\sum_{v\in N_u \cup Neg_{u}} -\log(\sigma(a_{u}^{T}a_{v})) \\
    &- 3\cdot E_{q\sim P_{neg}(u)}\log(\sigma(a_{u}^{T}a_{q})),
  \end{split}
\end{equation}
where $U$ denotes the user set, $N_u$ denotes the neighbor set of $u$, $Neg_{u}$ represents negative samples, $P$ is the sampling distribution, and $\sigma(\cdot)$ is the sigmoid function. The total loss takes the sum of them and is formulated as:
\begin{equation} 
    \mathcal{L}_{SemiGNN} = \alpha \mathcal{L}_{sup} + (1- \alpha) \mathcal{L}_{unsup} + \lambda \mathcal{L}_{reg},
\end{equation}
where $\alpha$ is a balancing parameter and $\mathcal{L}_{reg}$ regularizes all trainable variables. 

\subsection{GAN Based Techniques}

\begin{figure*}[!t]
\setlength{\belowcaptionskip}{-0.25cm}
    \centerline{\includegraphics[width=0.98\textwidth]{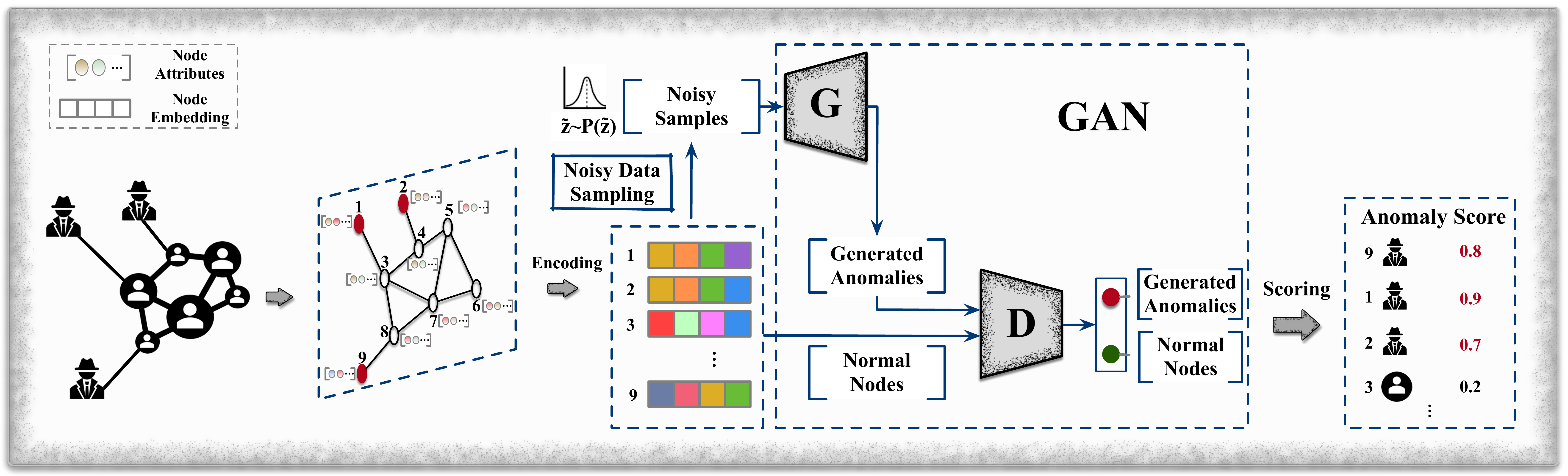}}
    \caption{ANOS ND on static graphs -- GAN based approaches. The generator, G, generates potential anomalies by sampling noisy data from a prior distribution to fool the discriminator. The discriminator, D, distinguishes the generated anomalies and normal nodes. The scoring function then quantifies the anomaly score of each node according to D's output.}
    \label{pic:GAN}
\end{figure*}

Because GAN is effective at capturing anomalous/regular data distributions (as reviewed in Section~\ref{sec:ND:GAN}), Ding \etal~\cite{ding2020inductive} used GAN in their developed model, AEGIS, for improved anomaly discriminability on unseen data.
As shown in Fig.~\ref{pic:GAN}, this model first generates node embeddings through a GNN from the input attributed graph, and then a generator and a discriminator are trained to identify anomalies.    In the first phase, anomalous nodes and regular nodes are mapped to distinctive areas in the embedding space such that the GAN is able to learn a boundary between them. Accordingly, Ding \etal built an autoencoder network with graph differentiative layers to capture the attribute difference between each node and its $k$-th neighbors. Such difference information enables anomalies to be distinguished easily. The embeddings are encoded as follows:
\begin{equation} 
    h_{i}^{l} = \sum_{k=1}^{K} \beta_{i}^{k} \sigma \left( W_1h_{i}^{l-1} + \sum_{j\in N_{k}(i)}\alpha_{ij}W_{2}\Delta_{ij}^{l-1}\right),
\end{equation}
where $h_{i}^{l}$ is embedded features of node $i$ through the $l$-th layer, $\beta_{i}^{k}$ is the attention for each hop, $N_{k}(i)$ is the set of $k$-th order neighbors, $\alpha_{ij}$ is the attention for each neighbor, and $\Delta_{ij}^{l-1}$ is the difference between $i$ and $j$, $W_1$ and $W_2$ are the trainable variables.
The autoencoder is fine-tuned until the node attributes can be best reconstructed using the learned embeddings, after which the GAN is trained.

In the second phase, the generator follows a prior distribution $p(\tilde{z})$ to generate anomalies by sampling noisy data, while the discriminator struggles to distinguish between the embeddings of the normal nodes and those of the generated anomalies. The training process is formulated as a mini-max game between the generator and discriminator as follows:
\begin{equation}
    \min_{G}\max_{D}\mathbb{E}_{z\sim Z}[\log(D(z))] + \mathbb{E}_{\tilde{z}\sim \tilde{Z}}[\log(1-D(G(\tilde{z})))],
\end{equation}
where $z$ is the node embeddings, and $\tilde{z}$ are the generated anomalies.
After training, AEGIS directly learns embedding $z_u$ for a test node $u$, and quantifies its anomaly score with regard to the discriminator's outputs, \ie the possibility that node is normal. The scoring function is formulated as:
\begin{equation} 
    Ascore(u) = 1 - D(z_u),
\end{equation}
and the top-k nodes are deemed to be anomalous.

\subsection{Network Representation Based Techniques}

With network representation, graphs are first encoded into a vector space before the anomalies detection procedure takes place.
As outlined in Section~\ref{sec:spg:nr}, numerous studies on ANOS ND in attributed graphs have exploited deep network representation techniques.

For instance, Zhang \etal~\cite{zhang2019robust} detected abnormal nodes that have attributes significantly deviating from their neighbors through a 3-layer neural network, REMAD, and residual analysis.
They explicitly divide the original node attribute matrix into a residual attribute matrix $R$ that captures the abnormal characters of anomalies and a structurally relevant attribute matrix $\hat{X}$ for network representation learning. Both matrices are jointly updated throughout the representation learning process so nearby nodes are encouraged to have similar representations.
Specifically, these node embeddings are generated by aggregating neighbor information with each node's own attributes, formulated as:
\begin{equation} 
    h_{i}^{l} = \sigma \left( W^l \cdot \text{CONCAT}\{h_{i}^{l-1},h_{N_i}^{l}\} + b^2\right),
\end{equation}
where $h_{i}^{l}$ is node $i$'s representation generated by the $l$-th layer ($h_{i}^{0} = \hat{X}$), $N_i$ contains $i$'s neighbors, $\sigma()$ is the activation function, $W^{l}$ and $b$ are the trainable variables. Finally, the residual matrix $R$ will contain the abnormal information of each node and the top-k nodes with the largest norms are considered anomalies.

Given partial node labels, Liang \etal~\cite{liang2018semi} developed a semi-supervised representation model, SEANO, that incorporates graph structure, node attributes and label information.
Similar to REMAD, SEANO also aggregates neighbor information to center nodes, and the node representations are obtained through an embedding layer, formulated as:
\begin{equation} \label{SEANO:embedding}
    z_{i} = \lambda_{i}h^{l_{1}}(x_{i}) + (1-\lambda_{i})h^{l_{1}}(\Bar{x}_{N_{i}}),
\end{equation}     
where $z_{i}$ is $i$'s representation, $\lambda_{i}$ is a trainable variable that identifies the weight of $i$'s own attributes ($x_{i}$), $\Bar{x}_{N_{i}}$ is the average of node $i$'s neighbors' representations, and the function $h^{k}(x_{i}) = \phi(W^kh^{k-1}(x_i)+b^k)$ maps original node attributes into lower dimensional vectors.
Then, a supervised component, which takes the representations as input, predicts node labels through a softmax classifier, and an unsupervised component is trained to reconstruct node contexts (node sequences).
The context of each node is not only generated through random walks on the graph but also from the labeled nodes that belong to the same class. After training, SEANO interprets $\lambda_{i}$ as the normality score of node $i$ and the top-k nodes with the highest scores are classed as anomalies.

Learning node representations via aggregating neighbor information has proven effective for capturing comprehensive information from graph structure and node attributes. 
But, Liu \etal~\cite{GraphConsis} demonstrated such an approach can help anomalies aggregate features from regular nodes, making them look normal and leading to sub-optimal detection performance. 
They identified three concrete issues that should be considered when applying aggregation operations for anomaly detection: 
1) Anomalies are rare objects in a network. Hence, directly aggregating neighborhood information will smooth the difference between the anomalies and normal instances, blurring boundaries between them. 
2) Directly connected nodes have distinctive features, and the assumption that connected nodes share similar features, which serves as the basis for feature aggregation, no longer holds in this scenario.
3) Real objects also form multiple types of relations with others, which means aggregation results for different types of relations will be distinctive.
With regard to these concerns, their proposed method, GraphConsis, follows a sampling strategy to avoid potential anomalous neighbors when aggregating node features.
This method also adopts an attention mechanism to aggregate neighbor information following different links.
The learned node representations, therefore, are more robust to anomalies. As such, GraphConsis takes them as input to train a classifier for predicting labels.

Dou \etal~\cite{CARE-GNN} further considered camouflage behaviors of fraudsters in their proposed model CARE-GNN to enhance detection performance.
As specified, the camouflages can be categorized as either feature camouflage or relation camouflage.
Respectively, anomalies either adjust their feature information or form connections with many benign objects to gloss over suspicious information.
Hence, directly employing aggregation will overlook the camouflages and smooth the abnormal patterns of anomalies, eliminating the distinctions between anomalies and normal objects. 
To alleviate over-smoothness, CARE-GNN also adopts a neighbor sampling strategy, as is the case with GraphConsis, to filter camouflaged anomalies and explores different types of relations formed between users.
Specifically, under each relation, Dou \etal employed a MLP to predict node labels using their features and measure the similarity ($l1$ distance) between each node and its neighbors according to the MLP's output.
Then, the top-k most similar neighbors are selected for feature aggregation, and CARE-GNN generates each node's representation through a combination of latent representations that are learned under different relations.
A classifier is eventually trained using the representations to predict the node labels.

As can be seen, the performance of these network representation based techniques is decided by their training objectives/loss functions.
Enhanced detection performance is probable if the loss function is able to separate normal nodes from abnormal nodes reasonably well. 
Motivated by this, a more recent work in~\cite{GAL} emphasizes the importance of anomaly-aware loss functions.
In order to adjust margins for the anomalies, the authors proposed a novel loss function to guide the representation learning process.
Specifically, this loss function is designed to find the relative scales between the margins of outlier nodes and normal nodes.
An MLP-based classifier is finally trained using the node representations generated by the anomaly-aware loss-guided GNNs and node labels. 
For unseen nodes, the classifier will label them upon their representations.

\section{ANOS ND on Dynamic Graphs With Traditional Non-Deep Learning Techniques} \label{appendix:node:dynamic}

To detect anomalous nodes in dynamic plain graphs, traditional non-deep learning techniques rely heavily on modeling the structural evolving patterns of nodes.
Representative works like~\cite{DBLP:conf/wsdm/RossiGNH13} and~\cite{wang2019detecting} assume that the evolutionary behaviors of regular nodes'  (\ie generate or remove connections with others) usually follow stable patterns, and their changes introduce less impact on the graph structure compared to anomalies.
Specifically, in~\cite{wang2019detecting}, Wang \etal proposed a novel link prediction method to fit the evolutionary patterns of most nodes such that anomalies can be identified because their observations significantly conflict with the prediction results.
They further quantified the impact of anomalous behaviors by assessing the perturbation imposed on the graph adjacency matrix.

Other traditional works also exploit node/edge attributes and their changes. For examples, Teng \etal~\cite{teng2017anomaly} took node and edge attributes as two different views to describe each node. By encoding both types of information into a shared latent space, their proposed model learns a hypersphere from historical records. When new observations of existing nodes are given, the model distinguishes between the benign and the anomalies according to their distances to the hypersphere centroid. Points lying outside the hypersphere are spotted as anomalies.

Unlike embedding techniques, Nguyen \etal~\cite{tam2019anomaly} proposed a non-parametric method to detect anomalous users, tweets, hashtags, and links on social platforms. Specifically, they modeled social platforms as heterogeneous social graphs such that the affluent relationships between users, tweets, hashtags and links were effectively captured. Through extensive analysis of the features, such as user registration information, keywords in tweets, the linguistic style of links and the popularity scores of hashtags, anomalous objects are spotted based on their deviating features. This work also uses relationships between individual objects as well as the detected anomalies, and detects groups of abnormal objects.

\begin{figure}[!t]
    \centerline{\includegraphics[width=0.48\textwidth]{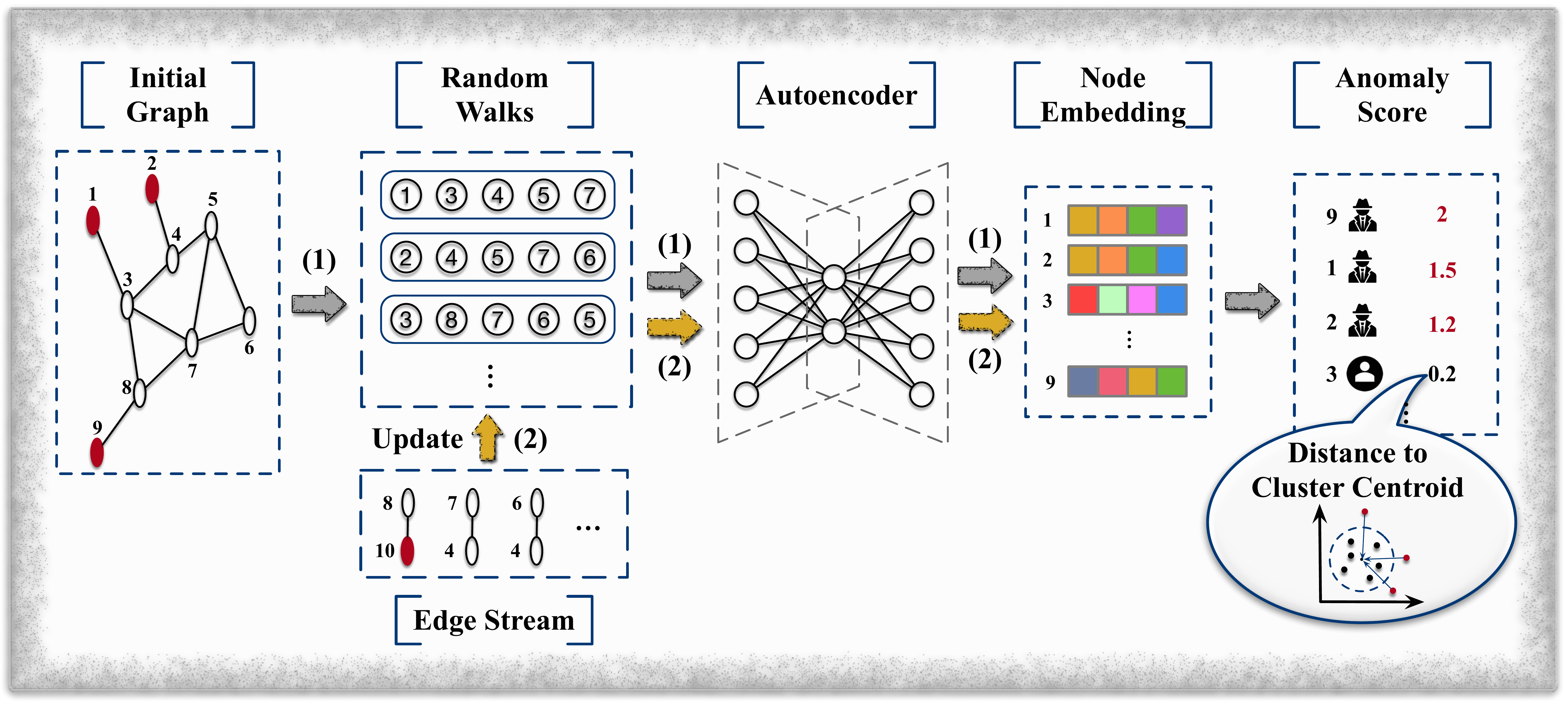}}
    \caption{The Framework of NetWalk~\cite{yu2018netwalk}. The input dynamic graph is modeled as an initial graph with an incoming edge stream. Given the initial graph, a trunk of node sequences is generated using random walks on the graph and the deep autoencoder encodes each node into an embedding space following process (1). The node sequences and node embeddings are updated based on the incoming edge stream following process (2). Finally, NetWalk assigns an anomaly score to each node according to its distance to the cluster centroid in the embedding space.}
    \label{pic:netwalk}
\end{figure}

\section{ANOS ED With Traditional Non-Deep Learning Techniques} \label{appendix:edge}

Traditional non-deep learning based approaches mainly focus on using temporal signals (\eg changes in graph structure), and applying specially designed statistical metrics to detect anomalous edges on dynamic graphs~\cite{ranshous2016scalable, aggarwal2011outlier}. 
As a concrete example, Eswaran and Faloutsos~\cite{eswaran2018sedanspot} modeled a dynamic graph as a stream of edges and exploited the graph structure as well as the structure evolving patterns. They identified two signs of anomalous edges: 1) connecting regions of the graph that were disconnected; and 2) connections that appear in bursts. For incoming edges, their model assigns anomaly scores to each edge, and the top-k edges with highest scores are anomalies. Another most recent work by Chang \etal~\cite{chang2021f} proposed a novel frequency factorization algorithm, aiming to spot anomalous incoming edges based on their likelihood of observed frequency. This method merges the advantages of both probabilistic models and matrix factorization for capturing both temporal and structural changes of nodes, and as reported, it only requires constant memory to handle edge streams.

\section{ANOS SGD With Traditional Non-Deep Learning Techniques} \label{appendix:subgraph}

\subsection{ANOS SGD on Static Graphs} \label{appendix:subgraph:static}

One motivation of ANOS SGD in static graphs is that anomalous sub-graphs often exhibit significantly different attribute distributions.
Therefore, traditional non-deep learning techniques, such as gAnomaly~\cite{li2014probabilistic}, AMEN~\cite{perozzi2016scalable}, and SLICENDICE~\cite{nilforoshan2019slicendice}, focus on modeling the attribute distributions and measuring the normality of sub-graphs.
Another line of investigation is graph residual analysis. The rich attribute information contained in real-world networks provides insight into the relationships formed between objects. Thus, the motivation behind several studies to spot anomalous sub-graphs has been to measure the residual between the expected structures and observed structures~\cite{miller2013efficient}.

\subsection{ANOS SGD on Dynamic Graphs} \label{appendix:subgraph:dynamic}

Devising metrics for ANOS SGD has been the subject of many traditional works. 
For instance, Chen \etal~\cite{DBLP:journals/jiis/ChenHS12} introduced six metrics to identify community-based anomalies, namely: grown community, shrunken community, merged community, split community, born community and vanished community.
Although these hand-crafted features or statistical patterns well fit some particular types of existing anomalies, their abilities to detect unseen and camouflage anomalies are limited and applying them directly might introduce high false negative rate, which is not optimal for applications like financial security. 
Other works, such as SPOTLIGHT by Eswaran \etal~\cite{eswaran2018spotlight} and another by Liu \etal~\cite{liu2008spotting}, explore sudden changes in dynamic graphs and identify anomalous sub-graphs that are related to such changes.

Motivated by the phenomena that social spam and fraud groups often form dense temporal sub-graphs in online social networks, plenty of works, including,~\cite{Densealert,mongiovi2013netspot}, use manually-extracted features and spot anomalous dense sub-graphs that have evolved significantly different from the reset of the graph. 

In addition to these studies, a large number of works discuss uses of various graph scan statistics for anomalous sub-graph detection, such as the Kulldorff statistic~\cite{kulldorff1997spatial}, Poisson statistic~\cite{neill2005detection}, elevated mean scan statistic~\cite{sharpnack2013near} and Berk-Jones statistic~\cite{berk1979goodness}. Specifically, Shao \etal~\cite{dGraphScan} proposed a non-parametric method to detect anomalous sub-graphs in dynamic graphs where the network structure is constant, but the node attributes change overtime. This approach measures the anomalous score of each sub-graph with regard to the p-values of nodes it comprises. Sub-graphs with higher scores are more anomalous. Another work, GBGP~\cite{GBGP}, instead, adopts the elevated mean scan statistic to identify nodes that might form anomalous sub-graphs and detects anomalous groups that follow predefined irregular structures.

\end{document}